\documentclass{iopjournal}

\usepackage{hyperref}
\usepackage{url}

\usepackage{amssymb}
\usepackage{amsmath}
\usepackage[utf8]{inputenc} % allow utf-8 input
\usepackage[T1]{fontenc}    % use 8-bit T1 fonts
\usepackage{booktabs}       % professional-quality tables
\usepackage{amsfonts}       % blackboard math symbols
\usepackage{nicefrac}       % compact symbols for 1/2, etc.
\usepackage{microtype}      % microtypography
\usepackage{lipsum}		% Can be removed after putting your text content
\usepackage{graphicx}
\usepackage{natbib}
\usepackage{doi}
\usepackage[dvipsnames, svgnames, x11names, table]{xcolor}
\usepackage{todonotes}
\usepackage{epstopdf, epsfig}
\usepackage{verbatim}
\usepackage[hang,splitrule,symbol]{footmisc}
\usepackage{amssymb}
\usepackage{multirow}
\usepackage{lineno}
\usepackage{mathrsfs}
\usepackage{graphicx}
\usepackage{adjustbox}
\usepackage{subcaption}
\usepackage{makecell}
\usepackage{array}

\usepackage{algorithm}
\usepackage{algpseudocode}
\usepackage{algorithmicx}
\usepackage{enumerate}
\usepackage{bigints}

\usepackage{xspace}
\usepackage{tcolorbox}
\usepackage{lipsum}
\usepackage{listings}

\usepackage{tikz}
\usetikzlibrary{patterns, arrows, shapes.misc}

\usepackage{longtable}
\begin{document}

\articletype{Article type} %	 e.g. Paper, Letter, Topical Review...

\title{Re4: Scientific Computing Agent with Rewriting, Resolution, Review and Revision}

\author{Ao Cheng$^{1,2*}$\orcid{0009-0001-6759-2595}, Lei Zhang$^{1,2*}$\orcid{0000-0001-8291-6124} Guowei He$^{1,2*}$\orcid{0000-0003-4738-0816}}

\affil{$^1$The State Key Laboratory of Nonlinear Mechanics, Institute of Mechanics, Chinese Academy of Sciences, Beijing 100190, China}

\affil{$^2$School of Engineering Science, University of Chinese Academy of Sciences, Beijing 100049, China}

\affil{$^*$Author to whom any correspondence should be addressed.}

\email{chengao23@mails.ucas.ac.cn; zhanglei@imech.ac.cn; hgw@lnm.imech.ac.cn}


\begin{abstract}
Large language models (LLMs) are a promising class of generative artificial intelligence models and have demonstrated abilities to perform complex tasks in multiple domains, including mathematical and scientific reasoning. In this work, we construct a novel agent framework for solving representative problems in scientific computing. The proposed agent, incorporating a ``rewriting-resolution-review-revision'' logical chain via three reasoning LLMs (functioning as the Consultant, Reviewer, and Programmer, respectively), is integrated in a collaborative and interactive manner. The Consultant module endows the agent with knowledge transfer capabilities to link problems to professional domain insights, thereby rewriting problem descriptions through text augmentation. The Programmer module is responsible for generating and executing well-structured code to deliver the problem resolution. The Reviewer module equips the agent with the capacity for self-debugging and self-refinement through interactive feedback with code runtime outputs. By leveraging the end-to-end review mechanism, the executable code provided by the Programmer undergoes iterative revision. A comprehensive evaluation is conducted on the performance of the proposed agent framework in solving typical partial differential equations (PDEs), \textcolor{black}{engineering-oriented flow past a cylinder}, ill-conditioned linear systems, and data-driven physical analysis problems. 
Compared with a single-model approach, this collaborative framework significantly improves the bug-free code generation rate and reduces the occurrence of non-physical solutions, thereby establishing a highly reliable framework for autonomous code generation based on natural language descriptions. 
The review mechanism improved the average execution success (bug-free code and non-NaN solutions) rate of the modern reasoning model DeepSeek R1 from $59\%$ to $82\%$, while ChatGPT 4.1-mini improved from $66\%$ to $87\%$ and Gemini-2.5 improved from $60\%$ to $84\%$. 
Automatic code generation and review provide a promising workflow for scientific computing; analysis of solver-development trajectories further reveals recurring implementation difficulties in boundary treatment, array indexing, and the preservation of numerical constraints.

\end{abstract}

\section{Introduction}\label{sec:Introduction}
Scientific computing, as a cornerstone of modern science and engineering, enables the modeling, simulation, and analysis of complex physical phenomena, including fluid dynamics~\citep{Brunton_etal_ARFM_2020}, computational mechanics~\citep{zhang2022hidenn}, electromagnetism~\citep{Oskooi_etal_CPC_2010}, materials science~\citep{ramprasad2017machine} and quantum computing~\citep{Meng_Yang_PRR_2023, meng2026toward}. 
Examples in turbulent flows include passive-scalar transport~\citep{wang25solvable}, Reynolds-stress modeling for separated and secondary flows~\citep{Liu2026factorized}, and data-driven wall modeling for large-eddy simulation~\citep{zhou2025wall, zhang2025knowledge, zhang2026differentiable}. However, solving scientific computing problems typically requires profound domain expertise, sophisticated algorithm design, and rigorous code implementation. 
In recent years, large language models (LLMs)~\citep{zhao2023survey} with advanced reasoning capabilities have emerged as transformative and low-barrier tools in scientific computing: automatically generating code based on a small amount of vague natural language descriptions. Implementing a natural-language-based solving paradigm faces two major challenges: (1) how to autonomously select and implement appropriate methods for specific problems; and (2) how to generate bug-free code. The first challenge demands the mastery and application of extensive domain-specific knowledge, necessitating that the LLMs complete the entire reasoning chain-from determining the type of equation (e.g., elliptic, parabolic, hyperbolic) to selecting the appropriate numerical method-without human intervention. The second challenge is a critical prerequisite for achieving autonomous scientific computing, requiring the model to translate vague natural language into precise code language.

Since the advent of LLMs like ChatGPT~\citep{achiam2023gpt}, researchers have begun to explore generating code through natural language descriptions~\citep{chen2021evaluating,wang2023codet5+,roziere2023code}. Early models heavily relied on meticulous prompt engineering~\citep{wei2022chain,kojima2022large}, requiring manual specification of numerical methods and model parameters, often resulting in logical and syntactical errors. At that stage, LLMs essentially functioned merely as code generation tools, providing rough program templates based on detailed customization from researchers, with low autonomy and necessitating extensive modification and debugging by users\citep{jiang2024survey}. Recently, the excellent reasoning capabilities of reasoning LLMs have shown promise in addressing autonomy issues. Studies by Jiang et al. \citep{Jiang_etal_TAML_2025} and Wang et al. \citep{Wang_etal_TAML_2025} have tested the capabilities of these reasoning models in scientific computing, indicating that they generally outperform non-reasoning models. However, it was found that even when addressing classical problems with established solutions, the code generated by reasoning models frequently contained errors requiring human correction. Furthermore, in tests conducted by Wang et al., reasoning models such as DeepSeek R1~\citep{guo2025deepseek} and OpenAI's o3-mini-high~\citep{openai_o3_mini_2025} proved ineffective in solving challenging problems like the Hilbert linear algebra problem, with DeepSeek R1 proposing potential strategies during thinking but failing to implement them. In addition to low bug-free rates in code generation, reasoning models currently face issues such as strong randomness in the selection of numerical methods, unstable output results, and reasoning hallucinations. \textcolor{black}{To mitigate such hallucinations, recent studies have explored robust query enrichment techniques driven by sparse autoencoders, which steer the language model toward semantically grounded features rather than misleading contexts~\citep{abdaljalil2025safe}. Without such mitigation strategies,} these factors can lead to the selection of suboptimal numerical schemes, thereby significantly affecting the reliability of LLMs in scientific computing.

The agent framework based on LLMs provides the capability to interact with external systems. Prompt-engineering-based interactive feedback enables self-refinement without the need for extra training. For natural language processing tasks (e.g., dialogue response or code readability), Madaan et al.~\citep{madaan2023self} leveraged a single non-reasoning LLM to generate an initial output, provided self-feedback, and iteratively optimized subsequent outputs using historical feedback and prior iterative outputs. Such a self-feedback framework, by simulating the iterative process of ``draft-reflection-revision'' of humans, significantly improves the quality of outputs compared with the traditional one-step generation approach. By encoding Standardized Operating Procedures (SOPs) into prompt sequences, the multi-agent collaboration framework MetaGPT~\citep{hong2023metagpt} empowered LLMs to simulate human professional roles, \textcolor{black}{deliver} structured outputs, and \textcolor{black}{collaborate} in an assembly-line manner, thereby reducing errors and boosting task efficiency. By assigning unique roles to multiple LLMs, the ChatEval framework constructed a multi-agent referee team that enhanced text evaluation accuracy through collaborative debate, and experimental results underscore the strengths of multi-agent collaboration~\citep{chan2023chateval}. \textcolor{black}{This collaborative architecture with specialized roles has recently shown significant promise across diverse domains. For instance, the xChemAgents framework employs a cooperative ``Selector-Validator'' architecture to achieve physically-grounded, explainable feature selection in quantum chemistry~\citep{polat2025xchemagents}. Similarly, Theorem-of-Thought (ToTh) structures complex inference by deploying independent abductive, deductive, and inductive agents to verify logical consistency through Bayesian belief propagation~\citep{abdaljalil2025theorem}.} In the field of scientific computing, such as computational fluid dynamics, an LLM-driven framework enabled end-to-end output from natural language input to user-interactive wall-bounded turbulence modeling problems~\citep{yang2025large}. Xu et al.~\citep{xu2025cfdagent} proposed a zero-shot, natural language-driven multi-agent system called CFDagent, which, coordinated by LLMs, completes the end-to-end process from natural language prompts to fully automated computational fluid dynamics (CFD) simulations. For solving partial differential equations (PDEs), the PINNsAgent~\citep{wuwu2025pinnsagent}, which integrates physics-informed knowledge replay and memory tree reasoning strategies, serves as an LLM-based surrogate framework to automatically construct and optimize physics-informed neural networks (PINNs) architectures, effectively bridging the gap between domain-specific knowledge and deep learning expertise. The task-specific-tuning-free framework CodePDE was validated on a range of representative PDE problems by enabling LLMs to perform iterative debugging~\citep{Li_etal_arXiv_2025}. The best solutions achieved by this framework on multiple PDE problems reach or exceed human levels, indicating the potential of LLM-based agents in solving PDEs. However, existing applications of LLMs in scientific computing often lack a structured framework to ensure the designed algorithms \textcolor{black}{are} adaptable to diverse tasks and code refinement through feedback that interacts with runtime results.

To address the autonomy and reliability issues encountered by LLMs in achieving solutions autonomously, while leveraging the merits of multi-step multi-agent collaboration, we propose a novel agent framework for scientific computing. The proposed framework, named \textbf{Re4}, is structured around a four-stage logical chain (\textbf{Re}writing, \textbf{Re}solution, \textbf{Re}view, and \textbf{Re}vision) orchestrated through three collaborative, role-decomposed modules (the Consultant, the Programmer, and the Reviewer):

$\bullet$ \textbf{Rewriting}. By expanding the task background through adaptive integration of domain knowledge and formulating structured alternative algorithms, the agent's understanding of specific scientific computing problems is significantly deepened. 

$\bullet$ \textbf{Resolution}. In the initial response, Python code \textcolor{black}{is} generated leveraging the augmented task text and run in the terminal, while capturing the runtime output text for subsequent assessment. 

$\bullet$ \textbf{Review}. As an independent third party, an independent LLM evaluates the solution quality of the designed algorithm, with suggestions for the code's implementation details, thereby endowing the agent with the traits of self-debugging and self-refinement. 

$\bullet$ \textbf{Revision}. In the review-revision loop, the executable code \textcolor{black}{is} comprehensively enhanced in execution/solving success rate, readability, modularity, and solution accuracy through feedback-driven refinement. 

We quantitatively assess the performance of the agent \textcolor{black}{both} with and without the Reviewer module in autonomously solving PDEs, challenging Hilbert systems and conducting data-driven physical analysis.
Apart from typical textbook-style PDEs, we further extend our evaluation to a realistic engineering scenario, namely flow past a circular cylinder. For this task, the Programmer is explicitly required to implement a solver utilizing staggered marker-and-cell (MAC) grids in conjunction with the immersed boundary method (IBM)~\citep{peskin2002immersed,mittal2005immersed}.
This assessment encompasses various dimensions, including bug-free rate, occurrence of non-physical solutions, high-precision solution generation rate (with errors below a certain threshold), and solution accuracy. Through multi-model cross-validation and the implementation of self-debugging, the review mechanism improved the average execution success (bug-free code and non-NaN solutions) rate of the modern reasoning models: DeepSeek R1 improved from $59\%$ to $82\%$, ChatGPT 4.1-mini improved from $66\%$ to $87\%$ and Gemini-2.5 improved from $60\%$ to $84\%$. Additionally, this framework significantly increased the probability of selecting high-precision methods, resulting in a noticeable improvement in average solution accuracy compared to single-model baselines. Furthermore, when extended to data-driven analyses of governing physical relationships, the agent demonstrates a robust capacity to reliably identify dominant dimensionless quantities from raw experimental data.

\begin{table}[!htbp]
\centering
\scriptsize
\renewcommand{\arraystretch}{1.6} 
\setlength{\tabcolsep}{2pt}
\begin{tabular}{c|c|c|c|c|c|c|c}
\cline{1-8}
& \textbf{Application Scope} & \textbf{Reasoning} & \textbf{Expansion} & \textbf{Debugging} & \textbf{Refinement} & \textbf{Review} & \makecell{\textbf{Single/Multiple}\\ \textbf{LLMs}} \\ \cline{1-8}

\makecell{Non-reasoning\\ LLMs} & General NLP & $\times$ & $\times$ & $\times$ & $\times$ & $\times$ & Single \\ \cline{1-8}

\makecell{Reasoning\\ LLMs} & \makecell{Complex reasoning\\ tasks} & $\checkmark$ & $\times$ & $\times$ & $\times$ & $\times$ & Single \\ \cline{1-8}

\makecell{CodePDE\\ Agent} & \makecell{From-scratch\\ code generation} & $\checkmark$ & $\times$ & $\checkmark$ & \makecell{based on\\ posterior error} & $\times$ & Single \\ \cline{1-8}

\makecell{PINNsAgent} & \makecell{NN architecture \&\\ hyperparameter tuning} & $\checkmark$ & $\times$ & $\checkmark$ & $\checkmark$ & $\times$ & Single \\ \cline{1-8}

\makecell{Madaan's\\ Agent} & \makecell{General NLP\\ refinement} & $\times$ & $\times$ & $\times$ & $\checkmark$ & $\times$ & Single \\ \cline{1-8}

\makecell{CFDAgent} & \makecell{CFD simulation via\\ external software} & $\checkmark$ & $\times$ & $\times$ & $\times$ & $\times$ & Multiple \\ \cline{1-8}

\makecell{\textbf{Our Agent}\\ \textbf{(Re4)}} & \makecell{From-scratch\\ code generation} & $\checkmark$ & $\checkmark$ & $\checkmark$ & $\checkmark$ & $\checkmark$ & \textbf{Multiple} \\ \cline{1-8}

\end{tabular}
\caption{\textcolor{black}{Comparison of application scopes and capabilities among existing LLM-based frameworks and the proposed Re4 agent.}}
\label{tab:agents_compare}
\end{table}

A detailed comparison of current LLM-based agents' capabilities is presented in table~\ref{tab:agents_compare}. Our contributions are as follows:

(1) We introduce a role-decomposed multi-agent framework for scientific computing that incorporates a ``rewriting-resolution-review-revision'' logical chain, effectively improving the bug-free code generation rate and reducing the occurrence of non-physical solutions, thereby establishing a highly reliable framework for autonomous code generation based on natural language descriptions.

(2) We present a robust multi-LLM collaborative workflow for scientific computing, demonstrating that the orchestrated cooperation among heterogeneous models effectively enhances both code execution reliability and solution accuracy compared to single-model baselines.

(3) We extend the agent framework to data-driven physical analysis, demonstrating its generality and applicability in uncovering governing physical relationships.

\textcolor{black}{Tracing the trial-and-error process of solver development reveals how an established numerical method can be implemented inconsistently and what checks are needed after a proposed repair. Across the evaluated tasks, the observed difficulties involve mapping staggered variables to arrays, maintaining consistency in boundary treatment and pressure–velocity operators, and handling ill-conditioning and dimensional-homogeneity constraints. These trajectories expose implementation decisions that aggregate success rates alone cannot explain.}

The rest of the paper is organized as follows: the construction of the agent framework is introduced in Section~\ref{sec:Methodology}; the evaluation on different scientific computing problems is presented in Section~\ref{sec:Experiments}; the ablation study and cost analysis are provided in Section~\ref{sec:ablation_tokenct_}; the solver-development trajectories are discussed in Section~\ref{sec:discussion}; and the conclusions are drawn in Section~\ref{sec:Conclusion}.

\section{Methodology}\label{sec:Methodology}
{\color{black}
To autonomously solve complex scientific computing problems from natural language descriptions, we construct the \textbf{Re4} framework, whose architecture comprises three LLM-based modules: the Consultant clarifies the problem and proposes solution plans; the Programmer generates and revises code; and the Reviewer assesses the code and execution results. As illustrated in figure~\ref{Fig_schematic_agent}, their interactions form a four-stage logical reasoning chain, represented by the acronym \textbf{Re4}: \textbf{Re}writing, \textbf{Re}solution, \textbf{Re}view, and \textbf{Re}vision. Distinct from frameworks that decompose roles based on cognitive reasoning types (e.g., abductive, deductive, and inductive agents~\citep{abdaljalil2025theorem}), Re4 assigns its module roles according to the practical workflow of solving scientific computing problems.

\begin{figure}[!t]
\centering{\includegraphics[width=\textwidth]{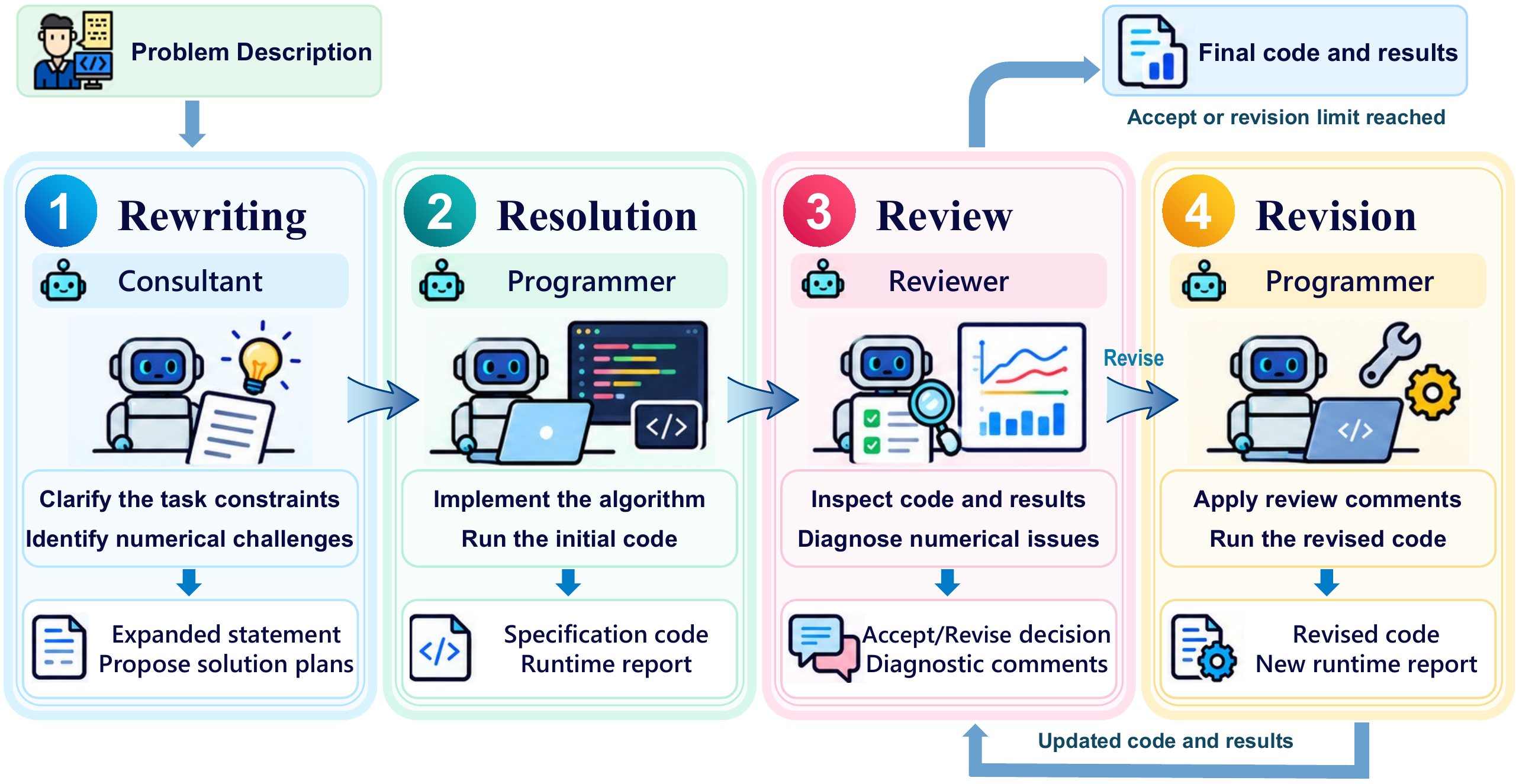}}
\caption{\textcolor{black}{Architecture and workflow of Re4. The three LLM-based modules are the Consultant (problem clarification and solution planning), Programmer (code generation and revision), and Reviewer (code and result assessment). Their interactions are organized into four stages: Rewriting, Resolution, Review, and Revision. The Programmer participates in both Resolution and Revision, with code execution handled by the Python execution tool.}}
\label{Fig_schematic_agent}
\end{figure}

\begin{enumerate}[(i)]
\item \textbf{Rewriting (Consultant).} The Consultant expands the original problem statement and proposes alternative numerical solution plans. This stage makes the mathematical challenges and task constraints explicit before code generation. Domain-specific guidance helps the Programmer assess numerical alternatives before committing to an implementation. It gives the Programmer an organized starting point for selecting and implementing a method.
\item \textbf{Resolution (Programmer and Python execution tool).} The Programmer selects a solution plan and produces an initial Python solver with a concise technical specification. This stage translates the numerical design into an implementation that can be tested. The execution tool runs the code and collects a runtime report of numerical outputs and errors, with optional textual assessment of the generated figures. These concrete execution results supply the evidence needed for the subsequent review.
\item \textbf{Review (Reviewer).} The Reviewer examines the problem context, technical specification, generated code, and execution results. This stage is needed because executable code can still violate numerical or physical requirements. The Reviewer returns an accept-or-revise decision and diagnostic comments addressing runtime failures, code structure, and numerical accuracy. Its independent role and explicit feedback give the Programmer actionable targets for refinement.
\item \textbf{Revision (Programmer and Python execution tool).} The Programmer uses the review comments and previous execution results to revise the solver. This stage addresses failures identified in the current candidate without restarting the problem formulation. The revised code is executed again and its new runtime report is returned to Review. Repetition permits both correction and detection of remaining or newly introduced problems.
\end{enumerate}

The agent is orchestrated via LangGraph~\citep{langgraph_2025}, where each operational module is defined as a functional node, and the workflow is governed through conditional edges. The general workflow can terminate when the Reviewer returns an ``Accept'' decision or when a preset revision limit is reached. For the stage-wise evaluation reported here, however, we retain two revision rounds (\texttt{rev-1} and \texttt{rev-2}) for all cases to enable consistent comparison across successive candidates. Additional implementation details, including routing logic, context isolation, structured outputs, and runtime-log handling, are provided in Appendix~\ref{app:implementation}, while the full role prompts are given in Appendix~\ref{app:prompt_for_agent}.
}

Regarding the configuration of LLMs, distinct from traditional single-model workflows, our framework supports a collaborative multi-model architecture, allowing different LLMs to operate within specific functional nodes to leverage their respective strengths. Given the great potential of reasoning LLMs~\citep{Wang_etal_TAML_2025,Jiang_etal_TAML_2025}, we primarily evaluate the agent's performance using \textbf{GPT-4.1-mini}~\citep{gpt4p1_mini_050414}, \textbf{Gemini-2.5-flash-preview}~\citep{gemini_2.5flash_preview-0520}, and \textbf{Deepseek-R1}~\citep{deepseek_r1_20250528} \textbf{as Programmers}. Furthermore, to validate the proposed framework's capability in tackling highly sophisticated numerical algorithms in the simulation of flow past a cylinder, we employ the advanced \textbf{GPT-5-mini}~\citep{gpt5_mini_250807} as a dedicated Programmer. While other LLMs can hardly provide executable and bug-free code, GPT-5-mini exhibits superior reasoning capability under long-context scenarios, which are crucial for managing the complex implementation steps inherent in the immersed boundary method on a staggered grid. In contrast, we designate \textbf{GPT-4.1-mini} as both the Consultant and Reviewer due to its high response efficiency and 1-million-token context window, which is essential for managing historical code records or extensive execution logs.

The key innovation of the proposed agent lies in its collaborative framework structured upon the ``rewriting-resolution-review-revision'' logical chain. By establishing this pipeline, our framework presents two distinct advantages for autonomous scientific computing:
\begin{enumerate}[(i) ]
    \item \textbf{Proactive algorithmic planning:} The framework leverages the Consultant module to explicitly transform vague natural language into structured algorithmic formulations. This proactive domain-specific text augmentation provides explicit algorithmic guidance before code generation, thereby reducing reliance on massive trial-and-error parallel sampling to obtain a runnable initial baseline.

    \item \textbf{Ground-truth-free self-refinement:} The Reviewer module generates multi-dimensional diagnostic feedback on runtime, structure, and accuracy based on the problem context, generated code, and execution results. For the PDE solver tasks, reference solutions are reserved for external performance evaluation and excluded from the review–revision loop, allowing refinement without external ground-truth guidance.
\end{enumerate}

\section{Experiments}\label{sec:Experiments}

\textbf{Tasks.} In this section, we conduct a comprehensive performance evaluation of the proposed agent architecture across diverse specialized scenarios in scientific computing, with explicit focus on typical PDE benchmarks (including six specific equations), flow past a cylinder, ill-conditioned Hilbert linear systems, and data-driven physical analysis based on dimensional analysis. Tasks involving PDE solving range from typical textbook-style problems to an engineering-oriented case requiring the implementation of the IBM on staggered grids.

\textbf{Problem setup.} The task-specific workflows and agent responsibilities for the evaluated problems are defined as follows:
\begin{enumerate}[ (i)]
    \item \textbf{PDE-solving tasks:} For the PDEbench cases and flow past a cylinder problem, the agent is required to design numerical solvers from scratch. Specifically, the Programmer autonomously designs the solver algorithms guided by the domain-specific knowledge provided by the Consultant. Subsequently, the Reviewer monitors the execution results and generates structured comments to facilitate the Programmer's subsequent revision, covering aspects such as discretization schemes, time-stepping methods, and boundary condition treatments.
    \item \textbf{Hilbert linear systems:} The agent is required to explore various numerical methods and design a general solver from scratch capable of working across various matrix dimensions, rather than relying on a single pre-determined approach. In this scenario, the Reviewer provides feedback based on the computed $L^\infty$ relative error, guiding the Programmer to iteratively refine the algorithmic details specifically for handling ill-conditioned matrices.
    \item \textbf{Dimensional analysis problems:} The agent is required to infer the dimensions of physical variables and design a parameter search algorithm to identify dominant dimensionless quantities from an infinite set of candidate combinations. In this scenario, the Reviewer assesses constraints on the exponents, dimensional homogeneity, and the $R^2$ regression score to guide the systematic revision of the search algorithm's implementation.
\end{enumerate}

Specifically, 12 independent runs were conducted for each PDEbench and flow past a cylinder case, while 8 independent runs were completed for the Hilbert and dimensional analysis problems. Runs that failed to return an API response were excluded from the corresponding success-rate denominator, whereas returned code with execution or numerical failures was counted as unsuccessful. Here, ``independent'' denotes that the entire multi-agent pipeline (including the rewriting, resolution, review, and revision stages) was executed from scratch for the identical problem setup. Consequently, the variation in the final results across different runs arises entirely from the inherent stochasticity of the LLMs during API calls.

\textbf{Models.} We evaluate several representative and widely adopted LLMs as the Programmer, including GPT-4.1-mini, Gemini-2.5-flash-preview, and Deepseek-R1. Additionally, the advanced GPT-5-mini is employed as a dedicated Programmer to handle the flow past a cylinder task. For API calls across all modules (Programmer, Consultant, and Reviewer), the temperature is set to \texttt{temperature = 0} to minimize stochasticity and maximize code reproducibility, while other sampling parameters (such as \texttt{top\_p}) remain at their default values and no explicit random seeds are set. 

Furthermore, strict role control is implemented via the LangGraph architecture, where each module functions as an independent node rather than sharing a continuous chat history. The Reviewer module is completely decoupled from the code generation process, and only evaluates the explicit textual inputs (e.g., generated code and execution results) guided by an exclusive, role-specific system prompt template.

\textbf{Metrics.} The capabilities of generating high-quality code and solutions, compared to the original LLM without feedback from the Reviewer module, are measured by:

$\bullet$ \textbf{Code execution success rate.} The percentage of generating bug-free code and providing non-NaN solutions. This metric indicates the reliability of the code generation.

$\bullet$ \textbf{Solving success rate.} The percentage of errors below a specified threshold, particularly in the task of solving ill-conditioned Hilbert linear systems. For the data-driven physical analysis task, this refers to the percentage of successfully identifying dominant dimensionless quantities.

$\bullet$ \textbf{Accuracy.} The $L^2$-norm or $L^\infty$-norm relative error.

Finally, Section~\ref{sec:ablation_tokenct_} reports a per-stage ablation study and API costs across four sequential reasoning stages (baseline, \texttt{ans-0}, \texttt{rev-1}, and \texttt{rev-2}) to isolate the individual contributions of the Consultant and Reviewer modules, and further compares Re4 agent with existing LLM-based scientific computing frameworks in terms of workflow designs and problem configurations. \textcolor{black}{The complete task prompts are provided in Appendix~\ref{app:full_tasks}, while supplementary numerical results for the six PDE benchmarks are collected in Appendix~\ref{app:pdebench_details}. Section~\ref{sec:discussion} further analyzes representative solver-development trajectories across tasks, focusing on numerical decisions, implementation difficulties, and the checks needed to assess revised solvers.}

\subsection{Partial Differential Equation benchmark}\label{subsec: PDE bench}

\begin{figure}[!t]
\begin{center}
\footnotesize
\def~{\hphantom{0}}
\begin{tabular}{|c|c|c|}
\cline{1-3}
\multirow{2}{*}{PDE systems}&\multirow{2}{*}{Challenges}&\multirow{2}{*}{Reference solutions}\\
& & \\
\cline{1-3}

\makecell{\textbf{Burgers' equation}\\ $\frac{\partial u}{\partial t}+u u_x-\nu u_{xx}=0$} & \makecell{\textbf{Nonlinear,Unsteady \checkmark}\\Accurately resolve\\ oscillations and capture\\ shock waves}&
\begin{minipage}{0.25\textwidth}
    \centering
    \includegraphics[width=\linewidth]{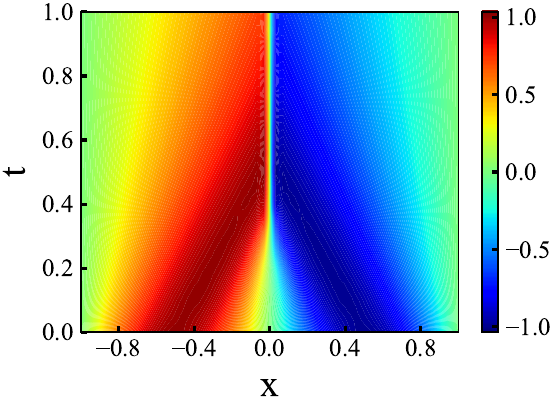}
  \end{minipage}\\
\cline{1-3}

\makecell{\textbf{Sod Shock Tube}\\ 
$\begin{cases}
\frac{\partial \rho}{\partial t} + \frac{\partial (\rho u)}{\partial x} = 0\\
\frac{\partial (\rho u)}{\partial t} + \frac{\partial (\rho u^2 + p)}{\partial x} = 0\\
\frac{\partial (\rho E)}{\partial t} + \frac{\partial (\rho E u + p u)}{\partial x} = 0
\end{cases}$}&\makecell{\textbf{Unsteady \checkmark} \\Precisely capture\\ time-dependent positions\\ and strengths of\\ multiple wave systems}&
\begin{minipage}{0.23\textwidth}
    \centering
    \includegraphics[width=\linewidth]{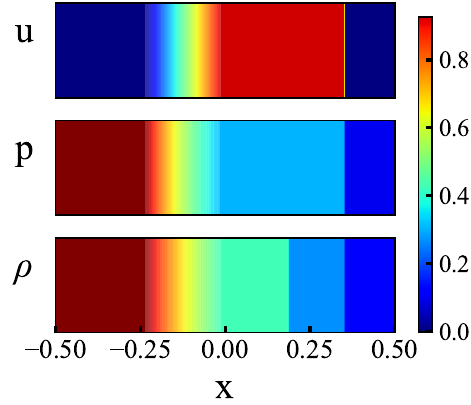}
  \end{minipage}\\
\cline{1-3}

\makecell{\textbf{Poisson equation}\\ $\Delta u = 0$}&\makecell{\textbf{Complex Geometry \checkmark}\\ Precise discretization of \\curvilinear circular holes \\and rectangular edges}&
\begin{minipage}{0.25\textwidth}
    \centering
    \includegraphics[width=\linewidth]{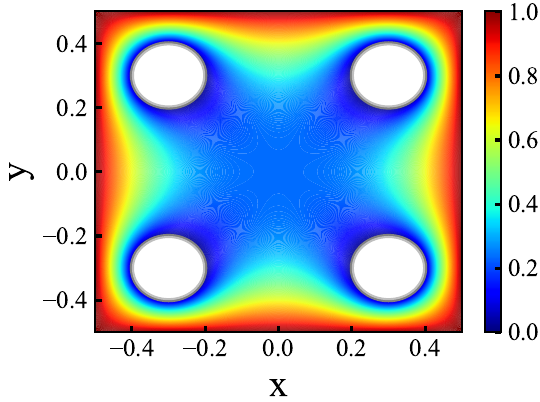}
  \end{minipage}\\
\cline{1-3}
 
\makecell{\textbf{Helmholtz equation}\\ $-\Delta u + k^2 u = f$}&\makecell{\textbf{Complex Geometry \checkmark}\\ Proper discretization of\\ complex boundaries\\ and handling of\\ high-frequency oscillation} &
\begin{minipage}{0.25\textwidth}
    \centering
    \includegraphics[width=\linewidth]{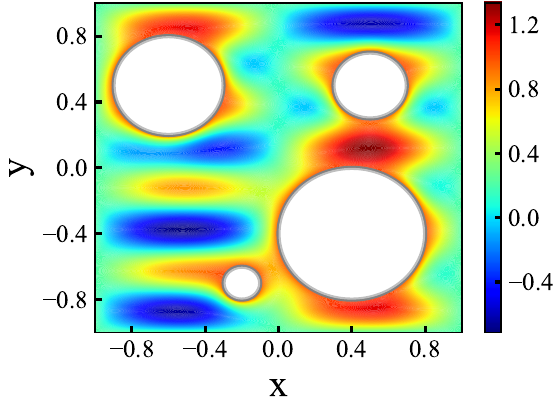}
  \end{minipage}\\
\cline{1-3}
 
\makecell{\textbf{Lid-Driven Cavity}\\ $\begin{cases}
\nabla \cdot \textbf{u}  = 0 \\
\textbf{u} \cdot \nabla \textbf{u} + \nabla p = \frac{1}{\mathrm{Re}} \Delta \textbf{u} \\
\end{cases}$}&\makecell{\textbf{Nonlinear \checkmark}\\ Complex flow details and\\ velocity-pressure coupling} &
\begin{minipage}{0.25\textwidth}
    \centering
    \includegraphics[width=\linewidth]{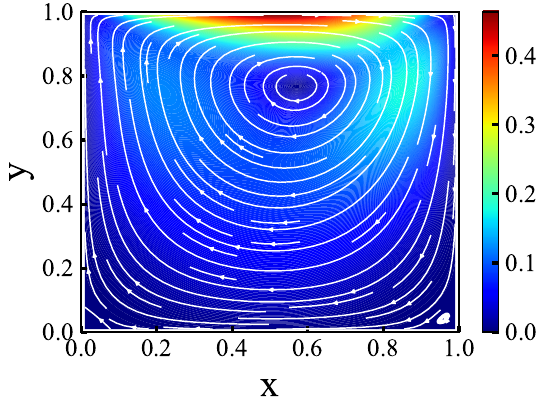}
  \end{minipage}\\
 \cline{1-3}
 
\makecell{\textbf{Unsteady NS equation}\\ $\begin{cases}
\nabla \cdot \textbf{u}  = 0 \\
\frac{\partial \mathbf{u}}{\partial t} + \mathbf{u} \cdot \nabla \textbf{u} + \nabla p - \frac{1}{\mathrm{Re}} \Delta \textbf{u} = \mathbf{f}\\
\end{cases}$}&\makecell{\textbf{Nonlinear,Unsteady \checkmark}\\ Temporal evolution and\\ velocity-pressure coupling} &
\begin{minipage}{0.25\textwidth}
    \centering
    \includegraphics[width=\linewidth]{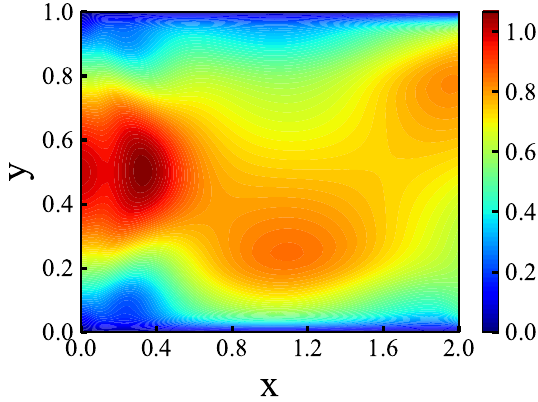}
  \end{minipage}\\
 \cline{1-3}

\end{tabular}
\caption{Summary of various equations in PDEbench.}
\label{tab:Pdebench}
\end{center}
\end{figure}

\begin{figure}[!t]
\centering
\begin{subfigure}[b]{0.6\textwidth}
\centering
{\includegraphics[width=\textwidth]{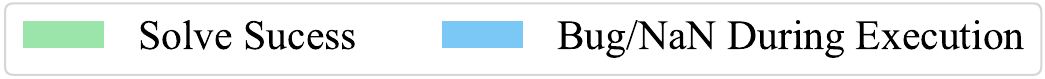}}
\end{subfigure}
\begin{subfigure}[b]{0.85\textwidth}
\centering
{\includegraphics[width=\textwidth]{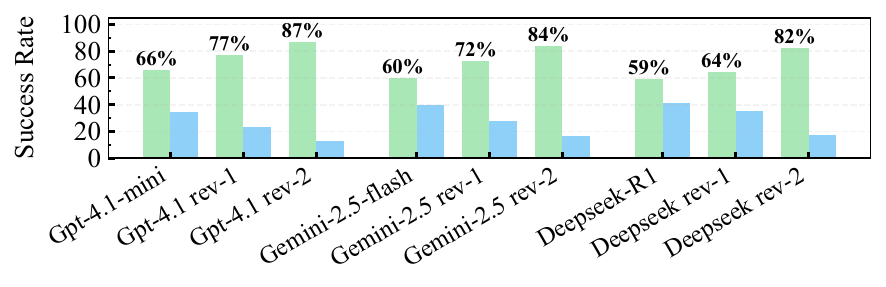}}
\end{subfigure}
  \caption{The overall average execution success rate of numerical algorithms employed by Programmers across all equations in the PDEbench.}
\label{fig:All_average_exe_rate}
\end{figure}

\begin{figure}[!t]
    \centering
\begin{subfigure}[b]{0.9\textwidth}
\centering
{\includegraphics[width = \textwidth]{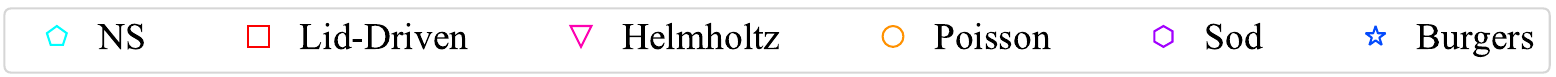}}
\end{subfigure}    
\begin{subfigure}[b]{0.85\textwidth}
\centering
{\includegraphics[width=\textwidth]{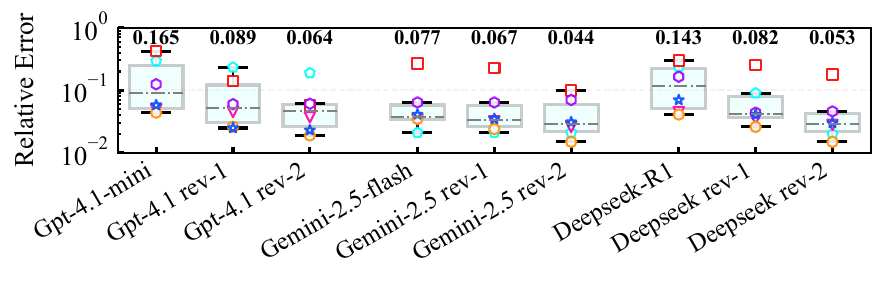}}
\end{subfigure}
    \caption{Boxplot of the average $L^2$ relative errors for all equations in the PDEbench. The \textbf{annotated numbers} in the figure denote the mean value of average relative errors.}
\label{fig:All_average_error_box}
\end{figure}

\begin{table}[!t]
\centering
\scriptsize
\renewcommand{\arraystretch}{1.2}
\setlength{\tabcolsep}{1.7pt}
\begin{tabular}{l|cccc|cccc|cccc}
\hline
\multirow{2}{*}{\textbf{PDE Systems}} & \multicolumn{4}{c|}{\textbf{GPT-4.1-mini}} & \multicolumn{4}{c|}{\textbf{Gemini-2.5-flash}} & \multicolumn{4}{c}{\textbf{Deepseek-R1}} \\ \cline{2-13}
& ans-0 & rev-1 & rev-2 & \textbf{Best} & ans-0 & rev-1 & rev-2 & \textbf{Best} & ans-0 & rev-1 & rev-2 & \textbf{Best} \\ \hline

Burgers & 5.8e-02 & 2.5e-02 & \underline{\textbf{2.3e-02}} & 1.98e-02 & 4.1e-02 & 3.5e-02 & \underline{3.1e-02} & \textbf{\textit{1.82e-02}} & 7.0e-02 & 4.1e-02 & \underline{3.0e-02} & 1.85e-02 \\

Sod Shock & 1.3e-01 & \underline{6.0e-02} & 6.1e-02 & 3.60e-02 & \underline{6.4e-02} & \underline{6.4e-02} & 7.0e-02 & 5.81e-02 & 1.7e-01 & \underline{\textbf{4.4e-02}} & 4.6e-02 & \textbf{\textit{2.31e-02}} \\

Poisson & 4.4e-02 & 2.6e-02 & \underline{1.9e-02} & 9.90e-03 & 3.5e-02 & 2.4e-02 & \underline{\textbf{1.5e-02}} & 1.18e-02 & 4.1e-02 & 2.6e-02 & \underline{\textbf{1.5e-02}} & \textbf{\textit{7.15e-03}} \\

Helmholtz & 4.9e-02 & 4.5e-02 & \underline{3.7e-02} & 2.30e-02 & 3.4e-02 & 3.2e-02 & \underline{\textbf{2.6e-02}} & 2.09e-02 & 4.5e-02 & 3.5e-02 & \underline{2.8e-02} & \textbf{\textit{1.47e-02}} \\

Lid-Driven & 4.2e-01 & 1.4e-01 & \underline{\textbf{5.7e-02}} & \textbf{\textit{2.74e-03}} & 2.7e-01 & 2.3e-01 & \underline{9.9e-02} & 6.55e-03 & 3.0e-01 & 2.6e-01 & \underline{1.8e-01} & 8.77e-03 \\

Unsteady NS & 2.9e-01 & 2.3e-01 & \underline{1.9e-01} & 1.89e-02 & \underline{2.1e-02} & \underline{2.1e-02} & \underline{2.1e-02} & 1.83e-02 & 2.4e-01 & 9.0e-02 & \underline{\textbf{2.0e-02}} & \textbf{\textit{1.81e-02}} \\ \hline

\end{tabular}
\caption{Summary of the relative $L^2$ errors across all cases in PDEbench. For each Programmer, \texttt{ans-0}, \texttt{rev-1}, and \texttt{rev-2} report the \textbf{average errors} across successful independent runs at each corresponding response stage. Underlined entries indicate a Programmer's minimum average error, while bold entries indicate the overall minimum average error among all Programmers. The \textbf{Best} columns present the minimum error achieved from the best-of-$N$ perspective at the final stage. The \textbf{\textit{bold and italic}} entries denote the overall best accuracy achieved for each equation.}
\label{tab:pdeb_summ_relerr}
\end{table}

In this subsection, we assess the performance of the proposed agent framework in designing algorithms to solve a set of PDE problems that represent a wide range of physical challenges. The benchmark encompasses three categories: \textbf{discontinuous problems} (Burgers' equation, Sod shock tube), \textbf{elliptic equations} (Poisson equation, Helmholtz equation), and \textbf{Navier-Stokes (NS) equations} (Lid-driven cavity, Unsteady NS).

The PDEs selected for this study encompass a diverse range of mathematical properties, ensuring that the benchmark does not favor a specific type of equation. The curated set of problems introduces several fundamental challenges, outlined as follows:
\begin{itemize}
    \item \textbf{Nonlinear Behavior:} When PDEs exhibit nonlinearity, minor deviations in initial conditions can lead to significant divergence in results.
    \item \textbf{Complex Geometry:} Irregular geometries pose direct challenges for the agent in accurately discretizing the domain and representing boundary behaviors.
    \item \textbf{Unsteady Evolution:} The solutions of certain PDEs change dynamically over time, requiring algorithms to accurately handle time-varying processes and robustly capture transient phenomena.
\end{itemize}

Figure~\ref{tab:Pdebench} provides a detailed overview of the selected PDEs, their inherent challenges, and the corresponding reference solutions. The reference datasets are collected from the Pinnacle benchmark~\citep{Hao_etal_arXiv_2023}. 
% The specific mathematical formulations and prompt templates for each equation are detailed in the subsequent subsections. 
To evaluate the deviation between the agent's answers and reference solutions, we adopt the relative $L^2$ error metric to rigorously quantify the global fidelity of the solutions, defined as follows:
\begin{equation}
error = \sqrt{\frac{\sum_{i=1}^n (y_i' - y_i)^2}{\sum_{i=1}^n (y_i)^2}}
\label{equ:rel2_err}
\end{equation}
where $\boldsymbol{y} = (y_i)_{i=1}^n$ is the ground truth and $\boldsymbol{y'} = (y_i')_{i=1}^n$ is the vector of computed values, and $n$ is the reference solution's grid point count.

The overall average execution success rate of the code generated by the agent framework across the six PDE benchmarks is presented in figure~\ref{fig:All_average_exe_rate}. Within the proposed Re4 framework, the initial Resolution stage is followed by an iterative Review--Revision loop. Consequently, comparing the initial response (\texttt{ans-0}) against the final output (\texttt{rev-2}) naturally serves as an ablation study to evaluate the contribution of the Reviewer module. It is evident that with the involvement of the Reviewer, the Programmers effectively perform self-debugging and self-refinement, leading to an increase of up to 24 percentage points in the overall code execution success rate.

Beyond merely achieving higher execution reliability, the framework also demonstrates a marked refinement in numerical accuracy. As evidenced by the stage-wise average error metrics and the best final-stage results (best-of-$N$ perspective) summarized in table~\ref{tab:pdeb_summ_relerr}, the overall $L^2$-norm relative-error statistics exhibit a clear downward trend for all three Programmers across the successive response stages. The subsequent case studies show that these improvements arise from revisions to specific numerical implementations rather than from syntax correction alone.

Moreover, the improvement in solution accuracy from a statistical perspective is further corroborated by the error distribution visualized in figure~\ref{fig:All_average_error_box}, which plots the distribution of the average $L^2$-norm relative errors aggregated across all evaluated cases. The boxplots show that, from \texttt{ans-0} to \texttt{rev-2}, the median error decreases for all three Programmers, while the interquartile range narrows for GPT-4.1-mini and DeepSeek-R1. The subsequent case-level analyses in Section~\ref{sec:discussion} further show that some of these high-error outcomes originate from numerically unreasonable or non-physical implementations rather than simple syntax failures.

\textcolor{black}{Beyond these aggregate statistics, the following case studies examine how the review--revision process changes specific numerical implementations for the Sod shock tube, lid-driven cavity, and unsteady NS benchmarks. These cases expose different difficulties involving reconstruction and time integration, convection stencils, boundary treatment, and pressure--velocity coupling.}

\subsubsection{\label{subapp:Sod_Shock}Sod shock tube}

\begin{figure}[!b]
\centering
\begin{tikzpicture}[scale=1.0, >=stealth]
    \definecolor{boxblue}{RGB}{210, 235, 255}
    \definecolor{lineblue}{RGB}{100, 149, 237}

    % 1. 绘制管壁 (Tube)
    \draw[thick] (0,0) rectangle (8,1.5);
    
    % 2. 隔膜 (Diaphragm)
    \draw[line width=2pt, lineblue] (4,0) -- (4,1.5);
    \node[draw=boxblue, fill=boxblue, rounded corners=2pt, font=\bfseries\footnotesize] at (4, 2.0) {Diaphragm ($x=0.5$)};

    % 3. 左右状态标注框 (State labels)
    \node[draw=boxblue, fill=boxblue, rounded corners=2pt, font=\bfseries\footnotesize] at (1, 2.0) {Left State};
    \node[draw=boxblue, fill=boxblue, rounded corners=2pt, font=\bfseries\footnotesize] at (7, 2.0) {Right State};

    % 4. 物理参数 (Physical properties)
    \node[align=left, font=\small] at (2, 0.75) {$\rho_L = 1.0$ \\ $u_L = 0.0$ \\ $p_L = 1.0$};
    \node[align=left, font=\small] at (6, 0.75) {$\rho_R = 0.125$ \\ $u_R = 0.0$ \\ $p_R = 0.1$};

    % 5. 尺寸线 (Dimension)
    \draw[<->, lineblue, thick] (0,-0.5) -- (8,-0.5) node[midway, below, black, font=\footnotesize] {Total length $L=1.0$};
\end{tikzpicture}
\caption{Schematic of the Sod shock tube with initial conditions at $t=0$.}
\label{Fig_schematic_sod_sb}
\end{figure}

\begin{figure}[!b]
\centering
\begin{subfigure}[b]{0.6\textwidth}
\centering
{\includegraphics[width=\textwidth]{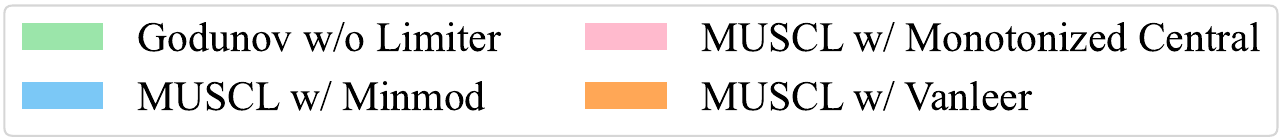}}
\end{subfigure}
\begin{subfigure}[b]{0.85\textwidth}
\centering
{\includegraphics[width=\textwidth]{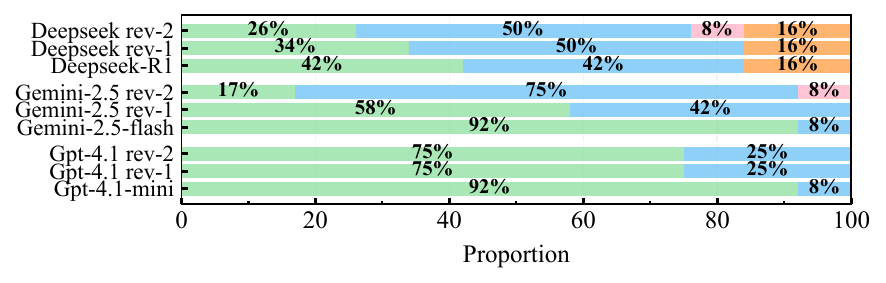}}
\end{subfigure}
  \caption{Proportional distribution of the spatial reconstruction schemes employed by Programmers. All Programmers selected the Harten-Lax-vanLeer-Contact (HLLC) Riemann solver during the sampling process.}
\label{fig:Sod_space_method}
\end{figure}

\begin{figure}[!ht]
\centering
\begin{subfigure}[b]{0.6\textwidth}
\centering
{\includegraphics[width=\textwidth]{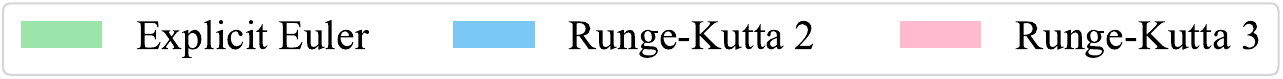}}
\end{subfigure}
\begin{subfigure}[b]{0.85\textwidth}
\centering
{\includegraphics[width=\textwidth]{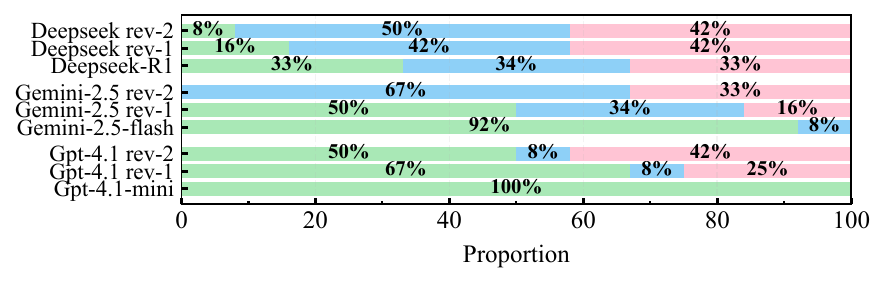}}
\end{subfigure}
  \caption{Proportional distribution of the time integration methods employed by Programmers.}
\label{fig:Sod_time_method}
\end{figure}

The Sod shock tube problem~\citep{sod1978survey} is a classic one-dimensional PDE system as depicted in figure~\ref{Fig_schematic_sod_sb}. The governing equations for this problem are the Euler equations simplified from the compressible NS equations, used to describe the wave systems in inviscid compressible flows.

At the initial time $t=0$, the left domain of the diaphragm is occupied by high-pressure gas, while the right domain contains low-pressure gas. Upon removal of the diaphragm at $t>0$, the high-pressure gas undergoes rapid expansion into the low-pressure region on the right, generating three distinct wave systems.

\begin{enumerate}[(i) ]
    \item a left-traveling rarefaction wave where pressure and density exhibit continuous variation across the wave front, with velocity gradually increasing to its peak.
    \item a right-traveling shock wave where pressure and density sharply increase across the shock wave front while velocity decreases abruptly.
    \item a contact between the two waves, where pressure and velocity remain continuous across the discontinuity surface, but density experiences an abrupt jump.
\end{enumerate}

{\color{black}To examine the numerical choices made during revision, the proportions of spatial reconstruction and time-integration schemes employed across different response stages are presented in figures~\ref{fig:Sod_space_method} and \ref{fig:Sod_time_method}. The distributions show increased adoption of MUSCL reconstruction with slope limiters and Runge--Kutta time integration during revision. These higher-order schemes can improve shock-capturing accuracy when implemented correctly. They also require consistent interface states, flux evaluations, and boundary updates at every time-integration stage; inconsistencies can produce incorrect fluxes or NaNs. Table~\ref{tab:pdeb_summ_relerr} shows that the mean relative $L^2$ errors increase slightly from \texttt{rev-1} to \texttt{rev-2} for all three Programmers. The best final-stage profiles reproduce the principal rarefaction wave, contact discontinuity, and shock wave of the Sod shock tube problem. Further run-level results are provided in Appendix figure~\ref{fig:Sod_detailed_results}.}

\subsubsection{\label{subapp:pde_Lid-Driven}Lid-driven Cavity}

\begin{figure}[!b]
\centering
\begin{tikzpicture}[scale=0.8, >=stealth]
    % 1. 定义颜色（保持全局统一）
    \definecolor{boxblue}{RGB}{210, 235, 255}
    \definecolor{lineblue}{RGB}{100, 149, 237}
    \definecolor{arroworange}{RGB}{255, 180, 130}

    % 2. 绘制三面静止壁面的斜纹填充 (Bottom, Left, Right)
    % Hatched Walls
    \fill[pattern=north east lines] (-0.3, -0.3) rectangle (5.3, 0); % Bottom
    \fill[pattern=north east lines] (-0.3, 0) rectangle (0, 5);      % Left
    \fill[pattern=north east lines] (5, 0) rectangle (5.3, 5);      % Right
    
    % 3. 绘制主边界线 (Top lid is a separate line)
    \draw[thick] (0, 5) -- (0, 0) -- (5, 0) -- (5, 5);
    \draw[ultra thick] (0, 5) -- (5, 5); % 顶盖加粗

    % 4. 绘制顶盖驱动箭头 (Moving Lid Arrow)
    \draw[->, arroworange, line width=2pt] (0, 5.2) -- (5, 5.2);
    \node[draw=boxblue, fill=boxblue, rounded corners=3pt, font=\bfseries\small] 
        at (2.5, 5.8) {Top wall (Moving lid)};

    % 5. 绘制静止壁面标注框 (Stationary Wall Labels)
    \node[draw=boxblue, fill=boxblue, rounded corners=3pt, font=\bfseries\small, rotate=90] 
        at (-1.0, 2.5) {Left wall (No-slip)};
    \node[draw=boxblue, fill=boxblue, rounded corners=3pt, font=\bfseries\small, rotate=90] 
        at (6.0, 2.5) {Right wall (No-slip)};
    \node[draw=boxblue, fill=boxblue, rounded corners=3pt, font=\bfseries\small] 
        at (2.5, -0.8) {Bottom wall (No-slip)};

    % 6. 尺寸线 (Dimension lines) - 使用蓝色 lineblue
    \draw[<->, lineblue, thick] (0,-1.6) -- (5,-1.6) node[midway, fill=white, inner sep=1pt, font=\small] {$L = 1$};
    \draw[<->, lineblue, thick] (-2.2, 0) -- (-2.2, 5) node[midway, fill=white, inner sep=1pt, font=\small] {$H = 1$};

    % 7. 内部流场示意 (可选：加一个旋转的小箭头表示涡流趋势)
    \draw[->, gray!40, line width=1pt] (3.5, 2.5) arc (0:-300:1.2);
    \node[gray!50, font=\footnotesize] at (2.5, 2.7) {Primary Vortex};

\end{tikzpicture}
\caption{Schematic of the computational setup for the lid-driven cavity flow.}
\label{fig:cavity_setup}
\end{figure}

Lid-driven cavity flow~\citep{ghia1982high} refers to a flow phenomenon in which a specified velocity is imposed on the top boundary of a closed cavity, whereas the other boundaries remain stationary. The governing equations are the incompressible steady NS equations, which describe the motion of the fluid inside the cavity under the influence of viscosity.

The inherent mathematical properties of these governing equations pose significant challenges for robust numerical solutions, primarily arising from the following two aspects:
\begin{enumerate}[(i)]
    \item \textbf{Pressure-Velocity Coupling:} Since the divergence-free continuity equation does not explicitly contain pressure while the momentum equations rely on pressure gradients, the pressure and velocity fields cannot be solved independently. This necessitates the use of special algorithms to effectively handle the coupling between pressure and velocity fields.
    \item \textbf{Nonlinear Convection:} The presence of nonlinear convection terms often leads to numerical instabilities, such as iterative divergence or slow convergence, making the solver design particularly demanding.
\end{enumerate}

\textcolor{black}{The computational setup is illustrated in figure~\ref{fig:cavity_setup}. The cavity results in table~\ref{tab:pdeb_summ_relerr} show lower mean errors after revision, while the best final-stage errors are much smaller than the corresponding means. This gap indicates that some executable runs remain inaccurate or insufficiently converged. The best final-stage solutions from all three Programmers reconstruct the primary vortex and agree closely with the reference velocity field. Additional run-level statistics and velocity fields are provided in Appendix~\ref{app:pdebench_details}.}

Beyond these quantitative results, the generated implementations highlight the importance of both convection discretization and array-index consistency for this steady incompressible-flow problem. The best final-stage implementations from all three Programmers employ the artificial compressibility method (ACM)~\citep{chorin1967numerical}. By introducing a pseudo-time derivative of pressure into the continuity equation, ACM advances pressure and velocity toward a steady incompressible solution without requiring a separate pressure-Poisson correction at each update. Compared with the projection method~\citep{chorin1968numerical}, this provides a relatively direct implementation route for the present benchmark. The collocated pressure--velocity arrangement used in these ACM implementations also avoids some of the indexing complexity associated with staggered grids.

\subsubsection{Unsteady NS equation}\label{appsub:pde_Unsteady_NS}

\begin{figure}[!b]
\centering
\begin{tikzpicture}[scale=0.8, >=stealth]
    % 1. 定义颜色（保持高度统一）
    \definecolor{boxblue}{RGB}{210, 235, 255}
    \definecolor{lineblue}{RGB}{100, 149, 237}
    \definecolor{arroworange}{RGB}{255, 180, 130}

    % 2. 绘制上下壁面的斜纹填充 (设定宽8, 高4)
    \fill[pattern=north east lines] (0, 4) rectangle (8, 4.3);
    \draw[thick] (0, 4) -- (8, 4);
    \fill[pattern=north east lines] (0, -0.3) rectangle (8, 0);
    \draw[thick] (0, 0) -- (8, 0);
    
    % 3. 绘制侧边边界线
    \draw[thick] (0, 0) -- (0, 4);
    \draw[thick] (8, 0) -- (8, 4);

    % 4. 内部强迫项标注 (居中)
    \node[draw=gray, dashed, fill=gray!5, rounded corners=3pt, align=center, font=\small] at (3.4, 2) 
        {\textbf{Internal Forcing $f(x,y,t)$} \\ $f_x = 0$ \\ $f_y = -\sin(\pi x)\sin(\pi y)\sin(\pi t)$};

    % 5. 边界条件标注框 (位置微调使其更紧凑)
    \node[draw=boxblue, fill=boxblue, rounded corners=3pt, font=\bfseries\small] 
        at (4, 4.8) {Top wall (No-slip)};
    \node[draw=boxblue, fill=boxblue, rounded corners=3pt, font=\bfseries\small] 
        at (4, -0.8) {Bottom wall (No-slip)};
    
    \node[draw=boxblue, fill=boxblue, rounded corners=3pt, font=\bfseries\small, align=center] 
        at (-2.7, 2) {Inlet\\ $u = u(0,y,t)$ \\ $v=0$};
    \node[draw=boxblue, fill=boxblue, rounded corners=3pt, font=\bfseries\small, align=center] 
        at (10.5, 2) {Outlet\\ $\partial u/\partial x = 0$ \\ $\partial v/\partial x = 0$ \\ $p=0$};

    % 6. 尺寸线 (Dimension lines)
    \draw[<->, lineblue, thick] (0,0.5) -- (8,0.5) node[midway, fill=white, inner sep=1pt, font=\small] {$L = 2$};
    \draw[<->, lineblue, thick] (7.2,0) -- (7.2,4) node[midway, fill=white, inner sep=1pt, font=\small] {$H = 1$};

    % 7. 入口箭头
    \foreach \y in {1, 2, 3} {
        \draw[->, arroworange, line width=1.5pt] (-1.2, \y) -- (-0.2, \y);
    }
    \foreach \y in {1, 2, 3} {
        \draw[->, arroworange, line width=1.5pt] (8.2, \y) -- (9.2, \y);
    }
    
\end{tikzpicture}
\caption{Schematic of the computational setup for the unsteady NS equations with explicit internal forcing at $Re=100$.}
\label{fig:unsteady_ns_setup}
\end{figure}

\begin{figure}[!t]
    \centering
\begin{subfigure}[b]{0.95\textwidth}
    \centering
{\includegraphics[width=\textwidth]{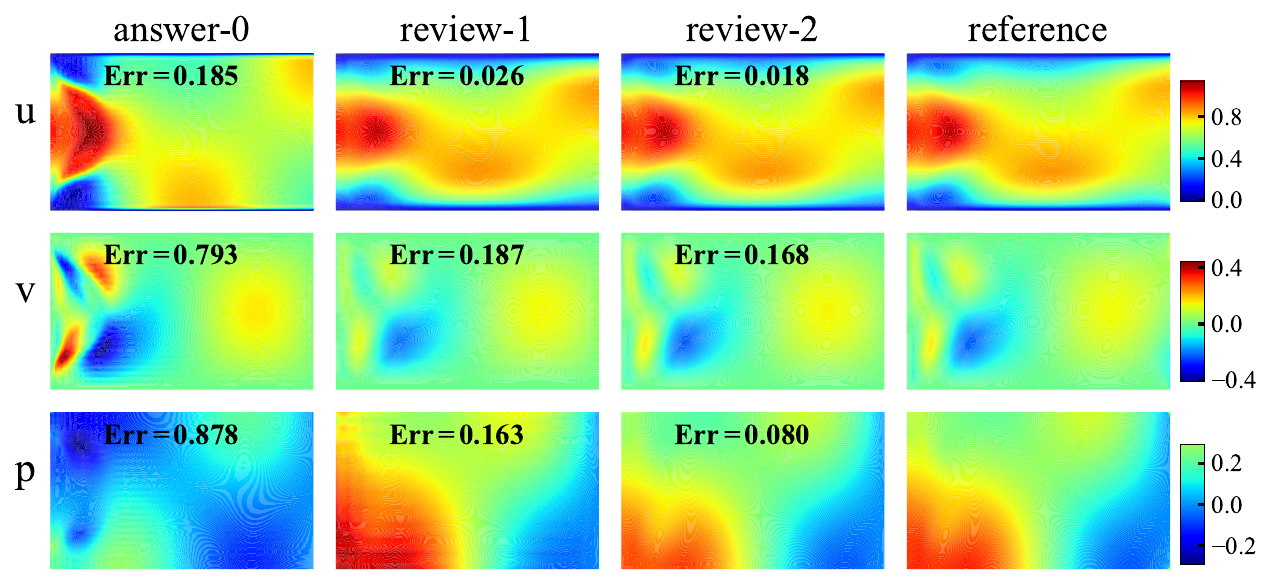}}
\end{subfigure}
\caption{The runtime outputs of the executable code provided by Programmer Deepseek-R1 in the initial response (\texttt{ans-0}), together with those in the Reviewer's first (\texttt{rev-1}) and second (\texttt{rev-2}) interventions, respectively.}
\label{fig:Unsteady_NS_R1_improvement}
\end{figure}

The unsteady incompressible NS equation serves as a spatiotemporal PDE system used to describe the transient dynamics of viscous flows. In this specific case, the fluid motion within the channel is jointly driven by a time-varying oscillatory inlet velocity and a spatiotemporal internal source term. This complex dynamic setup necessitates the accurate capture of evolving flow structures over the prescribed simulation interval. The inherent mathematical properties of these governing equations pose significant challenges for robust numerical solutions, primarily arising from the following aspects:
\begin{enumerate}[(i) ]
    \item \textbf{Coupled System:} The equations are a coupled system of PDEs, including the momentum equation and the continuity equation, which require simultaneous solution of the velocity $(u, v)$ and pressure $(p)$.
    \item \textbf{Unsteady Evolution:} The equations are unsteady in time, necessitating the handling of time-varying source terms and left inlet velocity boundary conditions, both of which are time-dependent, as well as the consideration of time integration stability.
    \item \textbf{Numerical Complexity:} The generated projection solvers typically employ the generalized minimal residual (GMRES) iterative method~\citep{saad1986gmres} for the pressure Poisson equation, presenting numerical challenges from the ill-conditioned discrete linear system and the computational cost of achieving convergence, both of which arise from the global continuity constraint in incompressible NS simulations.
\end{enumerate}

To overcome these inherent challenges, the agent needs to autonomously select appropriate spatiotemporal discretization schemes, rigorously enforce time-varying boundary conditions, and adopt efficient iterative solvers. The computational setup is illustrated in figure~\ref{fig:unsteady_ns_setup}. \textcolor{black}{Execution success reaches $63\%$ at \texttt{rev-2} for all three Programmers. Table~\ref{tab:pdeb_summ_relerr} shows that GPT-4.1-mini's final-stage mean relative $L^2$ error remains higher than those of the other two Programmers.}

By contrast, DeepSeek-R1 shows a more pronounced reduction in error across the response stages and provides a representative example of stage-wise refinement. Figure~\ref{fig:Unsteady_NS_R1_improvement} illustrates the evolution of the streamwise velocity $u$, normal velocity $v$, and pressure $p$. The initial response shows large deviations from the reference solution in all three quantities, whereas the final revision reproduces the streamwise velocity $u$ and pressure $p$ fields substantially more accurately. \textcolor{black}{Further comparisons across Programmers are given in Appendix~\ref{app:pdebench_details}.}

\textcolor{black}{Beyond the quantitative results, the inspected implementations highlight the importance of consistent boundary discretization and pressure correction in time-dependent incompressible-flow simulations. In the inspected DeepSeek-R1 implementation, the initial code computes spatial derivatives and pressure gradients using \texttt{np.roll}, so the difference formulas at array boundaries include values from the opposite edge. On this non-periodic domain, this introduces artificial coupling between opposite boundaries and can introduce errors into the computed divergence and pressure gradients. Subsequent revisions use central differences in the interior and one-sided differences at the boundaries, and add incomplete LU (ILU) preconditioning to the GMRES pressure solve. These changes address distinct numerical issues: the local stencils remove the artificial boundary coupling, while ILU preconditioning is used to accelerate the GMRES solve.}

\textcolor{black}{The inspected final-stage implementations differ in their treatment of pressure boundaries and boundary derivatives. GPT-4.1-mini approximates zero normal pressure gradients by copying adjacent pressure values, while Gemini-2.5-flash incorporates ghost-point Neumann conditions into the pressure matrix. DeepSeek-R1 applies second-order one-sided differences when evaluating divergence and pressure gradients at the boundaries. In a projection-based solver, convergence of the pressure solve alone does not guarantee a divergence-free corrected velocity unless the discrete pressure operator, divergence and gradient operators, and their boundary treatments are compatible. In the best-of-$N$ final-stage comparison, DeepSeek-R1 produces a pressure field closer to the reference solution than that produced by GPT-4.1-mini. The observed performance difference therefore cannot be attributed to boundary discretization alone.}

\subsection{Flow past a cylinder}\label{sec:clyinder_case}

\begin{figure}[!ht]
\centering
\begin{tikzpicture}[scale=0.8, >=stealth]
    % 定义颜色
    \definecolor{boxblue}{RGB}{210, 235, 255}
    \definecolor{lineblue}{RGB}{100, 149, 237}
    \definecolor{arroworange}{RGB}{255, 180, 130}

    % 1. 绘制上下壁面的斜纹填充 (Hatched Walls)
    \fill[pattern=north east lines] (-3, 3) rectangle (8, 3.3);
    \draw[thick] (-3, 3) -- (8, 3);
    \fill[pattern=north east lines] (-3, -3) rectangle (8, -3.3);
    \draw[thick] (-3, -3) -- (8, -3);

    % 2. 绘制侧边边界线
    \draw[thick] (-3, -3) -- (-3, 3);
    \draw[thick] (8, -3) -- (8, 3);

    % 3. 绘制圆柱及其斜纹填充
    \begin{scope}
        \clip (0,0) circle (0.5);
        \fill[pattern=north east lines] (-0.5,-0.5) rectangle (0.5,0.5);
    \end{scope}
    \draw[thick] (0,0) circle (0.5);
    \node at (0.8, 0.7) {\small No-slip};

    % 4. 绘制蓝色标注框 (Labels in blue boxes)
    \node[draw=boxblue, fill=boxblue, rounded corners=3pt, font=\bfseries\small] at (2.5, 3.8) {Top wall (Free-slip)};
    \node[draw=boxblue, fill=boxblue, rounded corners=3pt, font=\bfseries\small] at (2.5, -3.8) {Bottom wall (Free-slip)};
    
    \node[draw=boxblue, fill=boxblue, rounded corners=3pt, font=\bfseries\small, align=center, rotate=0] 
        at (-5.5, 0) {Inlet\\(u,v)=(1,0)};
    \node[draw=boxblue, fill=boxblue, rounded corners=3pt, font=\bfseries\small, align=center, rotate=0] 
        at (10, 0) {Outlet\\$\frac{\partial p}{\partial x}$=0};

    % 5. 绘制尺寸线 (Blue dimension lines)
    % x方向
    \draw[<->, lineblue, thick] (-3, 0) -- (0, 0) node[midway, below, black] {$x_1 = 3$};
    \draw[<->, lineblue, thick] (0, 0) -- (8, 0) node[midway, above, black] {$x_2 = 8$};
    % y方向
    \draw[<->, lineblue, thick] (0, 0) -- (0, 3) node[midway, right, black] {$y_2 = 3$};
    \draw[<->, lineblue, thick] (0, 0) -- (0, -3) node[midway, right, black] {$y_1 = 3$};
    % 直径标注 (Callout style)
    \draw[lineblue, thick] (-0.35, 0.35) -- (-1.2, 1.2) -- (-1.5, 1.2) node[left, black] {$d = 1$};

    % 6. 绘制入口/出口箭头 (Inlet/Outlet Arrows)
    \foreach \y in {-1.5, 0, 1.5} {
        \draw[->, orange!80, line width=1.5pt] (-4, \y) -- (-3.2, \y);
        \draw[->, orange!80, line width=1.5pt] (8.2, \y) -- (9, \y);
    }

    % 中心标记
    \draw[fill=white] (0,0) circle (0.05);

\end{tikzpicture}
\caption{Schematic of the computational domain and boundary conditions for the flow past a cylinder case at $Re=100$.}
\label{fig:cylinder_geometry}
\end{figure}

The 2-D incompressible flow past a circular cylinder at $Re=100$ represents a more realistic engineering scenario compared to the previous textbook-style PDE benchmarks~\citep{schafer1996benchmark}. Compared to the preceding rectangular-domain benchmarks, simulating flows over complex immersed geometries requires specialized interface treatments at the solid–fluid boundary to guarantee physical accuracy and numerical stability. To prevent the agent from arbitrarily adopting inappropriate numerical strategies, the prompt template for problem formulation explicitly mandates the autonomous implementation of the immersed boundary method using volume penalization on a staggered marker-and-cell grid. The computational setup is illustrated in figure~\ref{fig:cylinder_geometry}. Correctly and reliably solving this case poses challenges for the Programmer module in multiple technical implementation dimensions:

\begin{enumerate}[ (i)]
    \item handling the interpolation and cross-indexing logic unique to staggered grids, thereby correctly assembling the discretized momentum and pressure Poisson equations.
    \item implicitly embedding the volume penalization term into the diagonal entries of the momentum matrix to circumvent restrictive time-step limitation associated with explicit penalization.
    \item properly enforcing the no-slip boundary condition on the cylinder surface to support numerical stability and physically consistent solid--fluid coupling, while maintaining consistency with the pressure-projection step.
\end{enumerate}

Most LLMs struggle to generate executable, bug-free solvers that adhere to the aforementioned algorithmic requirements, as the task involves intricate matrix assembly formulations and indexing logic, requiring the agent to leverage domain-specific algorithmic knowledge to complete extensive long-context reasoning. Given its superior logical reasoning stability in long-context scenarios, \textbf{GPT-5-mini} is employed as the dedicated Programmer to handle such high-density algorithmic logic and implement the solver from scratch.

\begin{table}[!t]
\centering
\caption{Ablation study and cost analysis on the flow past a cylinder task over multiple independent runs, detailing the execution success rates, average input and output token consumption (in thousands, k), and average Programmer response times (in seconds, s) at each reasoning stage. The arrows ($\uparrow$) denote the individual contributions of the Consultant and Reviewer modules to the execution success rate.}
\label{tab:cylinder_token_rate}
\footnotesize
\renewcommand{\arraystretch}{1.2}
\begin{tabular}{l|l|l|cc|c}
\toprule
\multirow{2}{*}{\textbf{Programmer}} & \multirow{2}{*}{\textbf{Stage / Configuration}} & \multirow{2}{*}{\textbf{Success Rate}} & \multicolumn{2}{c|}{\textbf{Tokens (k)}} & \multirow{2}{*}{\textbf{Time (s)}} \\ \cmidrule(lr){4-5}
&  &  & \textbf{in} & \textbf{out} & \\ \midrule
\multirow{4}{*}{GPT-5-mini} & baseline (Programmer-Only) & 0\% & 0.94 & 5.88 & 36.47 \\ \cline{2-6}
 & ans-0 (with Consultant) & 9.1\% ~($\uparrow$ 9.1\%) & 4.02 & 6.82 & 51.51 \\
 & rev-1 (with Reviewer) & 45.5\% ~($\uparrow$ 36.4\%) & 8.15 & 7.27 & 48.62 \\
 & rev-2 (with Reviewer) & 54.5\% ~($\uparrow$ 45.4\%) & 8.68 & 7.93 & 48.87 \\ \bottomrule
\end{tabular}
\end{table}

\begin{figure}[!t]
\centering
\begin{subfigure}[b]{0.95\textwidth}
\centering
{\includegraphics[width=\textwidth]{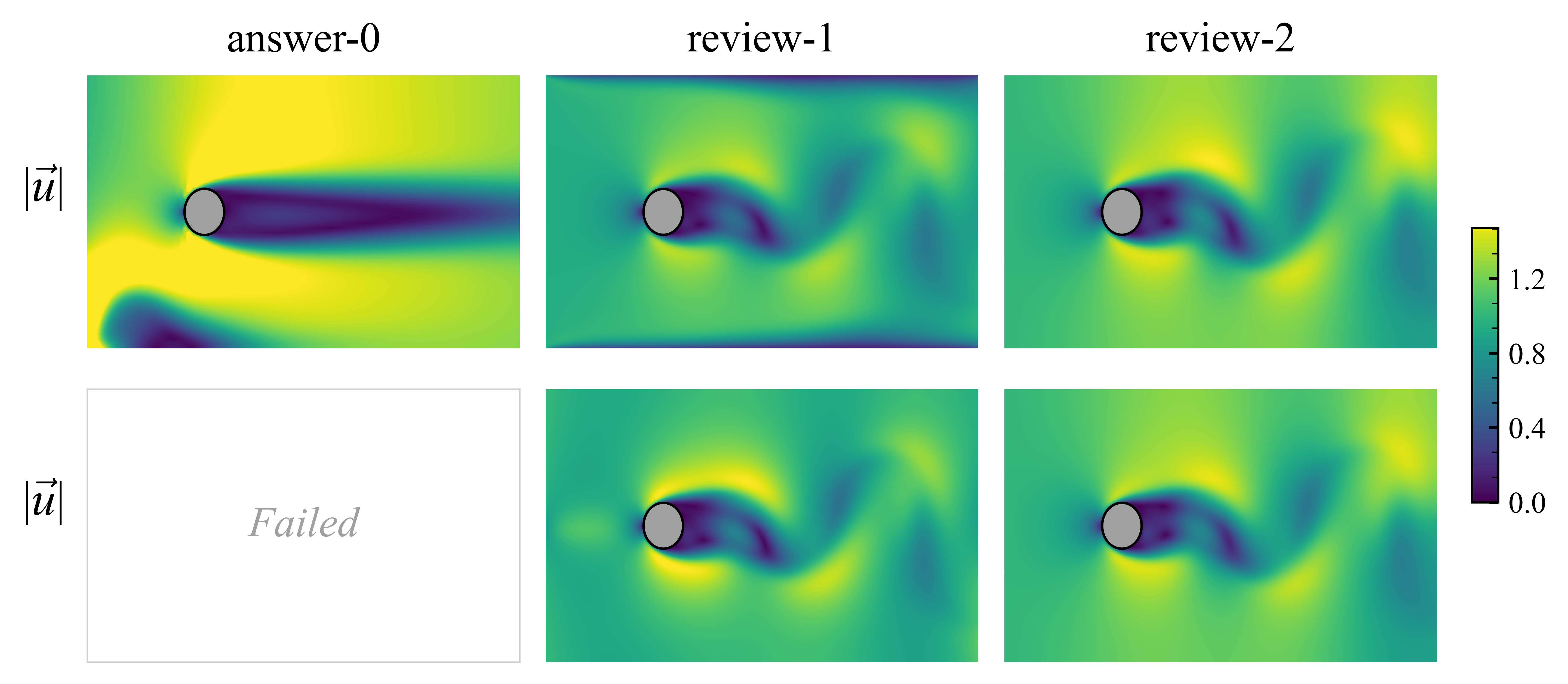}}
\end{subfigure}
\caption{Stage-wise outcomes of two independent cylinder-flow simulations generated by Programmer GPT-5-mini. The panels show the velocity magnitude $|\vec{u}|$ at \(t = 20.0\) when execution succeeds, while ``Failed'' indicates that no numerical solution was produced. The columns correspond to the initial response (\texttt{ans-0}) and two successive review–revision stages (\texttt{rev-1} and \texttt{rev-2}).}
\label{fig:cylinder_opt_steps}
\end{figure}

\begin{figure}[!t]
\centering
\begin{subfigure}[b]{0.95\textwidth}
\centering
{\includegraphics[width=\textwidth]{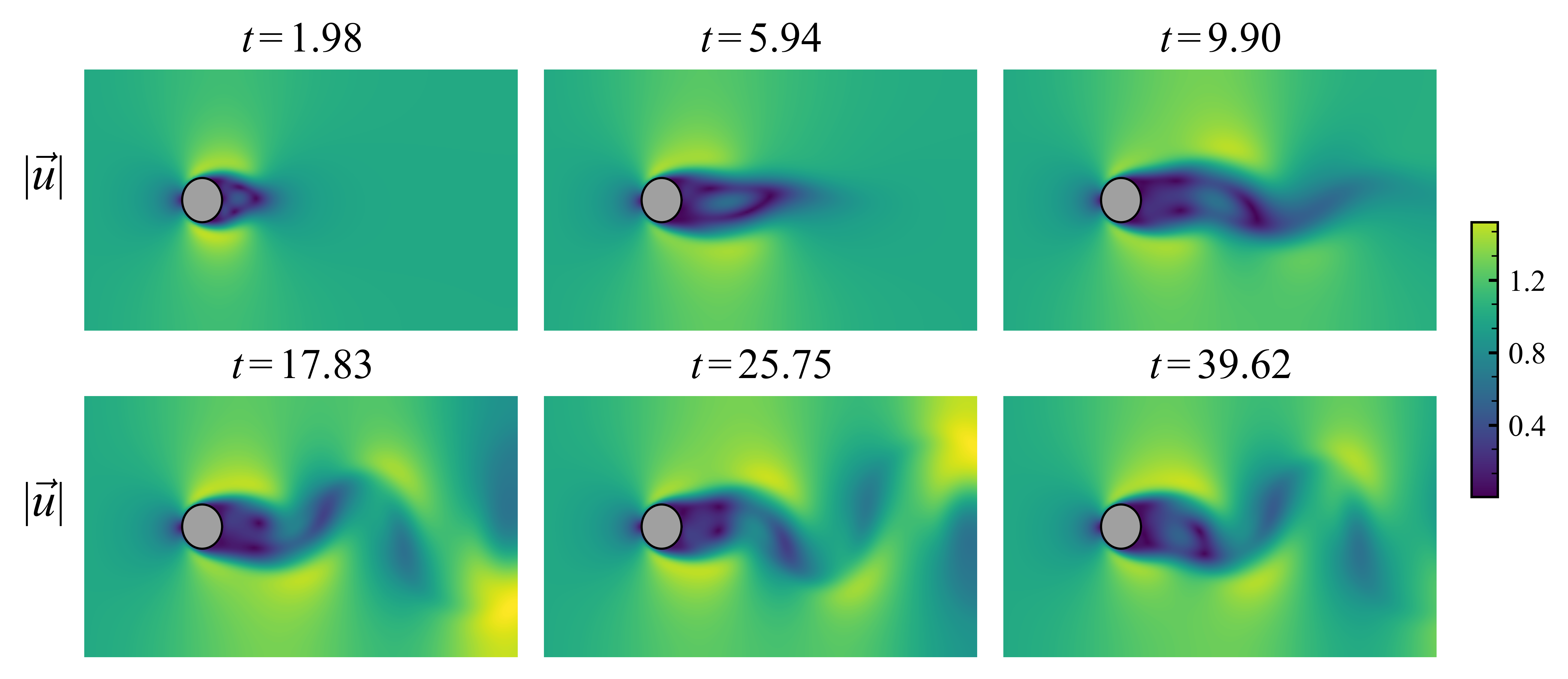}}
\end{subfigure}
\caption{Visualization of the periodic vortex shedding process at various time instants, generated by the final refined code (\texttt{rev-2}) from Programmer GPT-5-mini.}
\label{fig:cylinder_vor_sheding}
\end{figure}

Table~\ref{tab:cylinder_token_rate} presents an ablation study and a cost analysis for the flow past a cylinder task, aggregated across all independent runs. As observed, the baseline configuration (Programmer-Only) completely fails to generate executable code for this complex task ($0\%$ success rate). Introducing the Consultant's domain-specific knowledge increases the execution success rate to $9.1\%$ at stage \texttt{ans-0}. Subsequently, two rounds of Reviewer intervention increase this rate to $54.5\%$ at stage \texttt{rev-2}. Compared with the preceding PDEbench cases, the review--revision process for this task involves higher token consumption, reflecting the longer solver implementations and more extensive review context required by its numerical complexity. Nevertheless, GPT-5-mini's relatively high output-token throughput helps keep the average Programmer response time at approximately 49--52 seconds across the three response stages, as further discussed in Section~\ref{subsec:ablation_APIcost_}. These results demonstrate that the review mechanism improves the code execution success rate for this complex task.

Specifically, figure~\ref{fig:cylinder_opt_steps} presents the progressive refinement of the velocity fields across two representative independent runs. The upper row shows an initially executable solver whose intermediate response exhibits low-speed bands near the upper and lower boundaries. The \texttt{rev-2} implementation adjusts the boundary discretization and reapplies the prescribed free-slip conditions after pressure projection; these bands are no longer evident in the final plotted field. The lower row illustrates recovery from an initial broadcasting error caused by inconsistent cylinder-mask dimensions on the MAC grid. Corrections to mask construction and array--matrix indexing make the solver executable. The two revised implementations also differ in their pressure correction and cylinder-boundary treatment: the \texttt{rev-1} code explicitly zeroes velocities inside the cylinder after projection, whereas the \texttt{rev-2} code omits this reset and retains implicit volume penalization. \textcolor{black}{Since overwriting velocities after pressure projection can violate the discrete incompressibility constraint, the free-slip and cylinder-interior treatments need to remain consistent with the pressure correction.}

Unlike textbook-style PDEbench cases that rely on strict quantitative error evaluations against reference datasets, this engineering-oriented case serves as a qualitative capability demonstration. The primary objective is to evaluate the agent's ability to autonomously implement complex numerical algorithms and yield physically reasonable solutions. As shown in figure~\ref{fig:cylinder_vor_sheding}, time-resolved snapshots of the velocity magnitude $|\vec{u}|$ generated by the final revised code capture the development of an unsteady wake consistent with von K\'{a}rm\'{a}n vortex shedding.

\subsection{Hilbert linear algebraic systems}\label{subsec: Hilbert mat}

\begin{figure}[!t]
\centering
\begin{subfigure}[b]{0.65\textwidth}
\centering
{\includegraphics[width=\textwidth]{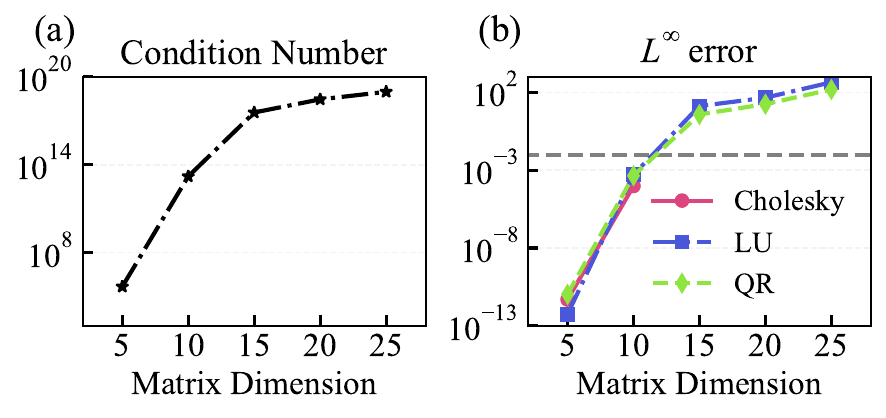}}
\end{subfigure}
  \caption{Schematic diagram of (a) variation of the 2-norm condition number with matrix dimension; (b) $L^\infty$ error obtained by direct solution using the naive methods.}
\label{fig:hil_naive_cond}
\end{figure}

In this subsection, we assess the performance of the proposed agent framework in solving the Hilbert linear algebraic system~\citep{kress1998ill}. The specific linear algebraic system involving Hilbert matrix is given by:
\begin{equation}
\label{Hilbert_equation}
\Large
\mathbf{b} = H_n \mathbf{x^*} = \begin{bmatrix}
1 & \frac{1}{2} & \cdots & \frac{1}{n} \\
\frac{1}{2} & \frac{1}{3} & \cdots & \frac{1}{n+1} \\
\vdots & \vdots & \ddots & \vdots \\
\frac{1}{n} & \frac{1}{n+1} & \cdots & \frac{1}{2n-1}
\end{bmatrix}
\normalsize
\begin{bmatrix}
1\\
1\\
\vdots\\
1
\end{bmatrix}
\end{equation}
where \( H_n \) is an \( n \times n \) Hilbert matrix with elements \( h_{i,j} = \frac{1}{i+j-1} \), and the vector \( \mathbf{b} \) is obtained by substituting the exact solution \( \mathbf{x}^* = (1, 1, \dots, 1)^{\top} \) into \( H_n \mathbf{x}^*\). The Hilbert matrix serves as a prototypical example of ill-conditioned matrices, whose 2-norm condition number $\kappa_2(H_n)$ grows exponentially with the dimension $n$, as shown in figure~\ref{fig:hil_naive_cond}.
The extremely high condition number renders naive methods ineffective, that is, even minor perturbations in input data could lead to drastic errors in the computed solution, thereby posing significant challenges for solving this ill-conditioned matrix. As reported by Wang et al.~\citep{Wang_etal_TAML_2025}, when directly solving this problem, single reasoning LLMs (e.g., DeepSeek R1 or o3-mini-high) fail to provide reliable solutions for this ill-conditioned linear system, even at a moderate dimension of $n=15$.

\begin{figure}[!t]
\centering
\begin{subfigure}[b]{0.8\textwidth}
\centering
{\includegraphics[width=\textwidth]{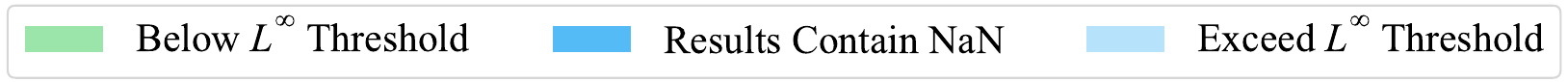}}
\end{subfigure}
\begin{subfigure}[b]{0.85\textwidth}
\centering
{\includegraphics[width=\textwidth]{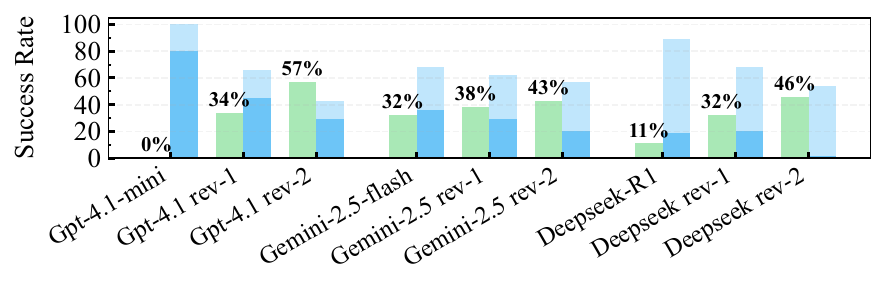}}
\end{subfigure}
  \caption{Proportional distribution of executable code provided by Programmers across three different completion statuses. The total number of evaluated method implementations at each response stage (\texttt{ans-0}, \texttt{rev-1}, \texttt{rev-2}) is as follows: GPT-4.1-mini (20, 30, 35), Gemini-2.5-flash (28, 34, 35), and Deepseek-R1 (27, 35, 46), respectively.}
\label{fig:hil_complete_rate}
\end{figure}

To bypass standard high-level built-in solvers (e.g., \texttt{numpy.linalg.solve}) and assess the reliability of the autonomously generated solving algorithms, the problem description explicitly required the agent to \textbf{implement various appropriate methods from scratch}. This setup evaluates both algorithm implementation and the selection of methods suited to severe ill-conditioning. It also allows us to trace how different solving strategies contribute to the solving success rate during review and revision. 

For a systematic evaluation, the tested matrix dimensions span $n \in \{5, 10, \dots, 25\}$, utilizing the $L^\infty$ error against the exact solution $\mathbf{x}^* = (1, \dots, 1)^\top$ as the quantitative metric. The statistics summarize method implementations pooled across eight independent runs. We categorize the completion statuses of the evaluated method implementations into three types: \textbf{results containing NaNs, errors exceeding the $L^\infty$ threshold, and errors below the $L^\infty$ threshold}, with the threshold set to $10^{-2}$.

The overall solving success rates across different response stages are shown in figure~\ref{fig:hil_complete_rate}. Notably, GPT-4.1-mini failed to provide any solution with errors below the threshold in its initial response, giving an initial solving success rate of $0\%$. Subsequently, with Reviewer intervention, the proportion of method implementations meeting the error threshold increases for all three Programmers. Although these results represent an improvement, the final solving success rates remain moderate, ranging from $43\%$ to $57\%$ after two revisions, under the $L^\infty$ error threshold of $10^{-2}$ and matrix dimensions up to $n=25$. 
Beyond the inherent difficulty of the problem itself, it is worth noting that this moderate performance is related to the evaluation metrics. 
Relaxing the threshold (e.g., to $10^{-1}$) or limiting the evaluation to lower dimensions (e.g., $n \le 15$) would make the success criterion less demanding. Therefore, the current evaluation scheme was adopted to avoid exaggerating the agent's performance on extremely ill-conditioned systems.

\textcolor{black}{Naive factorization methods predominantly produce NaNs or errors exceeding the threshold, whereas most below-threshold results are obtained using regularization or preconditioned conjugate-gradient methods. Regularized Cholesky implementations appear among both successful and unsuccessful results, indicating that regularization alone does not ensure the required accuracy. From the best-of-$N$ perspective, the final-stage results from all three Programmers meet the $L^\infty$ error threshold across the tested matrix sizes up to $n=25$. These results demonstrate that the framework can generate accurate solvers for the tested ill-conditioned Hilbert systems. Detailed method distributions and final-stage errors are provided in Appendix~\ref{app:hilbert_details}.}

\subsection{Data-driven physical analysis based on dimensional analysis}\label{subsec: Dimensionalanal}

\begin{figure}[!b]
\centering
\begin{subfigure}[b]{0.85\textwidth}
\centering
{\includegraphics[width=\textwidth]{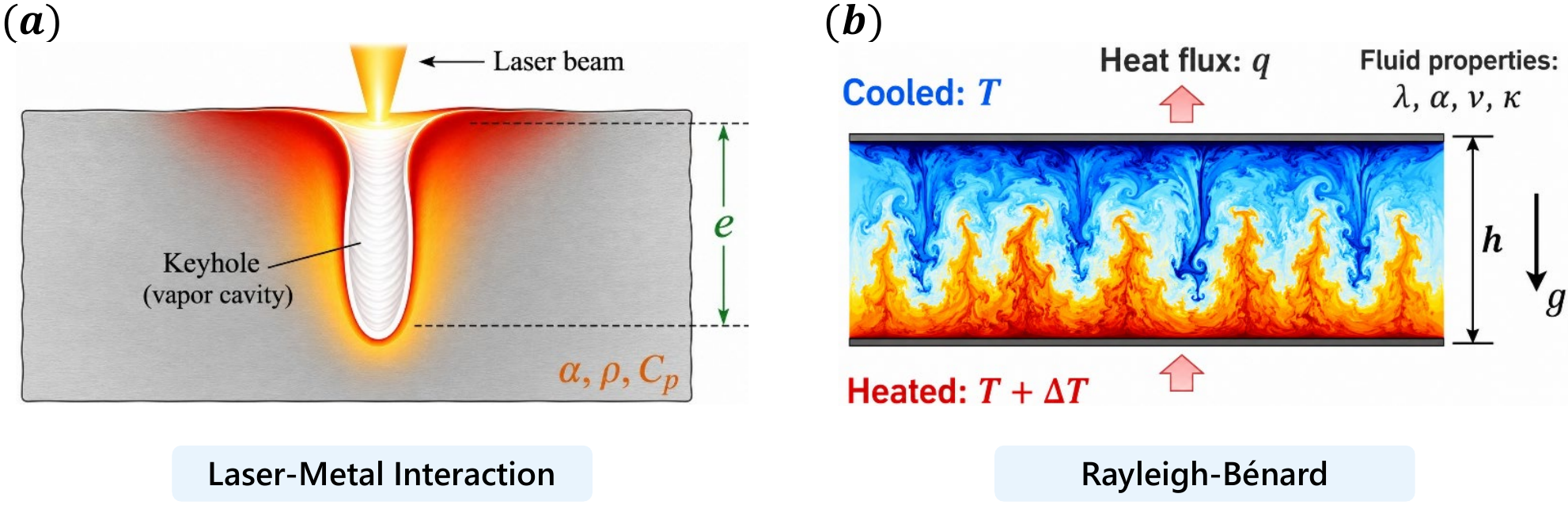}}
\end{subfigure}
  \caption{Schematic diagram of (a) vapor depression dynamics in laser–metal interaction; (b) turbulent Rayleigh–B\'enard convection.}
\label{fig:dimcase_schem}
\end{figure}

In this subsection, we assess the performance of the proposed agent framework in the data-driven physical analysis based on the dimensional analysis. As a data-driven physical analysis problem, distinct from solving specific governing equations, this task requires the agent to design a parameter search algorithm capable of identifying dominant dimensionless quantities from an infinite space of candidate combinations. In this scenario, the Reviewer assesses constraints on the exponents, dimensional homogeneity, and the $R^2$ regression score to steer the systematic revision of the search algorithm’s implementation. Subsequently, two distinct physical phenomena: keyhole dynamics in laser-metal interaction and Rayleigh-B\'enard convection in fluid mechanics are selected to demonstrate the generality of our Re4 agent.

\subsubsection{Vapor depression dynamics in laser–metal interaction}\label{subsubsec: keyhole}
During laser-metal interaction, a vapor-filled depression (i.e., keyhole) typically forms in the molten metal pool, as illustrated in figure~\ref{fig:dimcase_schem}(a). Induced by vaporization-driven recoil pressure, its dynamics are inherently complex due to multi-physical dependencies. Quantifying this phenomenon is critical, as it directly influences energy absorption and defect formation in diverse industrial applications.

By leveraging X-ray pulses, images of the keyhole region within metals can be captured at micrometer-level spatial resolution. The keyhole depth $e$, measurable from these X-ray images, is contingent on diverse materials and process parameters, including effective laser power ($\eta_p$), laser scan speed ($V_s$), and laser beam radius ($r_0$). A dataset of keyhole X-ray images was collated from the work of \citet{Zhao_etal_PRX_2019} and \citet{Gan_etal_NatCommun_2021}, encompassing 90 experiments across three materials (titanium alloy Ti6Al4V, aluminium alloy Al6061, and stainless steel SS316) under varied process conditions. These materials are characterized by thermal diffusivity ($\alpha$), density ($\rho$), heat capacity ($C_p$), and the temperature difference ($T_l-T_0$) between melting and ambient temperatures. Thus, the causal relationship is formalized as:
\begin{equation}
e^* = f(\eta_p,V_s,r_0,\alpha,\rho, C_p, T_l-T_0)
\label{equ:Keyhole_}
\end{equation}
where the output variable is normalized as the keyhole aspect ratio $e^*=\frac{e}{r_0}$, a dimensionless parameter widely used to characterize keyhole features~\citep{Fabbro_etal_JLA_2018}. As established in the work of \citet{Xie_etal_NatCommun_2022}, for the present dimensional analysis problem, the dominant dimensionless number emerging from keyhole dynamics is
\begin{equation}
\text{Ke} = \frac{\eta_P}{(T_l - T_0)\rho C_p \sqrt{\alpha V_s r_0^3}}
\label{equ:Ke_dimensionless}    
\end{equation}
This dimensionless number has the same physical-parameter exponents as the newly discovered keyhole number Ke~\citep{Gan_etal_NatCommun_2021} (also referred to as normalized enthalpy~\citep{Ye_etal_AEM_2019}), which can be derived from heat transfer theory. Therefore, we adopt the dimensionless number Ke as the metric to evaluate the success rate of the executable code provided by the agent. Given that dimensionless numbers can be mathematically transformed into equivalent forms by raising them to arbitrary powers (e.g., $\Pi^2, \sqrt{\Pi}$), we propose several constraints on the exponents of physical quantities to restrict the search space and remove the ambiguity associated with equivalent powers of a dimensionless group:
\begin{enumerate}[(i) ]
    \item \textbf{Normalization for Uniqueness:} To avoid identifying equivalent dimensionless numbers with different powers and reduce the computational cost, we select the laser scan speed $V_s$ for normalizing the exponents.
    \item \textbf{Rational Exponents:} The exponents of dimensionless numbers tend to be rational numbers to constrain the search space~\citep{Xie_etal_NatCommun_2022}, with a preference for small rational powers such as -1, 1, or 2, etc.
    \item \textbf{Range Limits:} The typical range of the exponents for dimensionless numbers is limited, with the absolute value of coefficients in most dimensionless numbers and scaling laws being less than 4~\citep{barenblatt2003scaling}.
\end{enumerate}

Based on these formulated constraints, we can explicitly define the evaluation criteria (i.e., the solving success rate) by whether the agent can correctly reconstruct the Ke form from the raw experimental data. Specifically, a successful reconstruction dictates that the exact combination of the derived physical quantity exponents must perfectly match the theoretical Ke number after being normalized by $V_s$. During the search process, the coefficient of determination ($R^2$) between the discovered dimensionless candidate and the keyhole aspect ratio $e^*$ serves as a quantitative metric to evaluate the goodness of fit.

\begin{figure}[!t]
\centering
\begin{subfigure}[b]{0.6\textwidth}
\centering
\includegraphics[width=\textwidth]{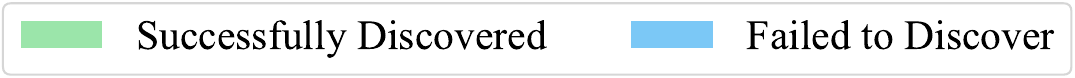}
\end{subfigure}
\begin{subfigure}[b]{0.85\textwidth}
\centering
\includegraphics[width=\textwidth]{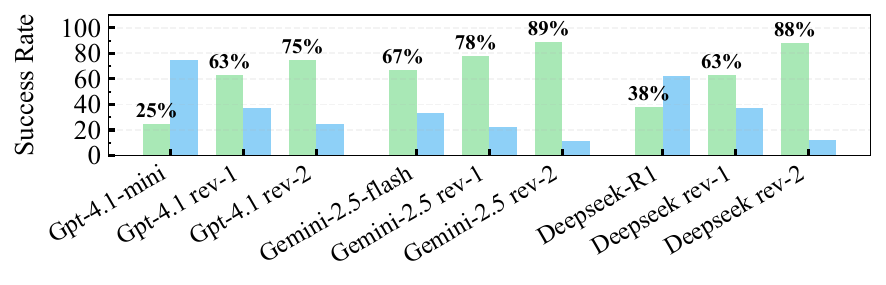}
\end{subfigure}
\caption{The success rate of Programmers' search algorithms in identifying dominant dimensionless quantities Ke.}
\label{fig:keyhole_success_rate}
\end{figure}

Buckingham's $\Pi$ theorem allows the construction of independent dimensionless groups but does not by itself determine which group best describes the observed physical behavior. To identify the dominant dimensionless group, the Programmers consistently use grid search paired with linear regression under the prescribed dimensional and exponent constraints. The solving success rates across eight independent runs are shown in figure~\ref{fig:keyhole_success_rate}. From \texttt{ans-0} to \texttt{rev-2}, the success rate of reconstructing the target Ke form increases by up to 50 percentage points.

\textcolor{black}{By tracing the generated codes, we identify separate revisions to exponent construction and regression assessment. In the inspected DeepSeek-R1 implementation, an incorrect laser-beam-radius exponent leaves a residual length dimension, so the candidate is not dimensionless and cannot satisfy the Ke reconstruction criterion, regardless of its regression fit. Revision corrects this exponent relation. GPT-4.1-mini and Gemini-2.5-flash determine dependent exponents by solving the dimensional constraints, followed by rational approximation and exponent-bound checks. K-fold cross-validation is introduced in the DeepSeek-R1 and Gemini revisions, while GPT-4.1-mini provides regression residual diagnostics. These changes address complementary requirements: dimensional consistency establishes the physical admissibility of a candidate, whereas regression assessment evaluates its ability to describe the observed data.}

\begin{figure}[!t]
\centering
\begin{subfigure}[b]{0.9\textwidth}
\centering
\includegraphics[width=\textwidth]{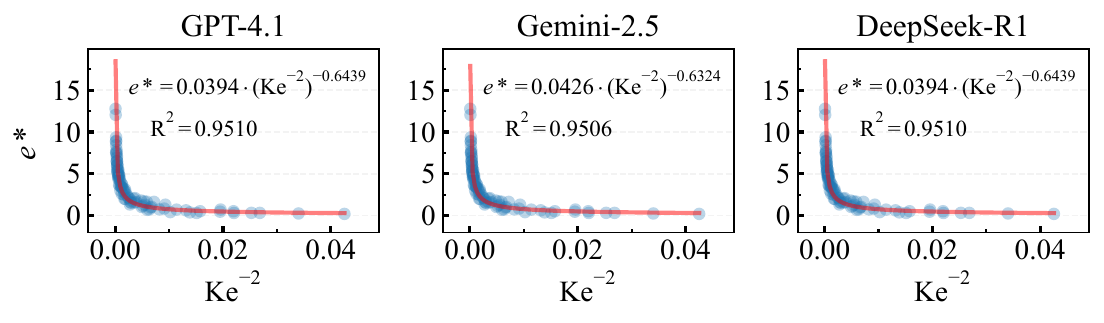}
\end{subfigure}
\caption{Fitting curves of the keyhole aspect ratio $e^*$ as a function of the discovered dimensionless number $\mathrm{Ke}^{-2}$ derived by all Programmers in the final response (\texttt{rev-2}). The red solid lines denote the power-law relationships obtained via linear regression between $\mathrm{Ke}^{-2}$ and $e^*$ in the log-log space.}
\label{fig:keyhole_scale}
\end{figure}

Figure~\ref{fig:keyhole_scale} shows the keyhole aspect ratio $e^*$ as a function of the reconstructed dimensionless group Ke$^{-2}$. The final-stage power-law fits yield $R^2$ values of approximately $0.95$ in log-log space for all three Programmers. These results demonstrate that the Re4 agent can autonomously reconstruct the established dimensionless group and capture its scaling relationship with the dependent variable.

\subsubsection{Turbulent Rayleigh–B\'enard convection}\label{subsubsec: Reyleigh_bernard}

Turbulent Rayleigh-B\'enard convection is a classic fluid mechanics phenomenon describing buoyancy-driven flow in a planar horizontal fluid layer heated from below, as illustrated in figure~\ref{fig:dimcase_schem}(b). As the temperature difference ($\Delta T$) increases, buoyancy forces overcome viscous dissipation and thermal diffusion, developing complex turbulent thermal flows. To quantify the heat transfer efficiency through the fluid layer, the dimensionless Nusselt number ($Nu$) is defined as:
\begin{equation}
Nu = \frac{q h}{\lambda \Delta T}
\label{equ:Nu_definition}
\end{equation}
where $q$ is the heat flux, $h$ is the container height, $\lambda$ is the fluid's thermal conductivity, and $\Delta T$ is the temperature difference. The governing physical parameters of this system include the container height ($h$), thermal conductivity ($\lambda$), temperature difference ($\Delta T$), gravitational acceleration ($g$), thermal expansion coefficient ($\alpha$), kinematic viscosity ($\nu$), and thermal diffusivity ($\kappa$). Thus, the high-dimensional causal relationship is formalized as:
\begin{equation}
Nu = f(h, \lambda, \Delta T, g, \alpha, \nu, \kappa)
\label{equ:Rayleigh_causal}
\end{equation}

For this analysis, we collect an experimental dataset containing 182 cases from the literature~\citep{niemela2006turbulent, chavanne2001turbulent}. According to classical fluid mechanics and similarity theory, the dominant dimensionless group governing such buoyancy-driven convection is the Rayleigh number ($Ra$):
\begin{equation}
Ra = \frac{g \alpha \Delta T h^3}{\nu \kappa}
\label{equ:Ra_dimensionless}    
\end{equation}

To evaluate the agent's solving success rate, we apply the identical fitting metric $R^2$ and physical exponent constraints (including the rational fractional requirements and the absolute range limit of $\le 3$) as described in Section~\ref{subsubsec: keyhole}. Here, normalization fixes the exponent of the thermal expansion coefficient $\alpha$ to 1, removing the ambiguity associated with equivalent powers of the derived dimensionless group. A successful reconstruction is achieved when the normalized exponents of physical quantities derived from the raw data exactly match the classical $Ra$ formulation.

Although their specific physical backgrounds and dimensional variables differ, turbulent Rayleigh--B\'enard convection and keyhole dynamics share the challenge of identifying the dominant dimensionless group while satisfying dimensional homogeneity and exponent constraints. To assess the applicability of the Re4 agent to data-driven physical analysis while maintaining main text brevity, we specifically employ GPT-4.1-mini as the Programmer for this additional case.

\begin{figure}[!t]
\centering
\begin{subfigure}[b]{0.9\textwidth}
\centering
\includegraphics[width=\textwidth]{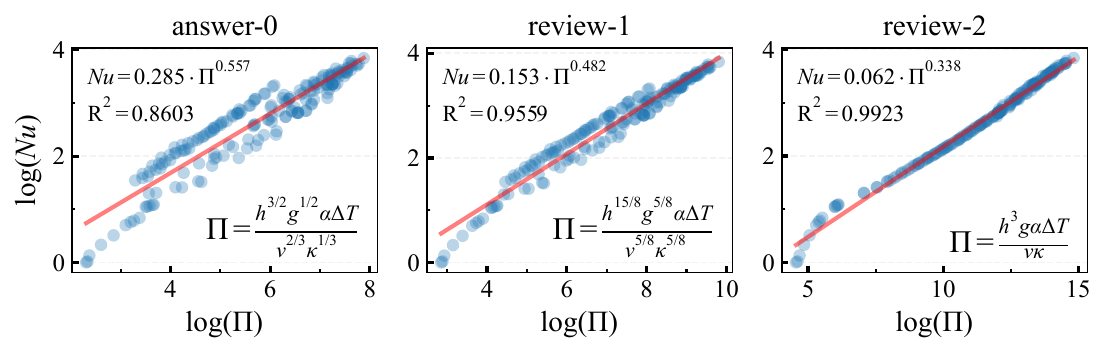}
\end{subfigure}
\caption{The progressive refinement of the discovered dimensionless groups and their log-log linear fits with the Nusselt number ($Nu$) generated by Programmer GPT-4.1-mini across the \texttt{ans-0}, \texttt{rev-1}, and \texttt{rev-2} stages. The red solid lines denote the power-law relationships obtained via linear regression. As the review--revision loop progresses, the selected candidate group changes and ultimately matches the classical Rayleigh number ($Ra$).}
\label{fig:rayleigh_scale}
\end{figure}

\textcolor{black}{Across eight independent runs, GPT-4.1-mini's solving success rate increases from $37\%$ at \texttt{ans-0} to $75\%$ at \texttt{rev-2}, as summarized in Appendix table~\ref{tab:comprehensive_ablation_cost}.} Across both dimensional-analysis tasks, the baseline configuration, prompted directly with the raw natural-language problem description, exhibits the lowest solving success rate. Failures include execution errors and failure to reconstruct the prescribed target groups. Following the Consultant's rewriting and the Reviewer's feedback, the solving success rates increase.

The earlier candidate groups in figure~\ref{fig:rayleigh_scale} satisfy dimensional homogeneity but differ from the classical $Ra$ formulation. Through revision, GPT-4.1-mini reconstructs $Ra$ and improves the fit to the Nusselt-number data, demonstrating a clear trajectory of self-refinement. This comparison highlights the role of regression fitting in selecting among dimensionally homogeneous candidate groups.

From a code-level assessment, GPT-4.1-mini implementations provided in the repository retain a three-step search framework across the response stages: first, computing a null-space basis of the dimensional matrix using SVD to satisfy dimensional homogeneity; second, performing a grid search over linear combinations of the basis vectors; and finally, evaluating the candidate groups through regression with $Nu$ using the $R^2$ score. The revised implementations use log--log linear fitting to represent power-law relationships. 

During the statistical evaluation across independent runs, some initial responses encounter matrix-dimension and indexing errors during SVD-based null-space extraction. Reviewer feedback helps resolve these execution failures in subsequent stages. In the repository examples, the revisions refine normalization with respect to the $\alpha$ exponent and rational approximations using Python's \texttt{Fraction} module. Candidates with a zero $\alpha$ exponent are excluded rather than normalized using another variable. The final-stage regression routine also masks non-positive $Nu$ and candidate-group values before taking logarithms.

\textcolor{black}{The two dimensional-analysis tasks highlight a distinction between enforcing dimensional constraints and selecting the target quantity from physically admissible candidates. In the Keyhole problem, the inspected DeepSeek-R1 implementation contains an incorrect exponent relation that violates dimensional homogeneity. Correcting this relation is therefore necessary to construct a dimensionally valid candidate. In contrast, the earlier candidates in the Rayleigh--Bénard task are already dimensionally homogeneous, but do not recover the target Rayleigh number. The subsequent revisions improve normalization and candidate selection, enabling the identification of \(Ra\). These observations show that satisfying dimensional constraints is necessary but not sufficient for recovering the dominant dimensionless group; effective selection among admissible candidates remains a separate requirement.}

\section{\label{sec:ablation_tokenct_}Ablation study and cost analysis}

In this section, we present a comprehensive ablation study and cost analysis for the proposed Re4 framework based on all evaluated scientific computing tasks. To rigorously isolate the specific contributions of each module, we evaluate the workflow's performance across four distinct reasoning stages under three representative Programmer models:
\begin{enumerate}[ (i)]
    \item \textbf{baseline (Programmer-Only):} A naive, single-stage configuration where each Programmer LLM is prompted directly with the raw natural language problem description, completely bypassing the Consultant's domain augmentation and the Reviewer's iterative feedback loops.
    \item \textbf{ans-0 (with Consultant):} The initial resolution stage where the Programmer generates the first solver candidate utilizing the domain-specific textual expansions and structured alternative algorithms provided by the Consultant module.
    \item \textbf{rev-1 and rev-2 (with Reviewer):} The sequential review-revision stages where the Programmer iteratively refines and debugs the generated solver by incorporating the exact compiler error traces, execution runtime outputs, and qualitative code optimization suggestions provided by the Reviewer module.
\end{enumerate}

\subsection{Ablation of collaborative modules and API cost}
\label{subsec:ablation_APIcost_}
Table~\ref{tab:comprehensive_ablation_cost} summarizes the execution/solving success rates across all nine evaluated tasks, the average input and output token consumption (in thousands, k), and the average response latency (in seconds, s) at each reasoning stage for the different Programmer models.
As observed, for relatively straightforward textbook-style PDEs (e.g., Poisson, Helmholtz, and Burgers' equations), the baseline configurations already exhibit relatively stable execution success rates. For these tasks, the subsequent intervention of the Reviewer module helps raise the success rate to a range of $89\%\sim100\%$. In contrast, the remaining tasks involve more complicated numerical implementations and generally require more detailed instructions and higher token consumption. In these scenarios, the integration of the Consultant can play a more pronounced role in improving the initial solving success rate at the \texttt{ans-0} stage by systematically structuring the numerical solution strategies. With subsequent intervention from the Reviewer, the success rates are further improved. The numerical decisions, implementation difficulties, and review--revision trajectories underlying these stage-wise changes are further discussed in Section~\ref{sec:discussion}.

Beyond the individual contributions of each module to the overall success rates, table~\ref{tab:comprehensive_ablation_cost} also provides a detailed breakdown of the token counts and latency, clarifying the API cost and computational budget of the proposed Re4 framework. The token usage dynamically reflects the multi-agent collaborative process: the difference in input tokens between the baseline and \texttt{ans-0} stages directly corresponds to the expanded domain-specific knowledge and alternative pseudocode-style resolution strategies provided by the Consultant during the rewriting phase. Once the Programmer completes the initial resolution, the orchestration enters the review-revision loop. Here, the token increment between the input of the subsequent stage and the output of the previous stage represents the newly accumulated context, specifically the run-time execution logs and the structured comments and actionable suggestions meticulously provided by the Reviewer.

By analyzing the Programmer's output tokens against the response time, we can further estimate the Tokens Per Second (TPS) throughput for different LLMs. Gemini-2.5-flash and GPT-4.1-mini exhibit comparable response times across various stages, but Gemini consistently generates a larger volume of output tokens, yielding a higher throughput (approximately ranging from 110 to 130 TPS, compared to GPT-4.1-mini's $70 \sim 100$ TPS). In contrast, DeepSeek-R1 typically generates fewer output tokens but incurs a noticeably larger time overhead due to its intrinsic chain-of-thought reasoning process, resulting in a much lower TPS of approximately $20 \sim 35$. 
Furthermore, as a model with advanced reasoning capabilities specialized for the flow past a cylinder task (detailed in table~\ref{tab:cylinder_token_rate}), GPT-5-mini also exhibits a high throughput, with its TPS ranging from approximately 130 to 160.

\subsection{\label{subsec:cost_benefit_compare}Cost-benefit comparison with existing baselines}
Among existing LLM-based scientific computing agents, PINNsAgent~\citep{wuwu2025pinnsagent} primarily optimizes PDE solutions by tuning neural network architectures and training hyperparameters rather than designing numerical solvers from scratch. On the other hand, agents developed for CFD typically rely on calling external commercial or open-source software packages to execute simulations. For instance, CFDagent~\citep{xu2025cfdagent} integrates OpenAI's Point-E~\citep{nichol2022point} to generate computational grids from textual descriptions, which are then imported into in-house codes to perform CFD simulations employing diffuse-interface immersed boundary methods, while other OpenFOAM-focused agents~\citep{pandey2025openfoamgpt} leverage retrieval-augmented generation (RAG) to synthesize case configuration dictionaries and automate solver executions within the OpenFOAM suite. 
In contrast, CodePDE~\citep{Li_etal_arXiv_2025} serves as the most directly comparable and relevant baseline for generating from-scratch PDE solvers. 

CodePDE divides the PDE solver generation process into two sequential, resource-intensive stages: execution-driven \textit{debugging} followed by metric-guided \textit{refinement}. In the initial debugging stage, CodePDE utilizes a feedback loop (up to 4 iterations) to resolve syntax and runtime errors, aiming purely to produce a runnable script. Once a set of executable solvers is obtained, they enter the subsequent refinement stage, where the framework evaluates their numerical accuracy by calculating the normalized Root Mean Squared Error (nRMSE) against ground-truth solutions and prompts the LLM to iteratively optimize the solvers using this error feedback. Crucially, to achieve optimal performance, CodePDE heavily relies on massive test-time scaling across both stages, generating up to 32 parallel candidates (best-of-32) to guarantee a runnable baseline during debugging, and another 12 candidates during refinement. \textcolor{black}{Overall, single-shot generation in CodePDE achieves an average bug-free solver generation rate of $41.9\%$, increasing to $86.2\%$ after up to four rounds of iterative self-debugging. Detailed results for individual PDE families are reported by \citet{Li_etal_arXiv_2025}.}

Instead of relying on massive parallel sampling and repeated execution trials to filter out bugs, Re4 utilizes the Consultant to proactively  lay out alternative pseudocode-style solution algorithms, notably reducing the initial algorithmic trial-and-error space. Furthermore, CodePDE's refinement strictly depends on the nRMSE metric, inherently requiring access to ground-truth data, a constraint that is typically unavailable in realistic scientific discovery scenarios. 
To bypass these restrictive metric dependencies, the Reviewer module within our framework evaluates the generated code across three predefined dimensions: \texttt{runtime} (execution errors or failures), \texttt{structure} (code organization and quality), and \texttt{accuracy} (numerical stability, precision, and convergence behavior). This multidimensional assessment facilitates the autonomous correction of frequent coding pitfalls, such as the array indexing mismatches commonly encountered in staggered grids, while driving critical algorithmic upgrades without relying on an explicit ground-truth oracle. 

In addition to the fundamental differences in workflow designs discussed above, our framework and CodePDE also diverge significantly in their problem configurations (e.g., CodePDE predominantly employs periodic boundary conditions for time-dependent PDEs like the Burgers' equation, effectively bypassing the severe numerical complexities and stability risks associated with strict boundary treatments). Under these unequal conditions, executing a direct, head-to-head quantitative comparison of error metrics on identical benchmarks is neither directly comparable nor conceptually fair. Therefore, beyond the workflow-oriented cost-benefit analysis discussed above, we provide a detailed, code-level qualitative comparison of the generated solvers in Appendix~\ref{subapp:code_comparison} to further demonstrate the algorithmic robustness of our framework.

Overall, the proposed Re4 agent achieves improved execution success rates across the evaluated configurations within two defined review-revision iterations, providing a favorable cost-benefit balance for autonomous scientific computing without requiring exhaustive trial-and-error runs or massive parallel sampling.

\section{Discussion}
\label{sec:discussion}

\textcolor{black}{Beyond the aggregate results, the progressive trial-and-error process provides insights into the relationships among numerical choices, implementation difficulties, and subsequent corrections. We examine these relationships below, with supporting records identified in Appendix~\ref{app:sample_evidence}.}

\subsection{Numerical choices and implementation challenges}

The generated PDE solvers follow established numerical formulations, with different requirements for discretization, stability, and boundary treatment:

\begin{enumerate}[(i)]
\setlength{\itemsep}{0.3em}

\item \textbf{Sod shock tube.}
The common HLLC flux is combined with different reconstruction and time-integration schemes in figures~\ref{fig:Sod_space_method} and~\ref{fig:Sod_time_method}. Limited MUSCL reconstruction provides higher-order spatial approximations while controlling oscillations near discontinuities. Runge--Kutta integration introduces additional stages, requiring consistent interface states, flux evaluation, and boundary updates under the corresponding stability restrictions.

\item \textbf{Lid-driven cavity.}
The best final-stage implementations use ACM to advance pressure and velocity toward a steady solution in pseudo-time, without a separate pressure-Poisson correction at each update. This provides a direct coupling strategy, but correct convection directions, array indexing, and sufficient steady-state convergence remain necessary.

\item \textbf{Unsteady NS.}
Projection separates the momentum update from pressure correction, requiring compatibility between the pressure equation, discrete divergence, and pressure gradient. Revisions replace wrapped differences with non-periodic boundary stencils and introduce ILU-preconditioned GMRES. These changes address boundary representation and linear convergence, while continuity still depends on consistent operators and boundary enforcement after velocity correction.

\item \textbf{Flow past a cylinder.}
The revised solvers combine incompressible NS equations with immersed-boundary penalization on a MAC grid. 
Explicit convection using the second-order Adams--Bashforth (AB2) method is coupled with implicit diffusion and penalization, alleviating the restrictions of explicit treatment of the latter terms; reusable sparse LU factorizations reduce repeated solution costs. Implementation additionally requires consistent face locations, cylinder masks, sparse indices, and boundary treatment across the pressure--velocity update.

\end{enumerate}

\textbf{From-scratch PDE solvers.}
Higher-order shock-capturing updates, projection-based solvers, and immersed-boundary coupling impose substantial coordination requirements. The examined failures identify consistency between interface states, grid representations, discrete operators, and boundary updates as a demanding part of their implementation. Refinement can therefore improve an existing scheme without replacing its overall numerical formulation.

\textbf{Ill-conditioned linear systems.}
Severe ill-conditioning in the Hilbert problem amplifies numerical errors and requires special treatment. Revised implementations use LU with pivoting, regularized Cholesky, preconditioned CG, and truncated-SVD solution to address factorization difficulties and inaccurate recovery. Adopting a stabilization technique alone does not ensure the required accuracy for the original system.

\textbf{Dimensionless-group search.}
Dependent-exponent solving and SVD null-space search provide different ways to construct dimensionally homogeneous candidates. Normalization and rational bounds restrict the search, while regression ranks candidates by their fit to the keyhole aspect ratio or Nusselt number. Success requires both correct enforcement of physical constraints and effective selection within the admissible candidate space.

\subsection{Representative trial-and-error trajectories}

\textbf{Initial failure and diagnosis.}
In the lower row of figure~\ref{fig:cylinder_opt_steps}, the initial GPT-5-mini solver fails before time integration: the streamwise velocity array has dimensions $(221,120)$, whereas its cylinder mask has dimensions $(120,221)$. Review focuses on the correspondence between grid coordinates, staggered velocities, and masks, together with the indexing convention used to map arrays into sparse-system vectors. The latter inconsistency would remain even after correcting the visible shape mismatch.

\textbf{Implemented correction.}
The revised implementations align mask construction and array--vector mappings. These repairs make the solver executable while retaining its semi-implicit MAC formulation and immersed-boundary treatment. The example shows how a runtime exception can expose a broader inconsistency in the discrete representation, requiring coordinated corrections rather than a change of numerical method.

\textbf{Boundary refinement.}
The upper row presents an executable solver whose intermediate field contains low-speed bands near the upper and lower boundaries. The final revision adjusts boundary discretization and reapplies free-slip conditions after pressure projection; the bands are no longer evident. In the lower row, \texttt{rev-1} resets cylinder-interior velocities after projection, whereas \texttt{rev-2} retains implicit penalization without this reset and also changes pressure correction. These simultaneous changes highlight the need to coordinate pressure correction and boundary treatment, without isolating the effect of a single operation.

\textbf{Correcting physical constraints.}
In the Keyhole example, fixing the scan-speed exponent at one requires the beam-radius exponent to be $a+2f+1$, where $a$ and $f$ are the effective-power and heat-capacity exponents. The initial DeepSeek-R1 code instead uses $a+2f+2$, leaving a residual length dimension. Revision corrects this algebraic relation while preserving the search framework. The failure concerns the translation of dimensional constraints into dependent exponents, rather than incorrect dimensions assigned to the input quantities.

\textbf{Refining candidate selection.}
The earlier Rayleigh groups in figure~\ref{fig:rayleigh_scale} already satisfy dimensional homogeneity. Revisions retain the SVD-based search but exclude candidates with a zero exponent of the thermal expansion coefficient $\alpha$, enforce the prescribed normalization, and refine candidate filtering and log--log regression. The final group matches $Ra$, with an improved displayed fit. Unlike the Keyhole correction, this trajectory concerns selection among physically admissible candidates, demonstrating that constraint satisfaction and effective search are distinct requirements.

\subsection{Lessons for numerical solver development}

The examined trajectories show different diagnostic priorities: runtime exceptions direct attention to dimensions and indexing, whereas executable but non-physical results require examination of boundary treatment, coupling, and convergence. Several corrections preserve the numerical formulation while repairing state conversions, convection directions, or data mappings; others modify reconstruction, time integration, linear-system treatment, or candidate selection. The process therefore extends beyond a progression from simple to more advanced methods.

A recurring difficulty is maintaining consistency between individually reasonable operations. Pressure-system convergence does not establish a divergence-free final velocity, stabilization does not establish accurate recovery of a Hilbert solution, and a good fit does not establish dimensional homogeneity. These observations suggest checking discrete representations before integration, coupled constraints after the complete update, and revised results against the original mathematical requirements.

For researchers using LLMs to develop scientific-computing software, these trajectories help distinguish implementation repair from changes in numerical strategy and identify where quantitative checks can guide revision. Their contribution concerns the construction and refinement of executable solutions using established methods. However, because several tasks use standard benchmarks, we cannot rule out that the LLMs recall established algorithms encountered during training.

\section{Conclusions}\label{sec:Conclusion}

In this work, we proposed a reliable agent framework for autonomous code generation in scientific computing without additional domain-specific training. We quantitatively assess its performance on textbook-style PDEs, engineering-oriented flow past a cylinder using an immersed boundary method, ill-conditioned linear systems, and data-driven physical analysis. Evaluation across problems with distinct numerical characteristics demonstrates the framework's applicability to diverse scientific-computing tasks. The statistical results quantify improvements in execution reliability, while representative solver-development trajectories illustrate how runtime feedback supports progressive debugging and refinement. With guidance from the Consultant and feedback from the Reviewer, the generated solutions show overall improvements in execution and solving success, although solution accuracy does not improve monotonically across all response stages.

This agent framework provides a collaborative platform for multiple LLMs, overcoming the reasoning limitations of single models. The core modules, Consultant, Programmer, and Reviewer, support heterogeneous model combinations (such as ChatGPT, DeepSeek, and Gemini), and the Reviewer module allows for parallel evaluation by multiple models. Multi-model collaboration mitigates hallucination issues and algorithmic biases stemming from the inherent preferences of individual models. For instance, in solving the Sod shock tube problem, Gemini 2.5-flash consistently selected the forward Euler scheme; however, after review by GPT 4.1-mini, it shifted toward adopting higher-order time-integration schemes.

Looking ahead, several directions exist to enhance the current agent. First, while the multi-agent collaboration significantly boosts reliability, it inevitably incurs higher token usage and time overhead compared to single-step generation. Future optimizations will explore adaptive inference strategies to balance computational cost with solution quality. Second, the current workflow relies heavily on the Consultant's initial strategy. If the Consultant proposes a flawed strategy, the Reviewer may struggle to steer the Programmer back on track. To address this, we plan to introduce a backtracking mechanism that allows the agent to revisit the Consultant phase upon persistent failure.

Beyond these workflow improvements, while the Reviewer module categorizes code issues into runtime, structure, and accuracy, its comments remain primarily qualitative and text-based. This limitation highlights the need for a more comprehensive and quantitative evaluation system to formally assess solver quality. Additionally, the LLMs in the Reviewer module may encounter token limitations when handling long contexts (such as review comments combined with runtime logs), necessitating the development of information distillation algorithms or adaptations for long-context models. Furthermore, we observed that LLMs generate code with version lag issues (e.g., outdated Python syntax), indicating a need to update and optimize the LLMs themselves or integrate them with domain knowledge bases.

As LLMs' reasoning capabilities continue to evolve, the performance of this agent framework is expected to improve further. Finally, to further broaden the framework's applicability to complex engineering problems or tasks involving industrial software like OpenFOAM, future work will focus on augmenting each module with dynamic knowledge retrieval, enabling the agent to incorporate both domain-specific insights from technical literature and operational guidelines from software documentation.

In summary, the proposed framework supports automatic code generation and review for scientific computing. Analysis of its trial-and-error trajectories can yield insights into numerical choices, implementation difficulties, and corrective strategies. These insights may offer a new perspective on physically interpretable algorithm design grounded in explicit numerical reasoning.

\ack{The authors would like to thank Zhaobin Li for his suggestions on early topic selection for this paper. OpenAI's ChatGPT (GPT-6.1) was used to assist with language editing and improving the clarity of the manuscript, as well as to generate the robot illustrations used in Figure~\ref{Fig_schematic_agent}. All scientific analyses, interpretations, and conclusions were reviewed and verified by the authors, who take full responsibility for the content of this manuscript.
}

\funding{
This work was supported by the National Natural Science Foundation of China (NSFC) Excellence Research Group Program for ``Multiscale Problems in Nonlinear Mechanics'' (No. 12588201) and by the CAS Project for Young Scientists in Basic Research (No. YSBR-136). L. Zhang was also supported by the National Natural Science Foundation of China (NSFC) under Grant No. 12202451.}

\data{The source code and data that support the findings of this study are openly available at the following URL/DOI:~\url{https://github.com/ChengAo21/Re4_Sci_Agent}.}

\appendix

\section{Agent implementation and prompts}\label{app:agent_protocols}

\subsection{Implementation protocols and structured interfaces}\label{app:implementation}

The four stages in Section~\ref{sec:Methodology} exchange information through predefined inputs and schema-constrained structured outputs, as summarized in table~\ref{tab:module_interfaces}. The implementation details below describe the associated routing logic, context management, and execution reporting.

\begin{table}[!b]
\centering
\color{black}
\footnotesize
\renewcommand{\arraystretch}{1.2}
\setlength{\tabcolsep}{3pt}
\begin{adjustbox}{max width=\textwidth}
\begin{tabular}{>{\raggedright\arraybackslash}p{2.1cm}|>{\raggedright\arraybackslash}p{5cm}|>{\raggedright\arraybackslash}p{8cm}}
\hline
\textbf{Operation / role} & \textbf{Input to the call} & \textbf{Structured output}\\
\hline
Rewriting / Consultant & Original problem (\texttt{prob\_todo}). & \texttt{expanded\_prob}: expanded context; \texttt{solution\_plans}: plan titles (\texttt{solu\_name}) and algorithm descriptions (\texttt{content}).\\
\hline
Resolution / Programmer & Original problem; expanded context and solution plans when the Consultant is enabled. & \texttt{technical\_spec}: algorithm and data flow; \texttt{python\_codes}: complete Python script.\\
\hline
Review / Reviewer & Original problem; expanded context when enabled; latest specification and code; runtime report; optional image-assessment text. & \texttt{review\_summary}; \texttt{review\_decision} (\texttt{accept}/\texttt{revise}); \texttt{review\_comments} (category, issue, severity, recommendation).\\
\hline
Revision / Programmer & Original problem; latest specification and code; runtime report; Reviewer comments. & Updated \texttt{technical\_spec} and \texttt{python\_codes}.\\
\hline
Python execution tool & Latest \texttt{python\_codes}. & \texttt{runtime\_outputs}: logs, numerical outputs, warnings, and errors; generated figures.\\
\hline
Optional image assessment & Original problem; latest code; runtime output; generated images. & \texttt{image\_description}: per-image descriptions and identified issues with severity; passed to Review as \texttt{rev\_image\_description}.\\
\hline
\end{tabular}
\end{adjustbox}
\caption{Structured interfaces and information flow across the four stages of Re4. The Python execution tool supports both Resolution and Revision, while optional image assessment returns textual observations to the Review stage.}
\label{tab:module_interfaces}
\end{table}

To ensure operational reliability and prevent parsing errors between nodes, each LLM module employs structured output protocols based on Pydantic-validated JSON schemas. These schemas constrain the generated responses to predefined formats (e.g., strictly isolating the Python script from explanatory text, or categorically formatting the review decisions), thereby ensuring robust data transfer within the automated pipeline. The detailed prompt templates for each module are provided in Appendix~\ref{app:prompt_for_agent}.

The collaborative framework encompasses the following four sequential stages, executed by three LLM-based modules (Consultant, Programmer, Reviewer) and a script execution tool invoking the local Python interpreter:

\begin{enumerate}[(i) ]

\item \textbf{Rewriting (Performed by the Consultant module).} Functioning as a mathematical consultant and numerical analyst, this stage expands the context of the original problem statement. \textbf{Input:} The vague natural language problem description. \textbf{Output:} Based on predefined schemas, it generates an \texttt{expanded\_prob} that explicitly identifies the underlying mathematical and numerical challenges, and \texttt{solution\_plans} containing a variety of alternative solution strategies (e.g., pseudocode or structured plans). This domain-specific text augmentation essentially functions as robust query enrichment, which has been studied as a way to reduce hallucinations and steer the model toward relevant features \citep{abdaljalil2025safe}. Consequently, it deepens task understanding and provides explicit algorithmic guidance, thereby effectively enhancing the reliability of the subsequently generated code.

\item \textbf{Resolution (Handled by the Programmer module \& the script execution tool).} Acting as an expert Python programmer, this stage translates the Consultant's algorithmic design into an initial executable solver. \textbf{Input:} The original problem statement, Consultant's augmented context and solution plans. \textbf{Output:} Adhering to structured output protocols, the Programmer generates a \texttt{technical\_spec} (a concise architectural description of the core algorithm and data flow) and a standardized \texttt{python\_codes} script block. Immediately following code generation, the script execution tool automatically runs the code using the local Python interpreter to produce a comprehensive runtime report that aggregates execution outputs, interpreter warnings, error tracebacks, and visualization images, serving as the input for the subsequent review–revision stage.

\item \textbf{Review (Conducted by the Reviewer module).} Serving as an independent code reviewer and scientific computing expert, this stage assesses the reliability of the execution results and the quality of code implementation. \textbf{Input:} Governed by an LLM independent of the Programmer, it evaluates inputs comprising the Consultant's augmented context, the generated code, and the execution tool's runtime logs. Furthermore, when image assessment is enabled, runtime-generated visualization images are encoded via \texttt{Base64}, enabling a separate multimodal review of the physical validity of the numerical solutions. \textbf{Output:} The Reviewer is constrained by the output schema to return a strict binary \texttt{review\_decision} (Accept or Revise) and a structured list of \texttt{review\_comments}. Each comment must be explicitly categorized into \textit{runtime} (execution failures), \textit{structure} (code quality), or \textit{accuracy} (numerical precision/stability), and assigned a severity level (MAJOR or MINOR), thus establishing a rigorous diagnostic report.

\item \textbf{Revision (Performed by the Programmer module during the revision loop).} Governed by the conditional edges of the graph architecture, a ``Revise'' decision dynamically routes the workflow back to the Programmer for refinement. \textbf{Input:} Rather than sharing a verbose and continuous chat history, the Programmer receives only the original problem, the latest technical specification and generated code, the execution tool's runtime logs, and the Reviewer's structured recommendations. \textbf{Output:} Based on the Reviewer's comments, the Programmer addresses the root causes of the failures in the previous script, improves the code quality, optimizes the relevant numerical algorithms, and ultimately outputs an updated technical specification and code, after which the revised code is re-executed to close the review-revision loop.
\end{enumerate}

\textbf{Connection and implementation:} The four stages are systematically integrated through the LangGraph architecture. The workflow initiates sequentially from Rewriting to Resolution, and then enters the Review-Revision stage. Within this loop, conditional edges evaluate the runtime state: an ``Accept'' decision terminates the process, while a ``Revise'' decision routes the workflow back to the Programmer node. To sustain reasoning quality throughout these iterative transitions, context histories are strictly isolated between nodes. Instead of maintaining a continuous, multi-turn chat history, each module receives only the specific inputs required for its task, which helps mitigate the attention decay and context dilution typically observed in long conversations. For the purpose of systematically analyzing how this review-revision loop drives progressive improvements, our experimental evaluation enforces exactly two iterations of the loop (yielding the \texttt{rev-1} and \texttt{rev-2} stages) across all evaluated tasks, regardless of early ``accept'' decisions. This ensures a clear, step-by-step tracking of the agent's self-refinement trajectory.

To manage context window constraints during the review-revision loop, we implement a selective context preservation strategy for terminal outputs. The system retains only the beginning and end segments of the runtime logs (e.g., 800 characters each), substituting the intermediate redundant content with a descriptive prompt marker (e.g., ``\textit{[Truncated: Content too long...]}''). By filtering out redundant intermediate logs, this strategy ensures that the agent preserves the most critical information (e.g., the error origin and final convergence state), thereby preventing reasoning failures caused by excessively long prompts.

\subsection{\label{app:prompt_for_agent} Detailed construction of agent prompt template}

\begin{tcolorbox}[
    title={Consultant module prompt}, 
    colback=white,                        
    colbacktitle=white,                   % 原为 black!60，改为纯白防灰度报警
    coltitle=black,                       % 标题字体设为黑色
    colframe=black,                       % 原为 gray!50，改为纯黑
    fonttitle=\bfseries\scriptsize,                  
    width=\linewidth,                     
    boxsep=6pt,                           
    top=3pt, bottom=3pt, left=3pt, right=3pt  
]
\scriptsize
\textbf{[Role]} You are a distinguished Mathematical Consultant and Numerical Analyst. 

\begin{tcolorbox}[
    colback=white,                 % 原为 gray!10，改为纯白
    colframe=black,                % 原为 black!30，改为纯黑
    boxrule=0.5pt,                 
    arc=2pt,                       
    width=\linewidth,
    boxsep=4pt,
    left=5pt, right=5pt, top=3pt, bottom=3pt
]
\textbf{[Task]}\\                  % 删除了 \textcolor{red!80}
Expand the context of a given scientific problem and generate multiple alternative solution options.\\
\textbf{[Goal]}\\                  % 删除了 \textcolor{red!80}
Accurately dissect the problem by preserving its original meaning, explicitly identifying the underlying mathematical and numerical challenges, and formulating rigorous and well-structured solution strategies.

\begin{tcolorbox}[
    colback=white,                 % 原为 GreenYellow!10，改为纯白
    colframe=black,                % 原为 gray!30，改为纯黑
    boxrule=0.5pt,                 
    arc=2pt,                       
    width=\linewidth,
    boxsep=4pt,
    left=5pt, right=5pt, top=3pt, bottom=3pt
]
\textbf{[Problem Statement]}\\     % 删除了 \textcolor{blue!80}
$<Problem\; Description>$

\end{tcolorbox}

\textbf{[Output Requirement]}\\    % 删除了 \textcolor{Melon!95}
Provide an expanded contextualization of the original problem and multiple solution plans.
\begin{enumerate}[(i)]
    \item Elaborate on the original problem outline while preserving its core meaning, and explicitly identify the primary mathematical and numerical challenges.
    \item Provide a variety of alternative solution strategies, each presented in a structured formulation (e.g., concise pseudocode or numbered steps), rather than long continuous paragraphs.
\end{enumerate}

\end{tcolorbox}
\end{tcolorbox}

% THE Programmer LLM--first time answer

\begin{tcolorbox}[
    title={Programmer module prompt (initial answer)}, 
    colback=white,                      
    colbacktitle=white,
    coltitle=black,
    colframe=black,
    fonttitle=\bfseries\scriptsize,
    width=\linewidth,
    boxsep=6pt,
    top=3pt, bottom=3pt, left=3pt, right=3pt
]
\scriptsize
\textbf{[Role]} You are an expert Python Programmer specialized in scientific computing.

\begin{tcolorbox}[
    colback=white,
    colframe=black,
    boxrule=0.5pt,               
    arc=2pt,
    width=\linewidth,
    boxsep=4pt,
    left=5pt, right=5pt, top=3pt, bottom=3pt
]
\textbf{[Task]}\\
For a specific problem description and the associated candidate solving algorithms, select an appropriate algorithm and implement it in Python to resolve the problem.\\
\textbf{[Goal]}\\
Produce a complete, well-structured, and executable Python script that correctly realizes the selected algorithm, with clear modular organization and quantitative outputs suitable for verification.     

\begin{tcolorbox}[
    colback=white,               
    colframe=black,              
    boxrule=0.5pt,                 
    arc=2pt,                       
    width=\linewidth,
    boxsep=4pt,
    left=5pt, right=5pt, top=3pt, bottom=3pt
]
\textbf{[Problem Statement]}\\
$<Problem\; Description>$
\end{tcolorbox}

\begin{tcolorbox}[
    colback=white,               
    colframe=black,              
    boxrule=0.5pt,                 
    arc=2pt,                       
    width=\linewidth,
    boxsep=4pt,
    left=5pt, right=5pt, top=3pt, bottom=3pt
]
\textbf{[Consultant expansion]}\\
$<Complete\;description\;of\;the\;problem>$\\
$<Descriptions\;of\;Multiple\;solution\;plans>$
\end{tcolorbox}

\textbf{[Output Requirement]}\\
Based on the specific description of the original problem and multiple solving algorithms, select an appropriate method to resolve the problem.
\begin{enumerate}[(i) ]
    \item Provide a concise architectural description of the core algorithm and data flow.
    \item Provide a complete, bug-free Python script.
\end{enumerate}
\textbf{The output code should be enclosed in a single ${```\text{python} ... ```}$ markdown block as follows:} % 【修正】删除了 \textcolor{Melon}
\begin{tcolorbox}[
    colback=white,               
    colframe=black,              
    boxrule=0.5pt,                 
    arc=2pt,                       
    width=\linewidth,
    boxsep=4pt,
    left=5pt, right=5pt, top=1pt, bottom=1pt
]
\scriptsize
\begin{verbatim}
# Technical explanation content

```python

# All required imports included at the top.
# Modular structure with clear function definitions.
# Use brief, clear code comments, avoid verbose or tutorial-style explanations.
# Must include quantitative outputs: printed metrics or labeled plots.

'''
\end{verbatim}
\normalsize  
\end{tcolorbox}
\end{tcolorbox}
\end{tcolorbox}

% THE Reviewer LLM

\begin{tcolorbox}[
    title={Reviewer module prompt}, 
    colback=white,                        
    colbacktitle=white,                    % 【修正】改为 white
    coltitle=black,                        % 【新增】确保标题字是黑色
    colframe=black,                        % 【修正】改为 black
    fonttitle=\bfseries\scriptsize,                  
    width=\linewidth,                     
    boxsep=6pt,                           
    top=3pt, bottom=3pt, left=3pt, right=3pt  
]
\scriptsize
\textbf{[Role]} You are a Code Reviewer and Scientific Computing Expert. 

\begin{tcolorbox}[
    colback=white,                  % 【修正】改为 white
    colframe=black,                 % 【修正】改为 black
    boxrule=0.5pt,                 
    arc=2pt,                       
    width=\linewidth,
    boxsep=4pt,
    left=5pt, right=5pt, top=3pt, bottom=3pt
]
\textbf{[Task]}\\                   % 【修正】删除了 \textcolor
Review the reliability of numerical results and the quality of code implementation, then provide comments.\\
\textbf{[Goal]}\\                   % 【修正】删除了 \textcolor
Conduct a structured review assessment based on the problem description, submitted code, and runtime output.\\
Be pragmatic in your decision: if results are reasonable, \textbf{\textsc{Accept}} it; but if you request a \textbf{\textsc{Revise}}, you must be thorough and list all technical blockers.

\begin{tcolorbox}[
    colback=white,                  % 【修正】改为 white
    colframe=black,                 % 【修正】改为 black
    boxrule=0.5pt,                 
    arc=2pt,                       
    width=\linewidth,
    boxsep=4pt,
    left=5pt, right=5pt, top=3pt, bottom=3pt
]
\textbf{[Problem Statement]}\\      % 【修正】删除了 \textcolor
$<Problem\; Description>$
\end{tcolorbox}

\begin{tcolorbox}[
    colback=white,                  % 【修正】改为 white
    colframe=black,                 % 【修正】改为 black
    boxrule=0.5pt,                 
    arc=2pt,                       
    width=\linewidth,
    boxsep=4pt,
    left=5pt, right=5pt, top=3pt, bottom=3pt
]
\textbf{[Consultant expansion]}\\   % 【修正】删除了 \textcolor
$<Complete\;description\;of\;the\;problem>$\\
\\
\textbf{[Programmer solution]}\\    
$<Technical\;Specification,\;Code,\;Runtime\;Report>$
\end{tcolorbox}

\textbf{[Output Requirement]} \\    % 【修正】删除了 \textcolor
Analyze the results output by the code and provide detailed comments. Ensure your structured review includes:
\begin{enumerate}[ (i)]
    \item \textbf{Review Decision}: Make a strict binary choice between \texttt{accept} and \texttt{revise}.
\begin{tcolorbox}[
    colback=white,                  % 【修正】改为 white
    colframe=black,                 % 【修正】改为 black
    boxrule=0.5pt,                 
    arc=2pt,                       
    width=\linewidth,
    boxsep=4pt,
    left=5pt, right=5pt, top=0pt, bottom=0pt
]
\scriptsize
\begin{verbatim}
- Accept: If code runs without errors, results are correct, and no MAJOR issues exist.
- Revise: If runtime errors, non-physical results, or any MAJOR issue exists.
\end{verbatim}
\normalsize
\end{tcolorbox}
     \item \textbf{Review Comments}: Provide a list of specific review comments. Each comment must strictly contain:
\begin{tcolorbox}[
    colback=white,                  % 【修正】改为 white
    colframe=black,                 % 【修正】改为 black
    boxrule=0.5pt,                 
    arc=2pt,                       
    width=\linewidth,
    boxsep=4pt,
    left=5pt, right=5pt, top=0pt, bottom=0pt
]
\scriptsize
\begin{verbatim}
- Category: "runtime" (execution errors), "structure" (code quality), or "accuracy".
- Issue: A concrete description of the identified problem.
- Severity: "MAJOR" (causes runtime failure/invalid results) or "MINOR" (cosmetic/stylistic).
- Recommendation: Specific suggestions and actionable steps for refinement.
\end{verbatim}
\normalsize
\end{tcolorbox}
\end{enumerate}

\end{tcolorbox}
\end{tcolorbox}

% THE Programmer LLM --in revise loop

\begin{tcolorbox}[
    title={Programmer module prompt (in revise loop)}, 
    colback=white,                      
    colbacktitle=white,                   % 【修正】改为 white
    coltitle=black,                       % 【新增】标题字设为黑
    colframe=black,                       % 【修正】改为 black
    fonttitle=\bfseries\scriptsize,                  
    width=\linewidth,                     
    boxsep=6pt,                           
    top=3pt, bottom=3pt, left=3pt, right=3pt  
]
\scriptsize
\textbf{[Role]} You are an expert Python Programmer specialized in scientific computing.

\begin{tcolorbox}[
    colback=white,                  % 【修正】改为 white
    colframe=black,                 % 【修正】改为 black
    boxrule=0.5pt,                 
    arc=2pt,                       
    width=\linewidth,
    boxsep=4pt,
    left=5pt, right=5pt, top=3pt, bottom=3pt
]
\textbf{[Task]}\\                   % 【修正】删除了 \textcolor
Revise the previously generated Python implementation based on the provided execution results and structured review comments.\\
\textbf{[Goal]}\\                   % 【修正】删除了 \textcolor
Analyze the error trace to identify the underlying root cause of the previous failure, and address the recommendations from review comments exactly.

\begin{tcolorbox}[
    colback=white,                  % 【修正】改为 white
    colframe=black,                 % 【修正】改为 black
    boxrule=0.5pt,                 
    arc=2pt,                       
    width=\linewidth,
    boxsep=4pt,
    left=5pt, right=5pt, top=3pt, bottom=3pt
]
\textbf{[Problem Statement]}\\      % 【修正】删除了 \textcolor
$<Problem\; Description>$
\end{tcolorbox}

\begin{tcolorbox}[
    colback=white,                  % 【修正】改为 white
    colframe=black,                 % 【修正】改为 black
    boxrule=0.5pt,                 
    arc=2pt,                       
    width=\linewidth,
    boxsep=4pt,
    left=5pt, right=5pt, top=3pt, bottom=3pt
]
\textbf{[Previous solution by Programmer]}\\
$<Technical\;Specification,\;Code,\;Runtime\;Report>$\\
\\
\textbf{[Reviewer comments]}\\                  % 【修正】删除了 \textcolor
$<Advice\;to\;Algorithm\;and\;Debug>$
\end{tcolorbox}

\textbf{[Output Requirement]}\\                 % 【修正】删除了 \textcolor
Generate revised Python code by incorporating reviewer's recommendations to improve solution plan's quality in terms of runtime, structure, and accuracy.
\begin{enumerate}[(i) ]
    \item Provide a concise architectural description of the core algorithm and data flow.
    \item Provide a complete, bug-free Python script.
\end{enumerate}
\textbf{The output code should be enclosed in a single ${```\text{python} ... ```}$ markdown block as follows:} % 【修正】删除了 \textcolor
\begin{tcolorbox}[
    colback=white,                  % 【修正】改为 white
    colframe=black,                 % 【修正】改为 black
    boxrule=0.5pt,                 
    arc=2pt,                       
    width=\linewidth,
    boxsep=4pt,
    left=5pt, right=5pt, top=1pt, bottom=1pt
]
\scriptsize
\begin{verbatim}
# Technical explanation content

```python

# All required imports included at the top.
# Modular structure with clear function definitions.
# Use brief, clear code comments, avoid verbose or tutorial-style explanations.
# Must include quantitative outputs: printed metrics or labeled plots.

'''
\end{verbatim}
\normalsize  
\end{tcolorbox}

\end{tcolorbox}
\end{tcolorbox}

\subsection{Task prompts for the ten benchmark cases}
\label{app:full_tasks}
The prompts below specify the mathematical tasks, constraints, and requested outputs. These specifications are referenced in Sections~\ref{sec:Experiments} and Appendix~\ref{app:pdebench_details} when discussing the corresponding results.

\begin{tcolorbox}[
    title={Problem Description Prompt: Sod Shock Tube},
    coltitle = black,
    colback=white,                      
    colbacktitle=white,            
    colframe=black,                  
    fonttitle=\bfseries\footnotesize,                  % 标题字体加粗
    width=\linewidth,                     % 宽度占满行
    boxsep=6pt,                           % 内边距（调整内容与边框距离）
    top=3pt, bottom=3pt, left=3pt, right=3pt  % 四边边距
]
\footnotesize
The PDEs of 1-D Euler equations are given by:
\[\begin{cases}
\vspace{2mm}
\frac{\partial \rho}{\partial t} + \frac{\partial (\rho u)}{\partial x} = 0,\\
\vspace{2mm}
\frac{\partial (\rho u)}{\partial t} + \frac{\partial (\rho u^2 + p)}{\partial x} = 0, \\
\frac{\partial (\rho E)}{\partial t} + \frac{\partial (\rho E u + p u)}{\partial x} = 0,
\end{cases}\]
where the total energy \(E = \frac{1}{2} u^2 + \frac{p}{(\gamma - 1)\rho}\) and the adiabatic index \(\gamma = 1.4\).\\
At \(t = 0\), the initial conditions in the interval \(x \in [0,1]\) are:
\[(\rho, u, p) = 
\begin{cases}
(1.0, 0.0, 1.0), & 0 < x \leq 0.5, \\
(0.125, 0.0, 0.1), & 0.5 \leq x < 1.
\end{cases}\]
Implement a stable and efficient method to solve this problem.\\
Plot the density \(\rho\), velocity \(u\), and pressure \(p\) at \(t = 0.2\) in one figure.
\end{tcolorbox}

\begin{tcolorbox}[
    title={Problem Description Prompt: Lid-driven flow},
    coltitle = black,
    colback=white,                      
    colbacktitle=white,            
    colframe=black,                  
    fonttitle=\bfseries\footnotesize,                  % 标题字体加粗
    width=\linewidth,                     % 宽度占满行
    boxsep=6pt,                           % 内边距（调整内容与边框距离）
    top=3pt, bottom=3pt, left=3pt, right=3pt  % 四边边距
]
\footnotesize
The PDEs of 2-D steady incompressible Navier-Stokes equations is given by:
\[\begin{cases}
u \frac{\partial u}{\partial x} + v \frac{\partial u}{\partial y} + \frac{\partial p}{\partial x} - \frac{1}{\mathrm{Re}} \left( \frac{\partial^2 u}{\partial x^2} + \frac{\partial^2 u}{\partial y^2} \right) = 0, (x,y) \in \Omega,\\
u \frac{\partial v}{\partial x} + v \frac{\partial v}{\partial y} + \frac{\partial p}{\partial y} - \frac{1}{\mathrm{Re}} \left( \frac{\partial^2 v}{\partial x^2} + \frac{\partial^2 v}{\partial y^2} \right) = 0, (x,y) \in \Omega,\\
\frac{\partial u}{\partial x} + \frac{\partial v}{\partial y} = 0, (x,y) \in \Omega,\\
\end{cases}\]
The Reynolds number Re = 100. The domain is \(\Omega = [0, 1]^2\), and the top boundary is \(\Gamma_1\), the left, right and bottom boundary is \(\Gamma_2\). The boundary conditions are:
\[\begin{cases}
(u,v) = (\alpha(x(1-x)), 0), (x,y) \in \Gamma_1\\
(u,v) = (0,0), (x,y) \in \Gamma_2\\
\text{Reference pressure: } p(x=0,y=0) = 0\\
\text{Zero normal pressure gradient: } \frac{\partial p}{\partial n} = 0, (x,y) \in \partial\Omega
\end{cases}\]
where \(\alpha\) is 2. Implement a stable and efficient method to solve this problem.\\
Plot the contour of the velocity magnitude overlaid with streamlines, and the convergence history in one figure.
\end{tcolorbox}

\begin{tcolorbox}[
    title={Problem Description Prompt: Unsteady Incompressible NS equation},
    coltitle = black,
    colback=white,                      
    colbacktitle=white,            
    colframe=black,                  
    fonttitle=\bfseries\footnotesize,     % 标题字体加粗
    width=\linewidth,                     % 宽度占满行
    boxsep=6pt,                           % 内边距（调整内容与边框距离）
    top=3pt, bottom=3pt, left=3pt, right=3pt  % 四边边距
]
\footnotesize
The governing PDEs for 2-D unsteady incompressible NS equations with explicit forcing are given by:
\[\begin{cases}
\frac{\partial u }{\partial t} + u \frac{\partial u}{\partial x} + v \frac{\partial u}{\partial y} + \frac{\partial p}{\partial x} - \frac{1}{\mathrm{Re}} \left( \frac{\partial^2 u}{\partial x^2} + \frac{\partial^2 u}{\partial y^2} \right) = f_x, (x,y) \in \Omega,\\
\frac{\partial v }{\partial t} + u \frac{\partial v}{\partial x} + v \frac{\partial v}{\partial y} + \frac{\partial p}{\partial y} - \frac{1}{\mathrm{Re}} \left( \frac{\partial^2 v}{\partial x^2} + \frac{\partial^2 v}{\partial y^2} \right) = f_y, (x,y) \in \Omega,\\
\frac{\partial u}{\partial x} + \frac{\partial v}{\partial y} = 0, (x,y) \in \Omega,\\
\end{cases}\]
The Reynolds number Re = 100. The domain is \(\Omega = [0, 2] \times [0, 1]\), and the forcing term \(f(x, y, t)\) is given by:
\[f_x = 0, \quad f_y = -\sin(\pi x) \sin(\pi y) \sin(\pi t).\]
The boundary conditions are:
\[\begin{cases}
(u, v) = (0, 0), \text{ on Top/Bottom walls } (y=0, 1) \\
u(0, y, t) = \sin(\pi y) \left(\sin(\pi t) + \sin(3\pi t) + \sin(5\pi t) \right), v = 0, \text{ at Inlet } (x=0) \\
\frac{\partial u}{\partial x} = \frac{\partial v}{\partial x} = 0, p(2, y, t) = 0, \text{ at Outlet } (x=2)
\end{cases}\]
The initial condition is:
\(u(x, y, 0) = v(x, y, 0) = 0, (x,y) \in \Omega\)\\

Implement a stable and efficient method to solve this problem.\\
Simulate until t=0.5. Plot contours of u, v, and p at the final step in one figure.
\end{tcolorbox}

\begin{tcolorbox}[
    title={Problem Description Prompt: Flow Past a Cylinder},
    coltitle = black,
    colback=white,                      
    colbacktitle=white,            
    colframe=black,                  
    fonttitle=\bfseries\footnotesize,
    width=\linewidth,
    boxsep=6pt,
    top=3pt, bottom=3pt, left=3pt, right=3pt
]
\footnotesize
The governing PDEs for 2-D incompressible flow around a cylinder are given by:
\[\begin{cases}
\frac{\partial u }{\partial t} + u \frac{\partial u}{\partial x} + v \frac{\partial u}{\partial y} + \frac{\partial p}{\partial x} - \frac{1}{\mathrm{Re}} \left( \frac{\partial^2 u}{\partial x^2} + \frac{\partial^2 u}{\partial y^2} \right) = 0, (x,y) \in \Omega,\\
\frac{\partial v }{\partial t} + u \frac{\partial v}{\partial x} + v \frac{\partial v}{\partial y} + \frac{\partial p}{\partial y} - \frac{1}{\mathrm{Re}} \left( \frac{\partial^2 v}{\partial x^2} + \frac{\partial^2 v}{\partial y^2} \right) = 0, (x,y) \in \Omega,\\
\frac{\partial u}{\partial x} + \frac{\partial v}{\partial y} = 0, (x,y) \in \Omega,
\end{cases}\]
The computational domain \(\Omega\) is a rectangle defined by the following distances from the cylinder center:
\begin{itemize}
    \item Cylinder diameter \(d = 1.0\); Distance to left boundary \(x_1 = 3.0\);
    \item Distance to right boundary \(x_2 = 8.0\); Distance to top/bottom boundaries \(y_1 = y_2 = 3.0\).
\end{itemize}
The boundary conditions are:
\[\begin{cases}
\text{Inlet (left): } (u,v) = (1, 0) \text{ Constant inflow} \\
\text{Walls (top/bottom): } \frac{\partial u}{\partial y}=0, v=0 \text{ (Free-slip condition)} \\
\text{Cylinder Surface: No-slip condition} \\
\text{Outlet (right): Zero pressure gradient}
\end{cases}\]
The Reynolds number Re = 100 (Based diameter \(d\) on inlet velocity \(u\)). \\

Implement a stable and efficient method to solve this problem. Simulate until \(t=20.0\).\\
Plot the contour of the velocity magnitude at the final step, and mark the cylinder in the plot. \\

\textbf{[HINTS]:}\\
Use the immersed boundary method with volume penalization on a staggered marker-and-cell grid.\\
Implement the volume penalization term implicitly within the momentum matrix diagonal.\\
Strictly ensure the no-slip condition on the cylinder surface.
\end{tcolorbox}

\begin{tcolorbox}[
    title={Problem Description Prompt: Burgers' Equation},
    coltitle = black,
    colback=white,                      
    colbacktitle=white,            
    colframe=black,                  
    fonttitle=\bfseries\footnotesize,                  % 标题字体加粗
    width=\linewidth,                     % 宽度占满行
    boxsep=6pt,                           % 内边距（调整内容与边框距离）
    top=3pt, bottom=3pt, left=3pt, right=3pt  % 四边边距
]
\footnotesize
The PDE of 1-D Burgers' equation is given by:\\
\[\frac{\partial u}{\partial t} + u u_x - \nu u_{xx} = 0, (x,t) \in \Omega\]\\
where the domain is defined as $(x,y)\in \Omega = [-1,1]\times [0,1]$, the the parameter is $\nu=\frac{0.01}{\pi}$. The initial and boundary conditions are:\\
\[\begin{cases}
u(x,0)= -sin(\pi x)\\
u(-1,t)=u(1,t)=0
\end{cases}\]
Implement a stable and efficient method to solve this problem.\\
Plot the contour of the velocity magnitude and velocity profile in only one figure.
\end{tcolorbox}

\begin{tcolorbox}[
    title={Problem Description Prompt: Poisson Equation},
    coltitle = black,
    colback=white,                      
    colbacktitle=white,            
    colframe=black,                  
    fonttitle=\bfseries\footnotesize,                  % 标题字体加粗
    width=\linewidth,                     % 宽度占满行
    boxsep=6pt,                           % 内边距（调整内容与边框距离）
    top=3pt, bottom=3pt, left=3pt, right=3pt  % 四边边距
]
\footnotesize
The PDE of 2-D Poisson equation is given by:
\[\Delta u = 0, (x,y) \in \Omega\]
The domain is a rectangle minus several circles \(\Omega = \Omega_{\text{rec}} \setminus \Omega_{\text{circle}}\), where \(\Omega_{\text{rec}} = [-0.5, 0.5]^2\) is the rectangle, and the circles \(\Omega_{\text{circle}} = \bigcup_{i = 1}^{4} R_i\) are defined as:
\[\begin{cases}
R_1 = \left\{ (x,y) : (x - 0.3)^2 + (y - 0.3)^2 \leq 0.1^2 \right\} \\
R_2 = \left\{ (x,y) : (x + 0.3)^2 + (y - 0.3)^2 \leq 0.1^2 \right\} \\
R_3 = \left\{ (x,y) : (x - 0.3)^2 + (y + 0.3)^2 \leq 0.1^2 \right\} \\
R_4 = \left\{ (x,y) : (x + 0.3)^2 + (y + 0.3)^2 \leq 0.1^2 \right\} \\
\end{cases}\]
The boundary conditions are:
\[\begin{cases}
u = 0,\, x \in \partial \Omega_{\text{circle}} \\
u = 1,\, x \in \partial \Omega_{\text{rec}} \\
\end{cases}\]
Implement a stable and efficient method to solve this problem.\\
Plot the contour of \(u(x,y)\) in one figure, and mark the circles in the plot.
\end{tcolorbox}

\begin{tcolorbox}[
    title={Problem Description Prompt: Helmholtz Equation},
    coltitle = black,
    colback=white,                      
    colbacktitle=white,            
    colframe=black,                  
    fonttitle=\bfseries\footnotesize,                  % 标题字体加粗
    width=\linewidth,                     % 宽度占满行
    boxsep=6pt,                           % 内边距（调整内容与边框距离）
    top=3pt, bottom=3pt, left=3pt, right=3pt  % 四边边距
]
\footnotesize
The PDE of 2-D Helmholtz equation is given by:
\[-\Delta u + k^2u = f(x,y), (x,y) \in \Omega\]
The function f(x) is defined as:
\[f(x)=A\cdot\left(\sum_{i}\mu_{i}^{2}+x_{i}^{2}\right)\sin(\mu_{1}\pi x_{1})\sin(\mu_{2}\pi x_{2})\]
The parameter values are:
\[\mu_1 = 1,\quad \mu_2 = 4,\quad k = 8,\quad A = 10\]
The domain is a rectangle minus several circles \(\Omega = \Omega_{\text{rec}} \setminus \Omega_{\text{circle}}\), where \(\Omega_{\text{rec}} = [-1, 1]^2\) is the rectangle, and the circles \(\Omega_{\text{circle}} = \bigcup_{i = 1}^{4} R_i\) are defined as:
\[\begin{cases}
R_{1}=\left\{(x,y):(x - 0.5)^{2}+(y - 0.5)^{2}\leq0.2^{2}\right\}\\
R_{2}=\left\{(x,y):(x - 0.4)^{2}+(y + 0.4)^{2}\leq0.4^{2}\right\}\\
R_{3}=\left\{(x,y):(x + 0.2)^{2}+(y + 0.7)^{2}\leq0.1^{2}\right\}\\
R_{4}=\left\{(x,y):(x + 0.6)^{2}+(y - 0.5)^{2}\leq0.3^{2}\right\}\\
\end{cases}\]
The boundary conditions are:
\[\begin{cases}
u = 0.2, & x\in\partial\Omega_{\text{rec}}\\
u = 1, & x\in\partial\Omega_{\text{circle}}
\end{cases}\]
Implement a stable and efficient method to solve this problem.\\
Plot the contour of \(u(x,y)\) in one figure, and mark the circles in the plot.
\end{tcolorbox}

\begin{tcolorbox}[
    title={Problem Description Prompt: Keyhole Dimensional Analysis},
    coltitle = black,
    colback=white,                      
    colbacktitle=white,            
    colframe=black,                   
    fonttitle=\bfseries\footnotesize,                  % 标题字体加粗
    width=\linewidth,                     % 宽度占满行
    boxsep=6pt,                           % 内边距（调整内容与边框距离）
    top=3pt, bottom=3pt, left=3pt, right=3pt,  % 四边边距
]
\footnotesize
Perform dimensional analysis for laser-melted liquid metal to derive the dimensionless form of the keyhole aspect ratio (denoted as \( e^* \)).\\

Please read the CSV data file from path \texttt{./dataset\_keyhole.csv} to perform a data-driven dimensional analysis.\\

The seven physical quantities for the analysis are in the columns named:
the effective laser power \( (\eta P) \), the laser scan speed \( (V_s) \), the laser beam radius \( (r_0) \), the thermal diffusivity \( (\alpha) \), the material density \( (\rho) \), the heat capacity \( (C_p) \), and the difference between melting and ambient temperatures \( (T_l - T_0) \).\\ 

The target variable for fitting is the keyhole aspect ratio \( (e^*) \).\\

Based on dimensional analysis and using the data of these physical quantities in the file, identify the optimal dimensionless quantity formed by combining these parameters, which exhibits the highest coefficient of determination \( (R^2) \) for predicting keyhole aspect ratio.\\

Implement a robust and reliable method from scratch for this data-driven dimensional analysis.\\
Ensure that the resulting exponents of the physical quantities are normalized by \( V_s \) and that the exponents of physical quantities be integers or rational fractions with absolute values not exceeding 3.
\end{tcolorbox}

\begin{tcolorbox}[
    title={Problem Description Prompt: Rayleigh-B\'enard Convection},
    coltitle = black,
    colback=white,                      
    colbacktitle=white,            
    colframe=black,                  
    fonttitle=\bfseries\footnotesize,                  
    width=\linewidth,                     
    boxsep=6pt,                           
    top=3pt, bottom=3pt, left=3pt, right=3pt,  
]
\footnotesize
Perform dimensional analysis for Rayleigh-B\'enard convection in a planar horizontal layer of fluid heated from below, to derive the dimensionless form of the Nusselt number (denoted as \( Nu \)).\\

Please read the CSV data file from path \texttt{./dataset\_rayben.csv} to perform a data-driven dimensional analysis.\\

The seven physical quantities for the analysis are in the columns named:
the height of the container \((h)\), the thermal conductivity \((\lambda)\), the temperature difference between the top and bottom surfaces \((\Delta T)\), the gravitational acceleration \((g)\), the thermal expansion coefficient \((\alpha)\), the kinematic viscosity \((\nu)\), and the thermal diffusivity \((\kappa)\).\\

The target variable for fitting is the Nusselt number \((Nu)\).\\

Based on dimensional analysis and using the data of these physical quantities in the file, identify the optimal dimensionless quantity formed by combining these parameters, which exhibits the highest coefficient of determination \( (R^2) \) for predicting Nu.\\

Implement a robust and reliable method from scratch for this data-driven dimensional analysis.\\
Ensure that the resulting exponents of the physical quantities are normalized by \(\alpha\) and that the exponents of physical quantities be integers or rational fractions with absolute values not exceeding 3.
\end{tcolorbox}

\begin{tcolorbox}[
    title={Problem Description Prompt: Hilbert Matrix},
    coltitle = black,
    colback=white,                      
    colbacktitle=white,            
    colframe=black,
    fonttitle=\bfseries\footnotesize,                  % 标题字体加粗
    width=\linewidth,                     % 宽度占满行
    boxsep=6pt,                           % 内边距（调整内容与边框距离）
    top=3pt, bottom=3pt, left=3pt, right=3pt  % 四边边距
]
\footnotesize
\textbf{Implement various appropriate methods from scratch} to solve the system of linear algebra equations \( H_{n}\mathbf{x} = \mathbf{b} \) accurately and efficiently, where \( H_{n} \) is an \( n \times n \) matrix defined by \( h_{ij} = 1/(i+j-1) \), and the vector \(\mathbf{b}\) is taken to ensure that the system admits an exact solution \( \mathbf{x^*} = (x_{i})_{n \times 1} = (1)_{n \times 1} \).\\
Compare the \( L_{\infty} \) error of the numerical results with the exact solution \( \mathbf{x^*} \) for \( n = 5, 10, 15, 20, 25 \).
\end{tcolorbox}

\section{Supplementary numerical results and comparative analysis}
\label{app:supplementary_results}

\subsection{Detailed results for the PDE benchmarks}
\label{app:pdebench_details}

\begin{figure}[!b]
    \centering

    % Top: execution success rate
    \begin{subfigure}{0.86\textwidth}
        \centering
        \includegraphics[width=0.60\textwidth]{LT_NS_Exe_Success_Rate_legend.pdf}\\[-2mm]
        \includegraphics[width=0.9\textwidth]{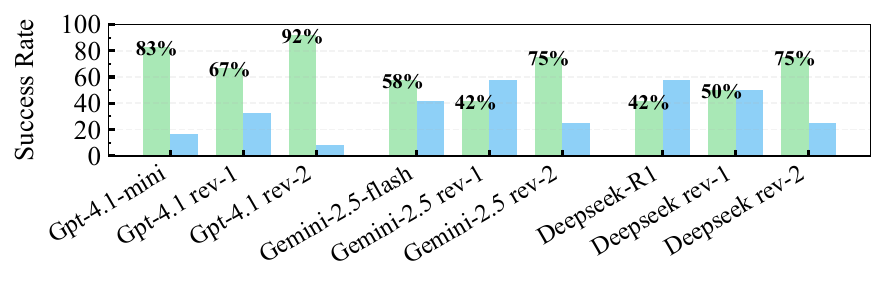}
    \end{subfigure}

    % Middle: error distributions
    \begin{subfigure}{0.86\textwidth}
        \centering
        \includegraphics[width=0.9\textwidth]{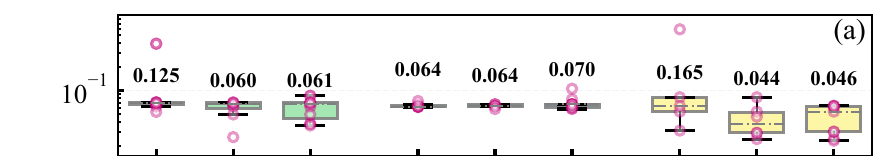}\\
        \includegraphics[width=0.9\textwidth]{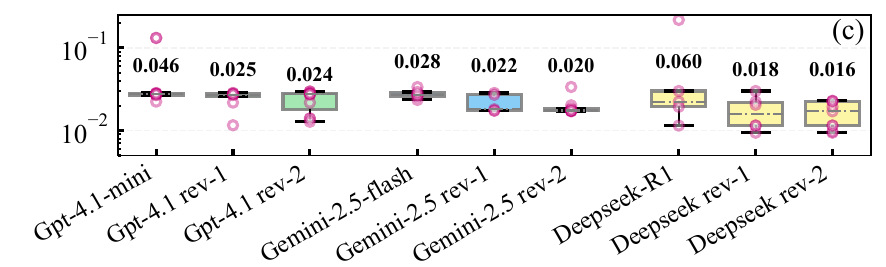}\\
        \includegraphics[width=0.9\textwidth]{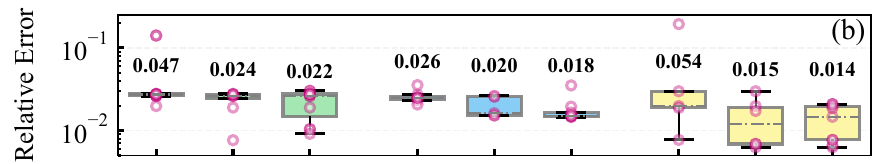}
    \end{subfigure}

    % Bottom: best final-stage outputs
    \begin{subfigure}{0.86\textwidth}
        \centering
        \includegraphics[width=0.92\textwidth]{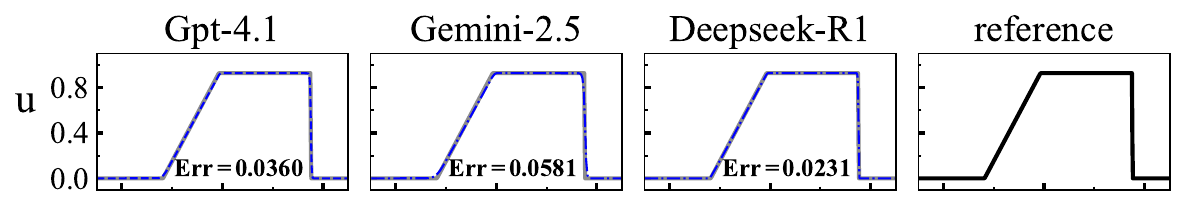}\\
        \includegraphics[width=0.92\textwidth]{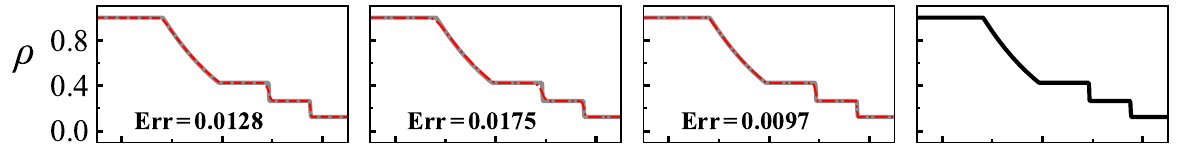}\\
        \includegraphics[width=0.92\textwidth]{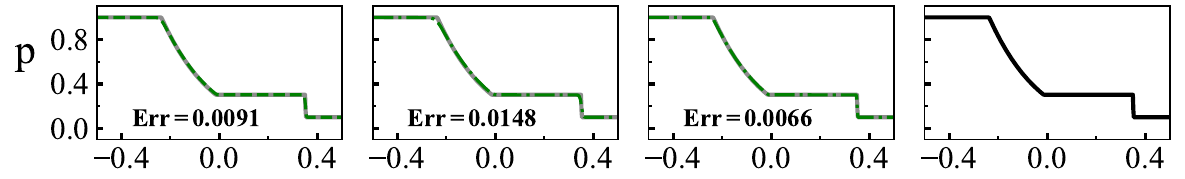}
    \end{subfigure}
\caption{
    Detailed results for the Sod shock tube problem.
    In the middle panel, (a) velocity $u$, (b) density $\rho$, and (c) pressure $p$.
    }
    \label{fig:Sod_detailed_results}
\end{figure}

\begin{figure}[!ht]
    \centering

    % Top: execution success rate
    \begin{subfigure}{0.86\textwidth}
        \centering
        \includegraphics[width=0.60\textwidth]{LT_NS_Exe_Success_Rate_legend.pdf}\\[-2mm]
        \includegraphics[width=0.9\textwidth]{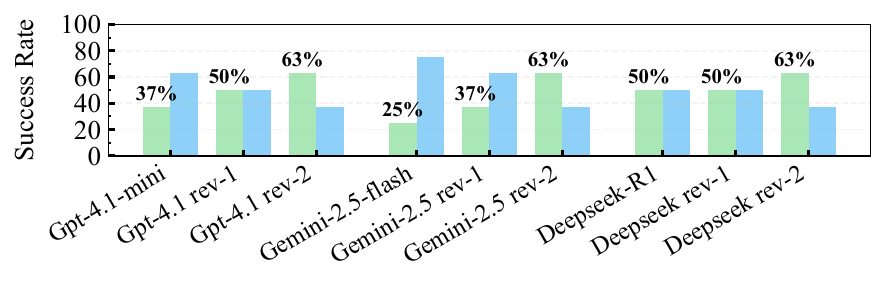}
    \end{subfigure}

    % Middle: error distribution
    \begin{subfigure}{0.86\textwidth}
        \centering
        \includegraphics[width=0.9\textwidth]{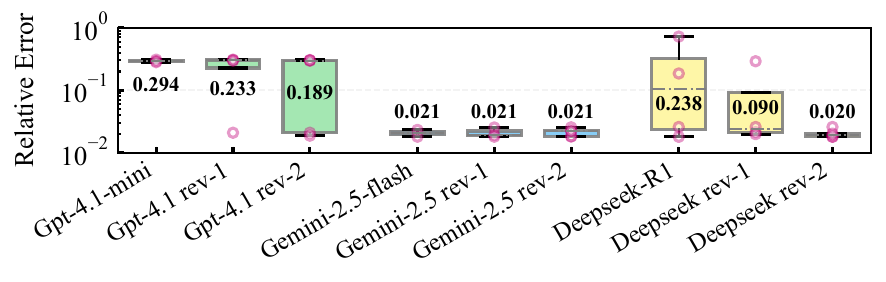}
    \end{subfigure}

    % Bottom: best final-stage outputs
    \begin{subfigure}{0.86\textwidth}
        \centering
        \includegraphics[width=0.99\textwidth]{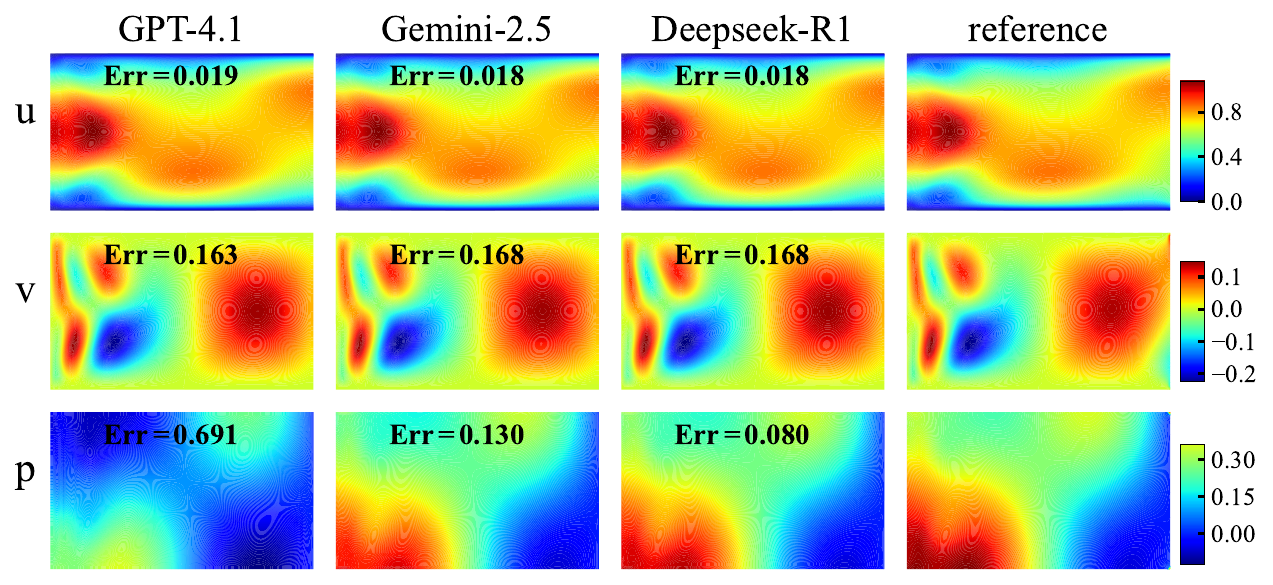}
    \end{subfigure}

    \caption{
    Detailed results for the unsteady NS equation.
    }
    \label{fig:UnsteadyNS_detailed_results}
\end{figure}

\begin{figure}[!ht]
    \centering

    % Top: execution success rate
    \begin{subfigure}{0.86\textwidth}
        \centering
        \includegraphics[width=0.60\textwidth]{LT_NS_Exe_Success_Rate_legend.pdf}\\[-2mm]
        \includegraphics[width=0.9\textwidth]{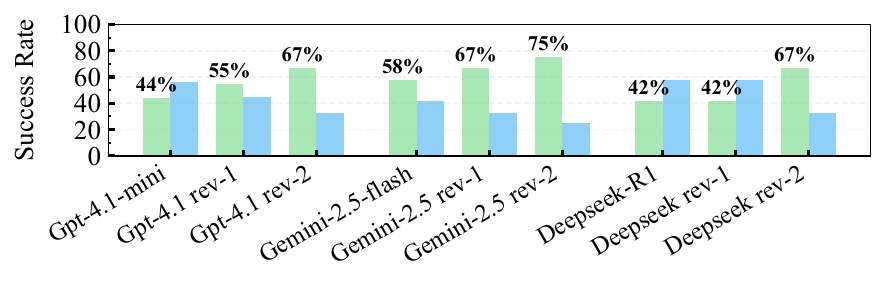}
    \end{subfigure}

    % Middle: error distribution
    \begin{subfigure}{0.86\textwidth}
        \centering
        \includegraphics[width=0.9\textwidth]{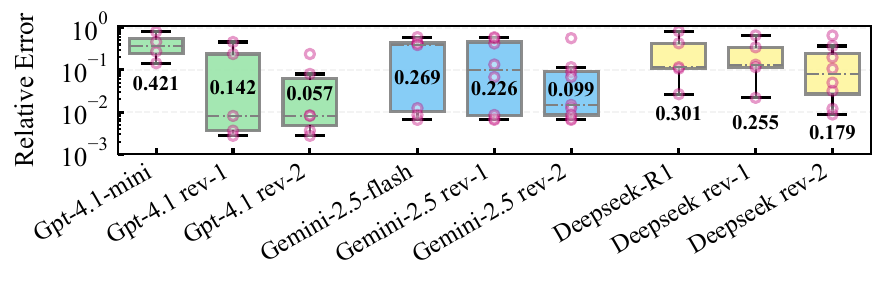}
    \end{subfigure}

    % Bottom: best final-stage outputs
    \begin{subfigure}{0.86\textwidth}
        \centering
        \includegraphics[width=0.99\textwidth]{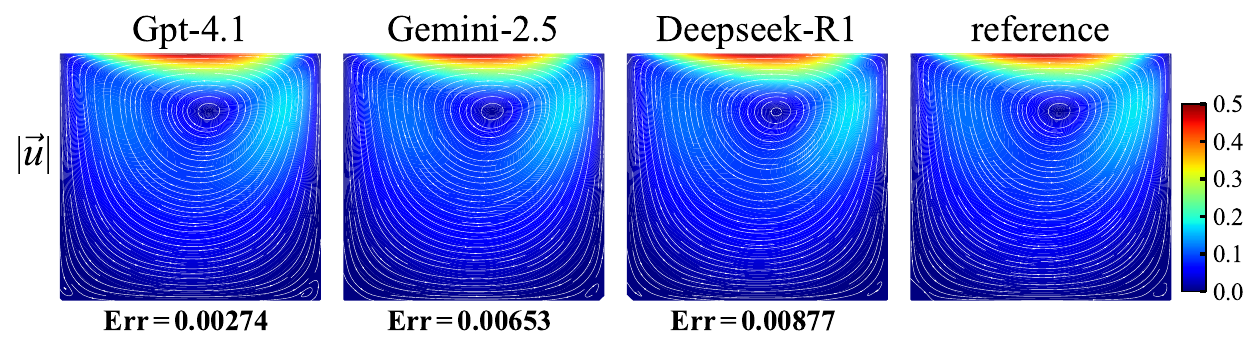}
    \end{subfigure}

    \caption{
    Detailed results for the lid-driven cavity flow.
    }
    \label{fig:LidDriven_detailed_results}
\end{figure}

\begin{figure}[!ht]
    \centering

    % Top: execution success rate
    \begin{subfigure}{0.86\textwidth}
        \centering
        \includegraphics[width=0.60\textwidth]{LT_NS_Exe_Success_Rate_legend.pdf}\\[-2mm]
        \includegraphics[width=0.9\textwidth]{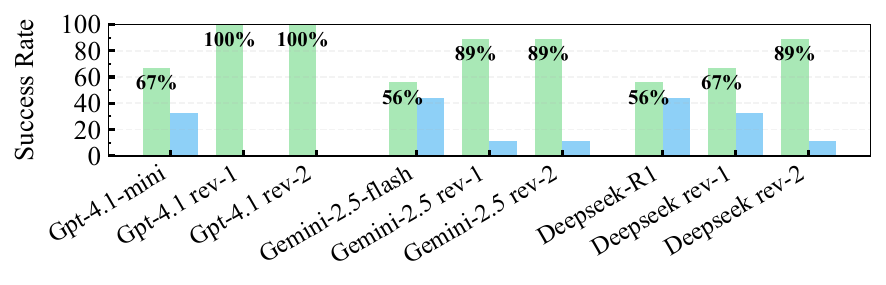}
    \end{subfigure}

    % Middle: error distribution
    \begin{subfigure}{0.86\textwidth}
        \centering
        \includegraphics[width=0.9\textwidth]{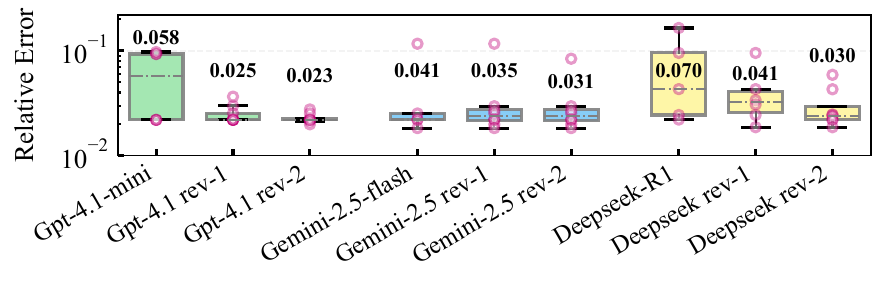}
    \end{subfigure}

    % Bottom: best final-stage outputs
    \begin{subfigure}{0.86\textwidth}
        \centering
        \includegraphics[width=0.99\textwidth]{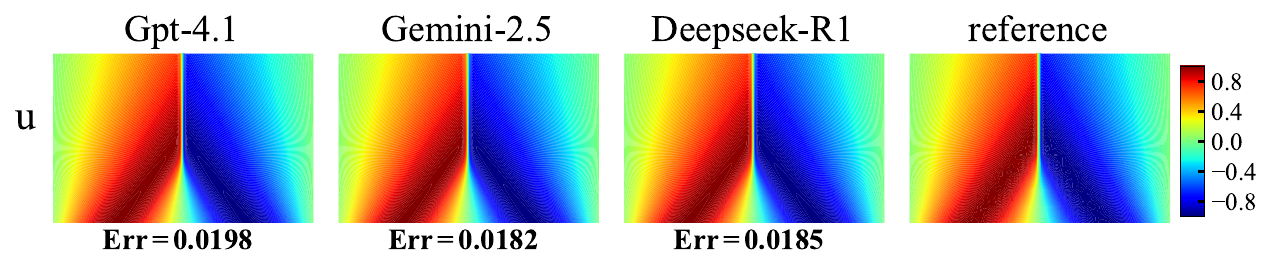}
    \end{subfigure}

    \caption{
    Detailed results for Burgers' equation.
    }
    \label{fig:Burgers_detailed_results}
\end{figure}

\begin{figure}[!ht]
    \centering

    % Top: execution success rate
    \begin{subfigure}{0.85\textwidth}
        \centering
        \includegraphics[width=0.60\textwidth]{LT_NS_Exe_Success_Rate_legend.pdf}\\[-2mm]
        \includegraphics[width=0.9\textwidth]{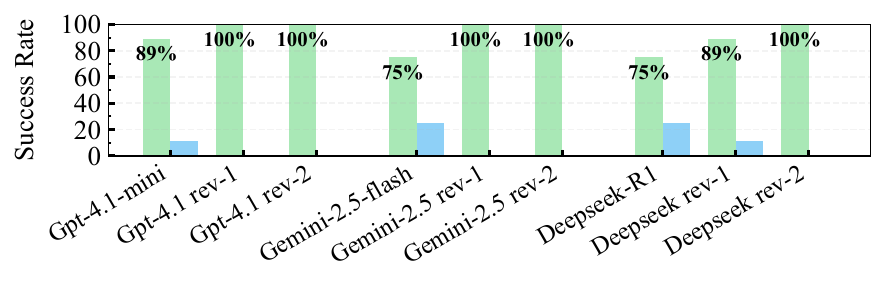}
    \end{subfigure}

    % Middle: error distribution
    \begin{subfigure}{0.85\textwidth}
        \centering
        \includegraphics[width=0.9\textwidth]{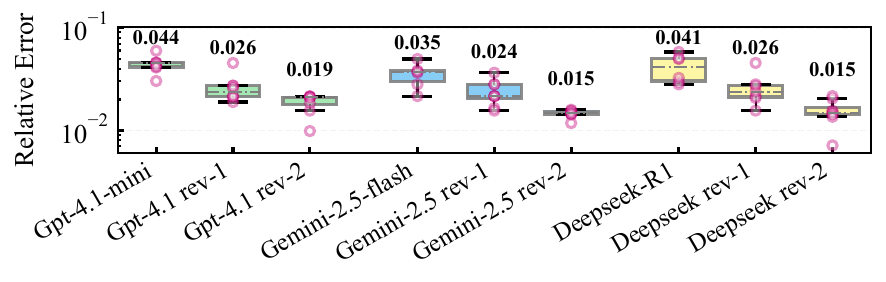}
    \end{subfigure}

    % Bottom: best final-stage outputs
    \begin{subfigure}{0.85\textwidth}
        \centering
        \includegraphics[width=0.99\textwidth]{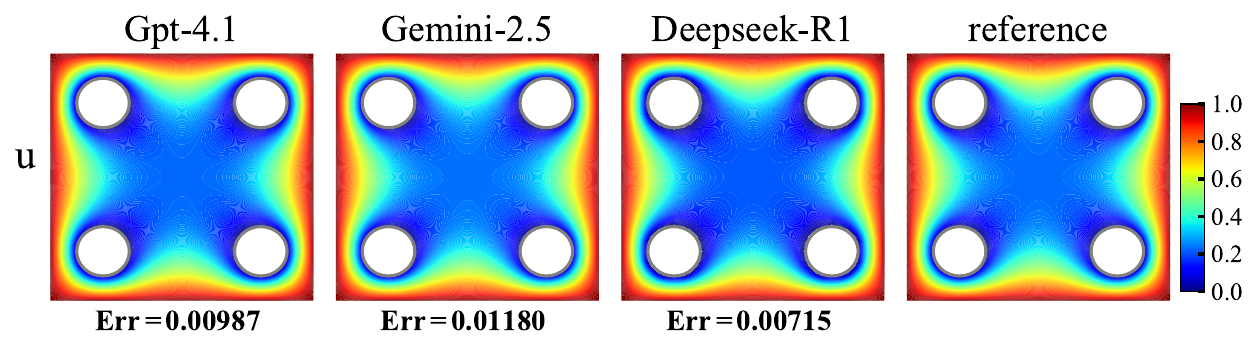}
    \end{subfigure}

    \caption{
    Detailed results for the Poisson equation.
    }
    \label{fig:Poisson_detailed_results}
\end{figure}

\begin{figure}[!ht]
    \centering

    % Top: execution success rate
    \begin{subfigure}{0.85\textwidth}
        \centering
        \includegraphics[width=0.60\textwidth]{LT_NS_Exe_Success_Rate_legend.pdf}\\[-2mm]
        \includegraphics[width=0.9\textwidth]{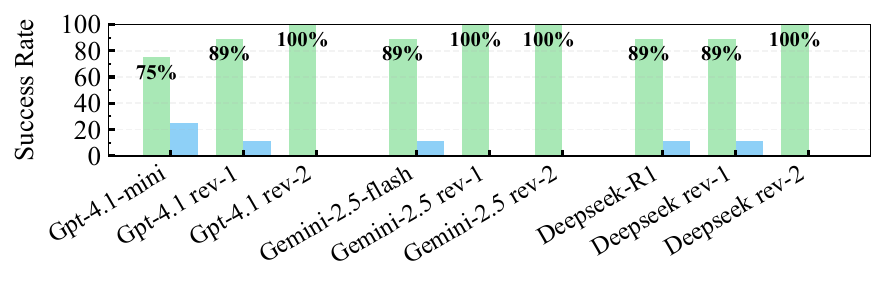}
    \end{subfigure}

    % Middle: error distribution
    \begin{subfigure}{0.85\textwidth}
        \centering
        \includegraphics[width=0.9\textwidth]{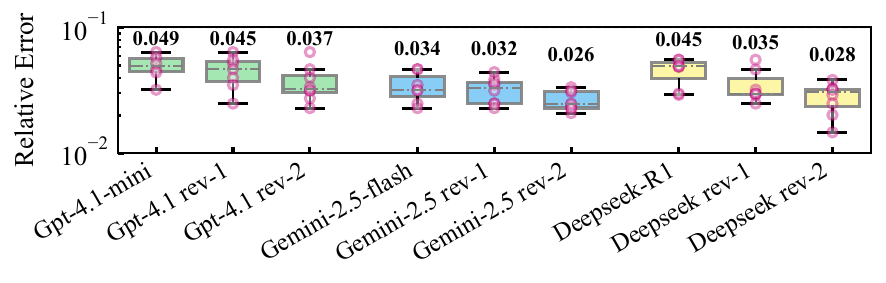}
    \end{subfigure}

    % Bottom: best final-stage outputs
    \begin{subfigure}{0.85\textwidth}
        \centering
        \includegraphics[width=0.99\textwidth]{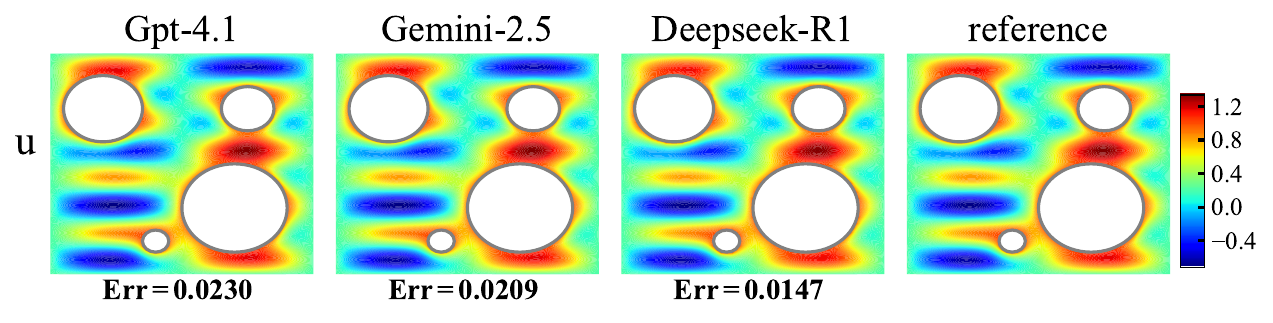}
    \end{subfigure}

    \caption{
    Detailed results for the Helmholtz equation.
    }
    \label{fig:Helmholtz_detailed_results}
\end{figure}

The six PDE benchmarks are presented using a common three-part layout. In each figure, the top panel reports the execution success rates at \texttt{ans-0}, \texttt{rev-1}, and \texttt{rev-2}; the middle panel shows the relative $L^2$ error distributions among successful executions, with annotated values indicating the corresponding mean errors; and the bottom panel presents the best executable \texttt{rev-2} output produced by each Programmer together with the reference solution. The complete task prompts are provided in Appendix~\ref{app:full_tasks}.

The \textbf{Sod shock tube} problem is a one-dimensional compressible Euler benchmark characterized by a rarefaction wave, a contact discontinuity, and a shock wave. \textcolor{black}{Figure~\ref{fig:Sod_detailed_results} shows that the code execution success rates of GPT-4.1-mini and Gemini-2.5-flash each decline at the \texttt{rev-1} stage and subsequently recover at \texttt{rev-2}. The slight increase in mean relative $L^2$ error for velocity $u$ from \texttt{rev-1} to \texttt{rev-2} reflects the inclusion of less accurate solutions from newly executable runs, despite the higher execution success rate.}

\textcolor{black}{The \textbf{unsteady NS equation} describes time-dependent incompressible flow. Figure~\ref{fig:UnsteadyNS_detailed_results} shows that GPT-4.1-mini's mean relative error remains relatively high at \texttt{rev-2}. Several independent runs become executable only at this stage. In some of these runs, deficiencies in the numerical treatment of the pressure Poisson equation can produce relatively large velocity divergence and non-physical velocity fields, contributing to the relatively high final-stage mean error despite successful execution. Among the best-of-$N$ final-stage results, the three Programmers produce comparable streamwise-velocity fields, while larger deviations in the normal velocity $v$ remain primarily near the outlet. The pressure fields vary more substantially across Programmers than the streamwise-velocity fields.}

The \textbf{lid-driven cavity} problem represents a steady incompressible Navier--Stokes flow in which nonlinear convection and pressure--velocity coupling must be treated consistently. \textcolor{black}{Figure~\ref{fig:LidDriven_detailed_results} shows increasing execution success rates through the review--revision process for all three Programmers. DeepSeek-R1's mean relative $L^2$ error remains relatively large at \texttt{rev-2} ($0.179$). This partly reflects two factors: the inclusion of solutions with relatively large errors from newly executable runs, and incomplete steady-state convergence due to insufficient pseudo-time iterations in some runs.}

The \textbf{Burgers' equation} benchmark in figure~\ref{fig:Burgers_detailed_results} couples nonlinear convection with viscous diffusion~\citep{benton1972table}. When the viscosity coefficient $\nu$ is small, convection becomes dominant and an initially smooth velocity profile can develop steep, shock-like gradients.

The \textbf{Poisson equation} is a prototypical elliptic partial differential equation~\citep{strauss2007partial}. The benchmark in figure~\ref{fig:Poisson_detailed_results} uses a multiply connected domain with a rectangular outer boundary and four circular holes, requiring consistent treatment of both outer and internal boundaries.

The \textbf{Helmholtz equation} is a time-independent wave equation that commonly arises after separation of variables in wave problems~\citep{strauss2007partial}. The results in figure~\ref{fig:Helmholtz_detailed_results} are obtained on a multiply connected rectangular domain containing four circular holes with prescribed boundary conditions.

\subsection{Detailed results for the Hilbert linear systems}
\label{app:hilbert_details}

\begin{figure}[!t]
\centering
\begin{subfigure}[b]{0.65\textwidth}
\centering
{\includegraphics[width=\textwidth]{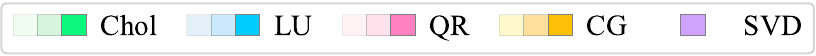}}
\end{subfigure}
\begin{subfigure}[b]{0.85\textwidth}
\centering
{\includegraphics[width=\textwidth]{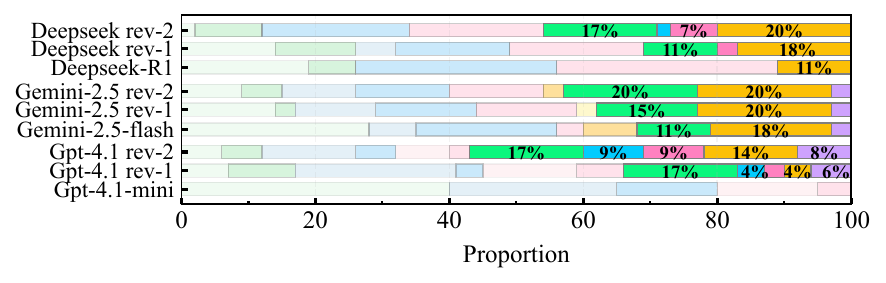}}
\end{subfigure}
  \caption{Proportional distribution of various algorithms employed by Programmers at different response stages. Percentages reflect the adoption frequency of each method across all independent runs. For each method, the bar is divided into three shaded segments based on completion status: the lightest shade (\textcolor{cyan!10}{$\blacksquare$}) represents results containing NaNs, the medium shade (\textcolor{cyan!35}{$\blacksquare$}) denotes results exceeding the $L^{\infty}$ threshold ($10^{-2}$), and the darkest shade (\textcolor{cyan!80}{$\blacksquare$}) indicates results below the $L^{\infty}$ threshold.}
\label{fig:hil_methodDist_3status}
\end{figure}

\begin{figure}[!t]
\centering
\begin{subfigure}[b]{0.8\textwidth}
\centering
{\includegraphics[width=\textwidth]{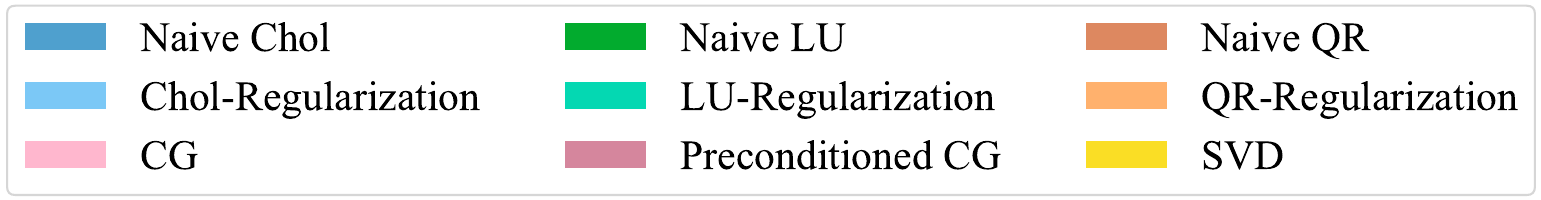}}
\end{subfigure}
\begin{subfigure}[b]{0.85\textwidth}
\centering
{\includegraphics[width=\textwidth]{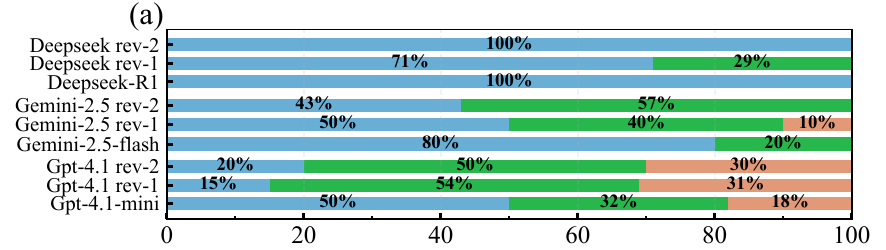}}
\end{subfigure}
\begin{subfigure}[b]{0.85\textwidth}
\centering
{\includegraphics[width=\textwidth]{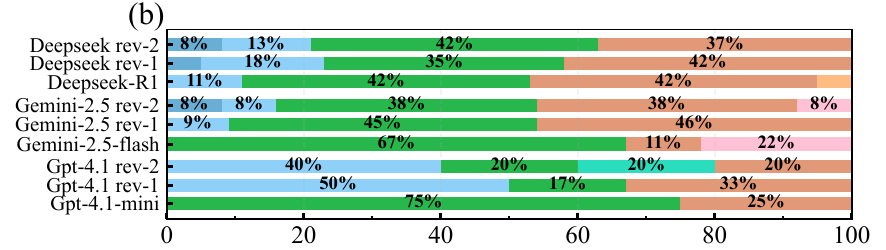}}
\end{subfigure}
\begin{subfigure}[b]{0.85\textwidth}
\centering
{\includegraphics[width=\textwidth]{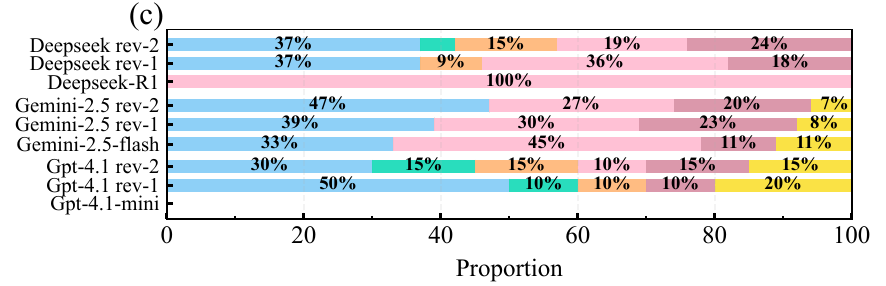}}
\end{subfigure}
\caption{Proportional distribution of specific numerical methods employed by Programmers across three completion statuses, serving as a detailed elaboration to the solution success rate. (a) Results Contain NaN; (b) Over $L^\infty$ threshold; (c) Below $L^\infty$ threshold.}
\label{fig:Hil_diff_stage_proportion}
\end{figure}

{\color{black}
Figure~\ref{fig:hil_methodDist_3status} presents the distribution of numerical methods across response stages. Since each generated script incorporates and compares multiple methods, the percentages represent shares of method implementations pooled across all independent runs for each Programmer and response stage. Cholesky-based and conjugate-gradient methods account for a large share of these implementations, and the proportion meeting the accuracy criterion increases across response stages.

Figure~\ref{fig:Hil_diff_stage_proportion} shows the method composition within each completion status. Below-threshold results mainly comprise implementations using regularization, including Tikhonov regularization~\citep{golub1999tikhonov}, and preconditioned conjugate-gradient methods~\citep{hestenes1952methods}, whereas naive factorization methods predominantly produce NaNs or errors exceeding the threshold. Unpreconditioned conjugate-gradient methods and SVD-based solvers also contribute to the successful results. Regularized Cholesky implementations occur in both the above- and below-threshold groups.

Table~\ref{tab:hil_Linf_error} lists the final-stage (\texttt{rev-2}) $L^\infty$ errors of the best-performing methods for each tested matrix dimension. From the best-of-$N$ perspective, the results show an order-of-magnitude reduction in $L^\infty$ errors at larger matrix dimensions compared with the naive methods in figure~\ref{fig:hil_naive_cond}.}

\begin{table}[!t]
\centering
\scriptsize
\renewcommand{\arraystretch}{1.2}
\begin{tabular}{c|c|ccccc}
\cline{1-7}
\multirow{2}{*}{Programmer} & \multirow{2}{*}{Methods}  & \multicolumn{5}{c}{$L^\infty$ error w.r.t. matrix size $n$} \\  \cline{3-7}
& & 5&10&15&20&25\\ \cline{1-7}
\multirow{5}{*}{\makecell{GPT-4.1-mini}} & Chol-Reg &  3.08e-04&6.81e-04&9.66e-04&1.36e-03&1.36e-03\\
% \\Answer-0 20\\Review-1 30\\Review-2 35
&  LU-Reg & 3.15e-04&8.77e-04&8.92e-04&1.38e-03&1.37e-03\\
&  QR-Reg & 3.20e-04&9.90e-04&1.18e-03&1.31e-03&1.62e-03\\
&  Pre-CG & \textbf{9.80e-12}&8.79e-05&7.16e-05&\textbf{4.32e-05}&\textbf{9.96e-05}\\
&  SVD & 1.46e-11&7.32e-04&3.91e-03&7.81e-03&7.32e-04\\ \cline{1-7}
\multirow{4}{*}{\makecell{Gemini-2.5-flash}} & Chol-Reg & 8.64e-04&1.12e-03&1.45e-03&1.61e-03&1.73e-03 \\
% \\Answer-0 28\\Review-1 34\\Review-2 35
% & Chol-Iter &3.26e-03&3.08e-03&2.99e-03&3.86e-03&4.01e-03\\ \cline{2-7}
& CG & 4.49e-04&9.34e-04&4.85e-04&6.60e-04&4.23e-04\\
& Pre-CG & 8.44e-04&5.57e-04&5.19e-04&7.50e-04&7.41e-04\\
& SVD & 7.13e-04&4.76e-04&5.36e-04&5.02e-04&6.30e-04 \\ \cline{1-7}
\multirow{5}{*}{\makecell{Deepseek-R1}} &Chol-Reg & 1.11e-09&\textbf{3.84e-05}&\textbf{5.05e-05}&1.45e-04&2.68e-04\\
% \\Answer-0 27\\Review-1 35\\Review-2 46
& LU-Reg & 1.11e-09&8.61e-05&1.81e-04&1.51e-04&2.01e-04\\
& QR-Reg & 3.80e-03&6.07e-04&1.01e-03&1.10e-03&1.31e-03\\
& CG & 5.42e-11&5.99e-04&3.93e-04&2.09e-04&4.30e-04\\
& Pre-CG & \textbf{9.80e-12}&6.01e-04&3.95e-04&2.08e-04&4.32e-04\\ \cline{1-7}
\end{tabular}
\caption{The summary of $L^\infty$ errors in solving linear systems with Hilbert matrices via various methods provided by each Programmer in the final response (\texttt{rev-2}). The \textbf{bold} entries denote the minimum error achieved for each matrix size $n$}
\label{tab:hil_Linf_error}
\end{table}

\subsection{\label{subapp:code_comparison}Code-level algorithmic analysis: CodePDE vs. Re4}

\begin{table}[!t]
\centering
\footnotesize % 采用 footnotesize，排版更加精致紧凑
\renewcommand{\arraystretch}{1.3} % 稍微加宽行高
\caption{Code-level algorithmic comparison of the generated 1-D Burgers' solvers between CodePDE and different Programmer LLMs under Re4.}
\label{tab:code_comparison}
% 使用 >{\raggedright\arraybackslash}p{...} 确保单元格文字靠左对齐，彻底消除多余的空格拉伸
\begin{tabular}{>{\raggedright\arraybackslash}p{2.6cm}|>{\raggedright\arraybackslash}p{2.4cm}>{\raggedright\arraybackslash}p{2.7cm}>{\raggedright\arraybackslash}p{2.7cm}>{\raggedright\arraybackslash}p{2.7cm}}
\hline
\textbf{Metric} & \textbf{CodePDE} & \textbf{Re4 (GPT)} & \textbf{Re4 (Gemini)} & \textbf{Re4 (R1)} \\ \hline
\textbf{Boundary Cond.} & Periodic ($x \in [0, 1]$) & Dirichlet ($x \in [-1, 1]$) & Dirichlet ($x \in [-1, 1]$) & Dirichlet ($x \in [-1, 1]$) \\ \hline
\textbf{Convection} & 2nd-order Central Diff. / Fourier Spectral & 1st-order Upwind & 1st-order Upwind & 2nd-order MUSCL (Minmod) \\ \hline
\textbf{Diffusion} & 2nd-order Central Diff. / Fourier Spectral & 2nd-order Central Diff. & 2nd-order Central Diff. & 2nd-order Central Diff. \\ \hline
\textbf{Time Integration} & 1st-order Explicit Euler / IMEX & Crank-Nicolson & Crank-Nicolson & IMEX SSP-RK3 \\ \hline
\textbf{Linear Solver} & Standard tensor operations & Banded matrix solve (\texttt{solve\_banded}) & Sparse matrix solve (\texttt{spsolve}) & LU pre-factorization (\texttt{splu}) \\ \hline
\end{tabular}
\end{table}

As previously discussed in Section~\ref{subsec:cost_benefit_compare}, existing LLM-based scientific computing agents exhibit variations in their target problem domains and underlying workflow designs, with CodePDE standing as the most directly comparable and relevant baseline for generating from-scratch PDE solvers. Beyond these workflow design discrepancies, CodePDE also differs significantly from our framework in its physical problem configurations. Specifically, taking the 1-D Burgers' equation as an illustrative example, the target problems of the two frameworks present the following differences:
\begin{itemize}
    \item \textbf{Setup of CodePDE:} Solves the Burgers' equation on a periodic domain $x \in [0, 1]$. Because of the periodic boundary assumption, the generated code can bypass complex boundary treatments by relying on Fast Fourier Transforms (FFT) or vector rolling (\texttt{np.roll}). For time integration, the generated code frequently defaults to a 1st-order explicit forward Euler scheme, which is typically unstable for purely advective transport and under low viscosity.
    \item \textbf{Setup of Re4:} Solves the equation on a Dirichlet domain $x \in [-1, 1]$ with $u(-1, t) = u(1, t) = 0$. Here, the non-linear convective term drives the initial sinusoidal profile to steepen into a sharp shock at $x = 0$ over time. Resolving this Dirichlet boundary while preventing numerical instability near the shock requires consistent boundary treatments and stable discretization schemes.
\end{itemize}

Given these discrepancies in workflow designs and problem configurations between the proposed Re4 agent and CodePDE, directly comparing quantitative error metrics on identical benchmarks is neither technically equivalent nor conceptually fair. Therefore, in table~\ref{tab:code_comparison}, we present a detailed code-level qualitative comparison to highlight the practical robustness and discretization strategies of the solvers produced by our Re4 framework (evaluated at the final \texttt{rev-2} stage across different Programmer LLMs).

CodePDE relies on massive test-time sampling (generating up to 32 candidates) during execution-driven debugging and uses reference-based error metrics in the subsequent refinement stage. The inspected Burgers' implementations include basic numerical schemes, such as second-order central differences combined with first-order explicit forward Euler time integration. Such setups are susceptible to non-physical oscillations near shock regions and can require small time steps to maintain stability.

In contrast, the Re4 feedback loop guides the agent to design stable numerical methods. As summarized in table~\ref{tab:code_comparison}, both GPT-4.1-mini and Gemini-2.5-flash implement 1st-order upwind schemes for the non-linear convection term to ensure physical windward propagation, coupled with stable implicit Crank-Nicolson schemes for the diffusion term, leveraging optimized banded matrix solvers (\texttt{solve\_banded}) and sparse solvers (\texttt{spsolve}). 
Notably, DeepSeek-R1 achieves a more robust solver by implementing a spatiotemporally high-order algorithm. It utilizes a 2nd-order MUSCL (Monotonic Upstream-centered Scheme for Conservation Laws) spatial reconstruction paired with a Minmod flux limiter to sharply capture the shock at $x=0$. Furthermore, it integrates an explicit-implicit Strong Stability Preserving Runge-Kutta 3rd-order (IMEX SSP-RK3) scheme, and caches the LU decomposition (\texttt{splu}) of the implicit Crank-Nicolson system matrix to maximize the computational efficiency of the implicit diffusion steps. This algorithmic comparison underscores that the proposed Re4 agent not only generates bug-free code but effectively elevates the physical reliability and numerical sophistication of the generated solvers.

\section{Ablation results and supporting evidence}\label{app:resources}

\subsection{Stage-wise ablation and API-usage statistics}
\label{app:ablation_records}
Table~\ref{tab:comprehensive_ablation_cost} reports the stage-wise success rates, input/output token counts, and Programmer response times for the tasks included in the ablation study. The corresponding results for the cylinder-flow case are reported separately in table~\ref{tab:cylinder_token_rate}. The reported response times refer to Programmer calls and are distinct from solver execution time and total pipeline API cost.

\begingroup
\footnotesize
\setlength{\tabcolsep}{3pt}
\begin{longtable}{l|l|l|l|cc|c}
\caption{Comprehensive ablation study and cost analysis of the proposed Re4 framework across different reasoning stages for all nine evaluated tasks. The arrows denote the individual contributions of the Consultant ($\uparrow$) and Reviewer ($\uparrow$/$\downarrow$) modules to the success rate.}
\label{tab:comprehensive_ablation_cost} \\
\hline
\multirow{2}{*}{\textbf{Task}} & \multirow{2}{*}{\textbf{Programmer}} & \multirow{2}{*}{\textbf{Stage}} & \multirow{2}{*}{\textbf{Success Rate}} & \multicolumn{2}{c|}{\textbf{Tokens (k)}} & \multirow{2}{*}{\textbf{Time (s)}} \\ \cline{5-6}
& & & & \textbf{in} & \textbf{out} & \\ \hline
\endfirsthead

\multicolumn{7}{c}{\tablename\ \thetable\ -- Continued from previous page} \\ \hline
\multirow{2}{*}{\textbf{Task}} & \multirow{2}{*}{\textbf{Programmer}} & \multirow{2}{*}{\textbf{Stage}} & \multirow{2}{*}{\textbf{Success Rate}} & \multicolumn{2}{c|}{\textbf{Tokens (k)}} & \multirow{2}{*}{\textbf{Time (s)}} \\ \cline{5-6}
& & & & \textbf{in} & \textbf{out} & \\ \cline{2-7}
\endhead

\cline{3-7} \multicolumn{7}{r}{Continued on next page} \\
\endfoot
\hline
\endlastfoot

\multirow{12}{*}{\textbf{Burgers}} & \multirow{4}{*}{GPT-4.1-mini} 
  & baseline & 50.0\% & 0.51 & 1.59 & 18.85 \\
& & ans-0 & 67.0\% ~($\uparrow$ 17.0\%) & 1.52 & 1.88 & 23.26 \\
& & rev-1 & 100\% ~($\uparrow$ 33.0\%) & 2.60 & 1.97 & 25.01 \\
& & rev-2 & 100\% ~($\uparrow$ 33.0\%) & 2.72 & 2.24 & 25.92 \\ \cline{2-7}
& \multirow{4}{*}{Gemini-2.5-flash} 
  & baseline & 50.0\% & 0.51 & 2.10 & 18.65 \\
& & ans-0 & 56.0\% ~($\uparrow$ 6.0\%) & 1.52 & 2.95 & 24.58 \\
& & rev-1 & 89.0\% ~($\uparrow$ 33.0\%) & 3.64 & 3.52 & 28.88 \\
& & rev-2 & 89.0\% ~($\uparrow$ 33.0\%) & 4.22 & 4.43 & 34.79 \\ \cline{2-7}
& \multirow{4}{*}{DeepSeek-R1} 
  & baseline & 50.0\% & 0.51 & 1.35 & 68.51 \\
& & ans-0 & 56.0\% ~($\uparrow$ 6.0\%) & 1.52 & 1.52 & 70.82 \\
& & rev-1 & 67.0\% ~($\uparrow$ 11.0\%) & 2.23 & 1.88 & 78.56 \\
& & rev-2 & 89.0\% ~($\uparrow$ 33.0\%) & 2.59 & 2.11 & 81.32 \\ \hline

\multirow{12}{*}{\textbf{Sod}} & \multirow{4}{*}{GPT-4.1-mini} 
  & baseline & 38.0\% & 0.61 & 2.18 & 22.89 \\
& & ans-0 & 83.0\% ~($\uparrow$ 45.0\%) & 1.69 & 2.43 & 26.30 \\
& & rev-1 & 67.0\% ~($\downarrow$ 16.0\%) & 3.23 & 2.97 & 31.50 \\
& & rev-2 & 92.0\% ~($\uparrow$ 9.0\%) & 3.68 & 3.12 & 34.10 \\ \cline{2-7}
& \multirow{4}{*}{Gemini-2.5-flash} 
  & baseline & 33.0\% & 0.61 & 3.15 & 24.47 \\
& & ans-0 & 58.0\% ~($\uparrow$ 25.0\%) & 1.69 & 3.98 & 29.87 \\
& & rev-1 & 42.0\% ~($\downarrow$ 16.0\%) & 4.69 & 4.80 & 36.20 \\
& & rev-2 & 75.0\% ~($\uparrow$ 17.0\%) & 5.37 & 5.87 & 43.15 \\ \cline{2-7}
& \multirow{4}{*}{DeepSeek-R1} 
  & baseline & 33.0\% & 0.61 & 2.24 & 84.56 \\
& & ans-0 & 42.0\% ~($\uparrow$ 9.0\%) & 1.69 & 2.78 & 98.46 \\
& & rev-1 & 50.0\% ~($\uparrow$ 8.0\%) & 3.53 & 3.09 & 106.28 \\
& & rev-2 & 75.0\% ~($\uparrow$ 33.0\%) & 3.73 & 3.50 & 117.05 \\ \hline

\multirow{12}{*}{\textbf{Poisson}} & \multirow{4}{*}{GPT-4.1-mini} 
  & baseline & 67.0\% & 0.67 & 1.81 & 17.65 \\
& & ans-0 & 89.0\% ~($\uparrow$ 22.0\%) & 1.63 & 1.92 & 24.32 \\
& & rev-1 & 100\% ~($\uparrow$ 11.0\%) & 2.75 & 2.15 & 25.14 \\
& & rev-2 & 100\% ~($\uparrow$ 11.0\%) & 2.96 & 2.80 & 29.68 \\ \cline{2-7}
& \multirow{4}{*}{Gemini-2.5-flash} 
  & baseline & 60.0\% & 0.67 & 2.61 & 19.96 \\
& & ans-0 & 75.0\% ~($\uparrow$ 15.0\%) & 1.63 & 3.23 & 24.99 \\
& & rev-1 & 100\% ~($\uparrow$ 25.0\%) & 3.99 & 4.03 & 30.19 \\
& & rev-2 & 100\% ~($\uparrow$ 25.0\%) & 4.75 & 4.72 & 35.68 \\ \cline{2-7}
& \multirow{4}{*}{DeepSeek-R1} 
  & baseline & 63.0\% & 0.67 & 1.38 & 63.56 \\
& & ans-0 & 75.0\% ~($\uparrow$ 12.0\%) & 1.63 & 1.55 & 70.61 \\
& & rev-1 & 89.0\% ~($\uparrow$ 14.0\%) & 2.43 & 1.94 & 80.28 \\
& & rev-2 & 100\% ~($\uparrow$ 25.0\%) & 2.79 & 2.09 & 84.47 \\ \hline

\multirow{12}{*}{\textbf{Helm.}} & \multirow{4}{*}{GPT-4.1-mini} 
  & baseline & 67.0\% & 0.79 & 1.63 & 17.33 \\
& & ans-0 & 75.0\% ~($\uparrow$ 8.0\%) & 1.92 & 2.31 & 24.50 \\
& & rev-1 & 89.0\% ~($\uparrow$ 14.0\%) & 3.20 & 2.36 & 29.05 \\
& & rev-2 & 100\% ~($\uparrow$ 25.0\%) & 3.23 & 3.06 & 31.56 \\ \cline{2-7}
& \multirow{4}{*}{Gemini-2.5-flash} 
  & baseline & 75.0\% & 0.79 & 2.34 & 19.21 \\
& & ans-0 & 89.0\% ~($\uparrow$ 14.0\%) & 1.92 & 3.16 & 25.54 \\
& & rev-1 & 100\% ~($\uparrow$ 11.0\%) & 4.04 & 4.16 & 32.94 \\
& & rev-2 & 100\% ~($\uparrow$ 11.0\%) & 4.96 & 4.64 & 35.16 \\ \cline{2-7}
& \multirow{4}{*}{DeepSeek-R1} 
  & baseline & 70.0\% & 0.79 & 1.48 & 68.92 \\
& & ans-0 & 89.0\% ~($\uparrow$ 19.0\%) & 1.92 & 1.75 & 76.38 \\
& & rev-1 & 89.0\% ~($\uparrow$ 0.0\%) & 2.68 & 2.17 & 85.83 \\
& & rev-2 & 100\% ~($\uparrow$ 11.0\%) & 3.03 & 2.81 & 98.62 \\ \hline

\multirow{12}{*}{\textbf{Lid.}} & \multirow{4}{*}{GPT-4.1-mini} 
  & baseline & 13.0\% & 0.73 & 3.35 & 37.78 \\
& & ans-0 & 44.0\% ~($\uparrow$ 31.0\%) & 1.82 & 3.47 & 43.65 \\
& & rev-1 & 55.0\% ~($\uparrow$ 11.0\%) & 4.27 & 4.04 & 48.61 \\
& & rev-2 & 67.0\% ~($\uparrow$ 23.0\%) & 4.78 & 4.34 & 51.25 \\ \cline{2-7}
& \multirow{4}{*}{Gemini-2.5-flash} 
  & baseline & 33.0\% & 0.73 & 3.57 & 28.20 \\
& & ans-0 & 58.0\% ~($\uparrow$ 25.0\%) & 1.82 & 3.97 & 31.80 \\
& & rev-1 & 67.0\% ~($\uparrow$ 9.0\%) & 4.69 & 4.95 & 38.17 \\
& & rev-2 & 75.0\% ~($\uparrow$ 17.0\%) & 5.60 & 5.99 & 44.93 \\ \cline{2-7}
& \multirow{4}{*}{DeepSeek-R1} 
  & baseline & 20.0\% & 0.73 & 2.24 & 89.08 \\
& & ans-0 & 42.0\% ~($\uparrow$ 22.0\%) & 1.82 & 2.65 & 98.81 \\
& & rev-1 & 42.0\% ~($\uparrow$ 0.0\%) & 3.53 & 3.16 & 107.92 \\
& & rev-2 & 67.0\% ~($\uparrow$ 25.0\%) & 3.98 & 4.01 & 123.12 \\ \hline

\multirow{12}{*}{\textbf{NS}} & \multirow{4}{*}{GPT-4.1-mini} 
  & baseline & 13.0\% & 0.92 & 3.18 & 44.85 \\
& & ans-0 & 37.0\% ~($\uparrow$ 24.0\%) & 2.01 & 3.24 & 51.98 \\
& & rev-1 & 50.0\% ~($\uparrow$ 13.0\%) & 4.27 & 3.76 & 51.34 \\
& & rev-2 & 63.0\% ~($\uparrow$ 26.0\%) & 4.74 & 3.88 & 55.80 \\ \cline{2-7}
& \multirow{4}{*}{Gemini-2.5-flash} 
  & baseline & 13.0\% & 0.92 & 3.67 & 28.85 \\
& & ans-0 & 25.0\% ~($\uparrow$ 12.0\%) & 2.01 & 4.51 & 35.31 \\
& & rev-1 & 37.0\% ~($\uparrow$ 12.0\%) & 5.53 & 5.13 & 40.34 \\
& & rev-2 & 63.0\% ~($\uparrow$ 38.0\%) & 6.06 & 5.72 & 43.18 \\ \cline{2-7}
& \multirow{4}{*}{DeepSeek-R1} 
  & baseline & 25.0\% & 0.92 & 2.39 & 89.68 \\
& & ans-0 & 50.0\% ~($\uparrow$ 25.0\%) & 2.01 & 2.56 & 95.72 \\
& & rev-1 & 50.0\% ~($\uparrow$ 0.0\%) & 3.64 & 3.28 & 112.86 \\
& & rev-2 & 63.0\% ~($\uparrow$ 13.0\%) & 4.29 & 3.91 & 126.92 \\ \hline

\multirow{12}{*}{\textbf{Hilbert}} & \multirow{4}{*}{GPT-4.1-mini} 
  & baseline & ~0.0\% & 0.48 & 1.73 & 16.98 \\
& & ans-0 & ~0.0\% ~($\uparrow$ ~0.0\%) & 1.59 & 2.16 & 36.18 \\
& & rev-1 & 34.0\% ~($\uparrow$ 34.0\%) & 2.89 & 2.99 & 37.35 \\
& & rev-2 & 57.0\% ~($\uparrow$ 57.0\%) & 3.63 & 4.02 & 44.03 \\ \cline{2-7}
& \multirow{4}{*}{Gemini-2.5-flash} 
  & baseline & 15.0\% & 0.48 & 2.94 & 22.11 \\
& & ans-0 & 32.0\% ~($\uparrow$ 17.0\%) & 1.59 & 3.33 & 26.64 \\
& & rev-1 & 38.0\% ~($\uparrow$ 6.0\%) & 4.01 & 4.50 & 34.25 \\
& & rev-2 & 43.0\% ~($\uparrow$ 11.0\%) & 5.05 & 5.91 & 43.41 \\ \cline{2-7}
& \multirow{4}{*}{DeepSeek-R1} 
  & baseline & ~0.0\% & 0.48 & 1.66 & 71.92 \\
& & ans-0 & 11.0\% ~($\uparrow$ 11.0\%) & 1.59 & 1.96 & 78.52 \\
& & rev-1 & 32.0\% ~($\uparrow$ 21.0\%) & 2.69 & 2.47 & 94.64 \\
& & rev-2 & 46.0\% ~($\uparrow$ 35.0\%) & 3.11 & 2.81 & 101.72 \\ \hline

\multirow{12}{*}{\textbf{Keyhole}} & \multirow{4}{*}{GPT-4.1-mini} 
  & baseline & 13.0\% & 0.50 & 2.62 & 27.13 \\
& & ans-0 & 25.0\% ~($\uparrow$ 12.0\%) & 1.44 & 3.21 & 31.55 \\
& & rev-1 & 63.0\% ~($\uparrow$ 38.0\%) & 3.84 & 3.40 & 33.15 \\
& & rev-2 & 75.0\% ~($\uparrow$ 50.0\%) & 4.03 & 3.81 & 39.21 \\ \cline{2-7}
& \multirow{4}{*}{Gemini-2.5-flash} 
  & baseline & 33.0\% & 0.50 & 2.55 & 19.57 \\
& & ans-0 & 67.0\% ~($\uparrow$ 34.0\%) & 1.44 & 3.18 & 25.67 \\
& & rev-1 & 78.0\% ~($\uparrow$ 11.0\%) & 3.86 & 5.19 & 39.73 \\
& & rev-2 & 89.0\% ~($\uparrow$ 22.0\%) & 5.70 & 5.95 & 45.76 \\ \cline{2-7}
& \multirow{4}{*}{DeepSeek-R1} 
  & baseline & 20.0\% & 0.50 & 1.72 & 72.64 \\
& & ans-0 & 38.0\% ~($\uparrow$ 18.0\%) & 1.44 & 1.79 & 76.48 \\
& & rev-1 & 63.0\% ~($\uparrow$ 25.0\%) & 2.53 & 2.85 & 99.20 \\
& & rev-2 & 88.0\% ~($\uparrow$ 50.0\%) & 3.42 & 3.49 & 114.88 \\ \hline

\multirow{4}{*}{\textbf{Rayleigh}} & \multirow{4}{*}{GPT-4.1-mini} 
  & baseline & 25.0\% & 0.53 & 2.48 & 24.57 \\
& & ans-0 & 37.0\% ~($\uparrow$ 12.0\%) & 1.48 & 2.78 & 30.95 \\
& & rev-1 & 63.0\% ~($\uparrow$ 26.0\%) & 3.49 & 3.07 & 32.27 \\
& & rev-2 & 75.0\% ~($\uparrow$ 38.0\%) & 3.78 & 3.38 & 39.43 \\ 
\hline
\end{longtable}
\endgroup

{\color{black}

\subsection{Implementation records supporting the discussion}
\label{app:sample_evidence}

The implementation records are available under
\texttt{src/ipynb\_ver/sample\_results/}
in the repository cited in the Data Availability statement:

\begin{itemize}
    \item \texttt{\_3Programmers\_for\_Reproduction/}
    contains the stage-wise implementations generated by the three Programmers for the PDE benchmarks, Hilbert systems, and dimensional-analysis tasks.

    \item \texttt{\_GPT5mini\_/cylinder\_for\_Reproduction/}
    contains the two cylinder-flow examples.
    \texttt{sample1} and \texttt{sample2} correspond to the upper and lower rows of figure~\ref{fig:cylinder_opt_steps}, respectively.

    \item \texttt{\_GPT5mini\_/other\_example\_trials/}
    contains agent-interaction records for additional examples, including expanded problem descriptions, solution plans, generated implementations, execution outputs, and Reviewer feedback.
\end{itemize}

}

\bibliographystyle{unsrtnat}
\bibliography{main}

@PREAMBLE{
 "\providecommand{\noopsort}[1]{}" 
 # "\providecommand{\singleletter}[1]{#1}%" 
}

@article{Wang_etal_TAML_2025,
  title   = {Evaluations of Large Language Models in Computational Fluid Dynamics: Leveraging, Learning and Creating Knowledge},
  author  = {Wang, L and Zhang, L and He, G-W},
  journal = {Theor. Appl. Mech. Lett.},
  pages   = {100597},
  year    = {2025},
  doi     = {10.1016/j.taml.2025.100597},
}

@article{Jiang_etal_TAML_2025,
  title   = {{DeepSeek} vs. {ChatGPT} vs. {Claude}: a comparative study for scientific computing and scientific machine learning tasks},
  author  = {Jiang, Q L and Gao, Z W and Karniadakis, G E},
  journal = {Theor. Appl. Mech. Lett.},
  volume  = {15},
  number  = {3},
  pages   = {100583},
  year    = {2025},
  doi     = {10.1016/j.taml.2025.100583},
}

@article{Hao_etal_arXiv_2023,
  title   = {Pinnacle: A comprehensive benchmark of physics-informed neural networks for solving {PDEs}},
  author  = {Hao, Z K and Yao, J C and Su, C and Su, H and Wang, Z and others},
  journal = {arXiv preprint arXiv:2306.08827},
  year    = {2023},
  doi     = {10.48550/arXiv.2306.08827},
}

@article{Xie_etal_NatCommun_2022,
  title   = {Data-driven discovery of dimensionless numbers and governing laws from scarce measurements},
  author  = {Xie, X Y and Samaei, A and Guo, J C and Liu, W K and Gan, Z T},
  journal = {Nat. Commun.},
  volume  = {13},
  number  = {1},
  pages   = {7562},
  year    = {2022},
}

@article{Zhao_etal_PRX_2019,
  title   = {Bulk-explosion-induced metal spattering during laser processing},
  author  = {Zhao, C and Guo, Q L and Li, X and Parab, N and Fezzaa, K and Tan, W D and Chen, L Y and Sun, T},
  journal = {Phys. Rev. X},
  volume  = {9},
  number  = {2},
  pages   = {021052},
  year    = {2019},
}

@article{Gan_etal_NatCommun_2021,
  title   = {Universal scaling laws of keyhole stability and porosity in 3D printing of metals},
  author  = {Gan, Z T and Kafka, O L and Parab, N and Zhao, C and Fang, L C and Heinonen, O and Sun, T and Liu, W K},
  journal = {Nat. Commun.},
  volume  = {12},
  number  = {1},
  pages   = {2379},
  year    = {2021},
}

@article{Fabbro_etal_JLA_2018,
  title   = {Analysis and possible estimation of keyhole depths evolution, using laser operating parameters and material properties},
  author  = {Fabbro, R and Dal, M and Peyre, P and Coste, F and Schneider, M and Gunenthiram, V},
  journal = {J. Laser Appl.},
  volume  = {30},
  number  = {3},
  year    = {2018},
}

@article{Ye_etal_AEM_2019,
  title   = {Energy coupling mechanisms and scaling behavior associated with laser powder bed fusion additive manufacturing},
  author  = {Ye, J C and Khairallah, S A and Rubenchik, A M and Crumb, M F and Guss, G and Belak, J and Matthews, M J},
  journal = {Adv. Eng. Mater.},
  volume  = {21},
  number  = {7},
  pages   = {1900185},
  year    = {2019},
}

@book{barenblatt2003scaling,
  author={Barenblatt, G I},  
  title={Scaling},
  year={2003},
  publisher={Cambridge University Press},
  doi={10.1017/CBO9780511814921}
}

@article{yang2025large,
  title={Large language model driven development of turbulence models},
  author  = {Yang, Z X and Bin, Y W and Shi, Y P and Yang, X I A},
  journal={Flow},
  volume={5},
  pages={E40},
  year={2025},
  publisher={Cambridge University Press}
}

@article{Li_etal_arXiv_2025,
  title   = {{CodePDE}: An Inference Framework for {LLM}-driven {PDE} Solver Generation},
  author  = {Li, S D and Marwah, T and Shen, J H and Sun, W W and Risteski, A and Yang, Y M and Talwalkar, A},
  journal = {arXiv preprint arXiv:2505.08783},
  year    = {2025},
}

@article{Oskooi_etal_CPC_2010,
  title   = {{MEEP}: A flexible free-software package for electromagnetic simulations by the {FDTD} method},
  author  = {Oskooi, A F and Roundy, D and Ibanescu, M and Bermel, P and Joannopoulos, J D and Johnson, S G},
  journal = {Comput. Phys. Commun.},
  volume  = {181},
  number  = {3},
  pages   = {687--702},
  year    = {2010},
}

@article{ramprasad2017machine,
  title={Machine learning in materials informatics: recent applications and prospects},
  author={Ramprasad, R and Batra, R and Pilania, G and Mannodi-Kanakkithodi, A and Kim, C},
  journal={npj Comput. Mater.},
  volume={3},
  number={1},
  pages={54},
  year={2017},
  publisher={Nature Publishing Group UK London}
}

@article{Brunton_etal_ARFM_2020,
  title   = {Machine learning for fluid mechanics},
  author  = {Brunton, S L and Noack, B R and Koumoutsakos, P},
  journal = {Annu. Rev. Fluid Mech.},
  volume  = {52},
  number  = {1},
  pages   = {477--508},
  year    = {2020},
}

@article{Meng_Yang_PRR_2023,
  title   = {Quantum computing of fluid dynamics using the hydrodynamic {Schr{\"o}dinger} equation},
  author  = {Meng, Z Y and Yang, Y},
  journal = {Phys. Rev. Res.},
  volume  = {5},
  number  = {3},
  pages   = {033182},
  year    = {2023},
}

@article{achiam2023gpt,
  title={Gpt-4 technical report},
  author={Achiam, J and Adler, S and Agarwal, S and Ahmad, L and Akkaya, I and Aleman, F L and Almeida, D and Altenschmidt, J and Altman, S and Anadkat, S and others},
  journal={arXiv preprint arXiv:2303.08774},
  year={2023}
}

@article{wang2023codet5+,
  title={Codet5+: Open code large language models for code understanding and generation},
  author={Wang, Y and Le, H and Gotmare, A D and Bui, N D and Li, J N and Hoi, S C},
  journal={arXiv preprint arXiv:2305.07922},
  year={2023}
}

@article{kojima2022large,
  title={Large language models are zero-shot reasoners},
  author={Kojima, T and Gu, S X Shane and Reid, M and Matsuo, Y and Iwasawa, Y},
  journal={Adv. Neural Inf. Process. Syst.},
  volume={35},
  pages={22199--22213},
  year={2022}
}

@article{zhao2023survey,
  title={A survey of large language models},
  author={Zhao, W X and Zhou, K and Li, J Y and Tang, T Y and Wang, X L and Hou, Y P and Min, Y Q and Zhang, B C and Zhang, J J and Dong, Z C and others},
  journal={arXiv preprint arXiv:2303.18223},
  year={2023}
}

@article{wei2022chain,
  title={Chain-of-thought prompting elicits reasoning in large language models},
  author={Wei, J and Wang, X Z and Schuurmans, D and Bosma, M and Xia, F and Chi, E and Le, Q V and Zhou, D and others},
  journal={Adv. Neural Inf. Process. Syst.},
  volume={35},
  pages={24824--24837},
  year={2022}
}

@article{roziere2023code,
  title={Code llama: Open foundation models for code},
  author={Roziere, B and Gehring, J and Gloeckle, F and Sootla, S and Gat, I and Tan, X Q and Adi, Y and Liu, J Y and Sauvestre, R and Remez, T and others},
  journal={arXiv preprint arXiv:2308.12950},
  year={2023}
}

@article{jiang2024survey,
  title={A survey on large language models for code generation},
  author={Jiang, J Y and Wang, F and Shen, J S and Kim, S J and Kim, S H},
  journal={arXiv preprint arXiv:2406.00515},
  year={2024}
}

@article{guo2025deepseek,
  title={Deepseek-r1: Incentivizing reasoning capability in llms via reinforcement learning},
  author={Guo, D Y and Yang, D J and Zhang, H W and Song, J X and Zhang, R Y and Xu, R X and Zhu, Q H and Ma, S R and Wang, P Y and Bi, X and others},
  journal={arXiv preprint arXiv:2501.12948},
  year={2025}
}

@article{madaan2023self,
  title={Self-refine: Iterative refinement with self-feedback},
  author={Madaan, A and Tandon, N and Gupta, P and Hallinan, S and Gao, L Y and Wiegreffe, S and Alon, U and Dziri, N and Prabhumoye, S and Yang, Y M and others},
  journal={Adv. Neural Inf. Process. Syst.},
  volume={36},
  pages={46534--46594},
  year={2023}
}

@article{chen2021evaluating,
  title={Evaluating large language models trained on code},
  author={Chen, M and Tworek, J and Jun, H W and Yuan, Q M and Pinto, H P D O and Kaplan, J and Edwards, H and Burda, Y and Joseph, N and Brockman, G and others},
  journal={arXiv preprint arXiv:2107.03374},
  year={2021}
}

@inproceedings{hong2023metagpt,
  title={MetaGPT: Meta programming for a multi-agent collaborative framework},
  author={Hong, S R and Zhuge, M C and Chen, J and Zheng, X W and Cheng, Y H and Wang, J L and Zhang, C Y and Wang, Z L and Yau, S K S and Lin, Z J and others},
  booktitle={The Twelfth International Conference on Learning Representations},
  year={2023}
}

@article{chan2023chateval,
  title={Chateval: Towards better llm-based evaluators through multi-agent debate},
  author={Chan, C-M and Chen, W Z and Su, Y S and Yu, J X and Xue, W and Zhang, S H and Fu, J and Liu, Z Y},
  journal={arXiv preprint arXiv:2308.07201},
  year={2023}
}

@article{wuwu2025pinnsagent,
  title={PINNsAgent: Automated PDE Surrogation with Large Language Models},
  author={Wuwu, Q P and Gao, C H and Chen, T Y and Huang, Y H and Zhang, Y K and Wang, J N and Li, J X and Zhou, H Y and Zhang, S H},
  journal={arXiv preprint arXiv:2501.12053},
  year={2025}
}

@article{xu2025cfdagent,
  title={CFDAgent: A language-guided, zero-shot multi-agent system for complex flow simulation},
  author={Xu, Z Y and Wang, L and Wang, C Y and Chen, Y X and Luo, Q Y and Yao, H-D and Wang, S Z and He, G-W},
  journal={Phys. Fluids},
  volume={37},
  number={11},
  year={2025},
  publisher={AIP Publishing}
}

@article{pandey2025openfoamgpt,
  title={OpenFOAMGPT: A retrieval-augmented large language model (LLM) agent for OpenFOAM-based computational fluid dynamics},
  author={Pandey, S and Xu, R and Wang, W K and Chu, X},
  journal={Phys. Fluids},
  volume={37},
  number={3},
  year={2025},
  publisher={AIP Publishing}
}

@article{zhang2022hidenn,
  title={HiDeNN-TD: reduced-order hierarchical deep learning neural networks},
  author={Zhang, L and Lu, Y and Tang, S Q and Liu, W K},
  journal={Comput. Methods Appl. Mech. Eng.},
  volume={389},
  pages={114414},
  year={2022},
  publisher={Elsevier}
}

@article{ghia1982high,
  title={High-Re solutions for incompressible flow using the Navier-Stokes equations and a multigrid method},
  author={Ghia, U and Ghia, K N and Shin, C},
  journal={J. Comput. Phys.},
  volume={48},
  number={3},
  pages={387--411},
  year={1982},
  publisher={Elsevier}
}

@article{sod1978survey,
  title={A survey of several finite difference methods for systems of nonlinear hyperbolic conservation laws},
  author={Sod, G A},
  journal={J. Comput. Phys.},
  volume={27},
  number={1},
  pages={1--31},
  year={1978},
  publisher={Elsevier}
}

@incollection{kress1998ill,
  title={Ill-conditioned linear systems},
  author={Kress, R},
  booktitle={Numer. Anal.},
  pages={77--92},
  year={1998},
  publisher={Springer}
}

@article{benton1972table,
  title={A table of solutions of the one-dimensional Burgers equation},
  author={Benton, E R and Platzman, G W},
  journal={Q. Appl. Math.},
  volume={30},
  number={2},
  pages={195--212},
  year={1972}
}

@article{chorin1967numerical,
  title={A numerical method for solving incompressible viscous flow problems},
  author={Chorin, A J},
  journal={J. Comput. Phys.},
  volume={2},
  number={1},
  pages={12--26},
  year={1967},
  publisher={Elsevier}
}

@book{strauss2007partial,
  title={Partial differential equations: An introduction},
  author={Strauss, W A},
  year={2007},
  publisher={John Wiley \& Sons}
}

@misc{openai_o3_mini_2025,
    author = {{OpenAI}},
    title = {o3-mini-2025-01-31},
    year = {2025},
    url = {https://platform.openai.com/docs/models/o3-mini},
    note = {Accessed: 2025-03-10}
}

@misc{deepseek_r1_20250528,
    author = {{DeepSeek-AI}},
    title = {DeepSeek-R1-0528},
    year = {2025},
    url = {https://api-docs.deepseek.com},
    note = {Accessed: 2025-07-10}
}

@misc{gemini_2.5flash_preview-0520,
    author = {{Google DeepMind}},
    title = {Gemini-2.5-flash-preview-05-20},
    year = {2025},
    url = {https://ai.google.dev/gemini-api/docs/models#gemini-2.5-flash},
    note = {Accessed: 2025-07-10}
}

@misc{gpt4p1_mini_050414,
    author = {{OpenAI}},
    title = {gpt-4.1-mini-2025-04-14},
    year = {2025},
    url = {https://platform.openai.com/docs/models/gpt-4.1-mini},
    note = {Accessed: 2025-07-10}
}

@misc{gpt5_mini_250807,
    author = {{OpenAI}},
    title = {gpt-5-mini-2025-08-07},
    year = {2025},
    url = {https://developers.openai.com/api/docs/models/gpt-5-mini},
    note = {Accessed: 2026-04-10}
}

@article{saad1986gmres,
  title={GMRES: A generalized minimal residual algorithm for solving nonsymmetric linear systems},
  author={Saad, Y and Schultz, M H},
  journal={SIAM J. Sci. Stat. Comput.},
  volume={7},
  number={3},
  pages={856--869},
  year={1986},
  publisher={SIAM}
}

@article{golub1999tikhonov,
  title={Tikhonov regularization and total least squares},
  author={Golub, G H and Hansen, P C and O'Leary, D P},
  journal={SIAM J. Matrix Anal. Appl.},
  volume={21},
  number={1},
  pages={185--194},
  year={1999},
  publisher={SIAM}
}

@article{hestenes1952methods,
  title={Methods of conjugate gradients for solving linear systems},
  author={Hestenes, M R and Stiefel, E and others},
  journal={J. Res. Natl. Bur. Stand.},
  volume={49},
  number={6},
  pages={409--436},
  year={1952}
}

@misc{langgraph_2025,
    author = {{LangChain AI}},
    title = {LangGraph: Build LLM applications with stateful graphs},
    year = {2025},
    url = {https://github.com/langchain-ai/langgraph},
    note = {Accessed: 2025-07-15}
}

@article{peskin2002immersed,
  title={The immersed boundary method},
  author={Peskin, C S},
  journal={Acta Numer.},
  volume={11},
  pages={479--517},
  year={2002},
  publisher={Cambridge University Press}
}

@article{mittal2005immersed,
  title={Immersed boundary methods},
  author={Mittal, R and Iaccarino, G},
  journal={Annu. Rev. Fluid Mech.},
  volume={37},
  number={1},
  pages={239--261},
  year={2005},
  publisher={Annual Reviews}
}

@incollection{schafer1996benchmark,
  title={Benchmark computations of laminar flow around a cylinder},
  author={Sch{\"a}fer, M and Turek, S and Durst, F and Krause, E and Rannacher, R},
  booktitle={Flow simulation with high-performance computers II: DFG priority research programme results 1993--1995},
  pages={547--566},
  year={1996},
  publisher={Springer}
}

@article{chorin1968numerical,
  title={Numerical solution of the Navier-Stokes equations},
  author={Chorin, A J},
  journal={Math. Comput.},
  volume={22},
  number={104},
  pages={745--762},
  year={1968}
}

@article{niemela2006turbulent,
  title={Turbulent convection at high Rayleigh numbers and aspect ratio 4},
  author={Niemela, J and Sreenivasan, K R},
  journal={J. Fluid Mech.},
  volume={557},
  pages={411--422},
  year={2006},
  publisher={Cambridge University Press}
}

@article{chavanne2001turbulent,
  title={Turbulent Rayleigh--B{\'e}nard convection in gaseous and liquid He},
  author={Chavanne, X and Chilla, F and Chabaud, B and Castaing, B and Hebral, B},
  journal={Phys. Fluids},
  volume={13},
  number={5},
  pages={1300--1320},
  year={2001},
  publisher={American Institute of Physics}
}

@article{nichol2022point,
  title={Point-e: A system for generating 3d point clouds from complex prompts},
  author={Nichol, A and Jun, H W and Dhariwal, P and Mishkin, P and Chen, M},
  journal={arXiv preprint arXiv:2212.08751},
  year={2022}
}

@inproceedings{abdaljalil2025theorem,
  title={Theorem-of-thought: a multi-agent framework for abductive, deductive, and inductive reasoning in language models},
  author={Abdaljalil, S and Kurban, H and Qaraqe, K and Serpedin, E},
  booktitle={Proceedings of the 3rd Workshop on Towards Knowledgeable Foundation Models (KnowFM)},
  pages={111--119},
  year={2025}
}

@article{polat2025xchemagents,
  title={xchemagents: Agentic ai for explainable quantum chemistry},
  author={Polat, C and Tuncel, M and Kurban, M and Serpedin, E and Kurban, H},
  journal={arXiv preprint arXiv:2505.20574},
  year={2025}
}

@article{abdaljalil2025safe,
  title={Safe: A sparse autoencoder-based framework for robust query enrichment and hallucination mitigation in llms},
  author={Abdaljalil, S and Pallucchini, F and Seveso, A and Kurban, H and Mercorio, F and Serpedin, E},
  journal={Findings Assoc. Comput. Linguist. EMNLP},
  year={2025}
}

@article{zhang2025knowledge,
  title={Knowledge-integrated additive learning for consistent near-wall modelling of turbulent flows},
  author={Zhang, F S and Zhou, Z D and Yang, X L and He, G-W},
  journal={J. Fluid Mech.},
  volume={1011},
  pages={R1},
  year={2025},
  publisher={Cambridge University Press}
}

@article{zhou2025wall,
  title={A wall model for separated flows: embedded learning to improve a posteriori performance},
  author={Zhou, Z D and Zhang, X-L and He, G-W and Yang, X L},
  journal={J. Fluid Mech.},
  volume={1002},
  pages={A3},
  year={2025},
  publisher={Cambridge University Press}
}

@article{meng2026toward,
  title={Toward end-to-end quantum simulation of rapidly distorted turbulence},
  author={Meng, Z Y and Chen, L Y and Liu, J-P and He, G-W},
  journal={J. Comput. Phys.},
  pages={114888},
  year={2026},
  publisher={Elsevier}
}

@article{zhang2026differentiable,
  title={A differentiable wall-modeled large-eddy simulation method for high-Reynolds-number wall-bounded turbulent flows},
  author={Zhang, F S and Yang, X L and He, G-W},
  journal={J. Comput. Phys.},
  pages={114835},
  year={2026},
  publisher={Elsevier}
}

@article{wang25solvable,
  title = {Solvable model for space-time correlations of passive scalars advected by turbulent flows with mean velocity},
  author = {Wang, L and He, G-W},
  journal = {Phys. Rev. Fluids},
  volume = {10},
  issue = {10},
  pages = {104607},
  numpages = {20},
  year = {2025},
  month = {Oct},
  publisher = {American Physical Society},
  doi = {10.1103/x644-pdqq},
}

@article{Liu2026factorized,
  author  = {Liu, H-C and Luo, Q Y and Zhang, X-L and He, G-W},
  title   = {Factorized-gating mixture of experts enables cross-regime turbulence modeling},
  journal = {Comms. Eng.},
  year    = {2026},
  doi     = {10.1038/s44172-026-00802-5},
  issn    = {2731-3395}
}

\end{document}